\PassOptionsToPackage{table}{xcolor}
\documentclass{article} 
\usepackage{iclr2025_conference,times}

\usepackage{hyperref}
\usepackage{url}
\usepackage[utf8]{inputenc} 
\usepackage[T1]{fontenc}    
\usepackage{booktabs}       
\usepackage{amsfonts}       
\usepackage{nicefrac}       
\usepackage{microtype}      

\usepackage{multirow}
\usepackage{graphicx}
\usepackage{makecell}
 \usepackage{multirow}
 \usepackage{arydshln}
 \usepackage{enumitem}
 \usepackage{svg} 

\usepackage{amssymb}
\usepackage{bbding}
\usepackage{pifont}
\usepackage{wasysym}
\usepackage{utfsym}
\usepackage{fontawesome}
\usepackage{caption}
\usepackage{tabularx}
\usepackage{wrapfig}
\usepackage[normalem]{ulem}

\usepackage{array}
\usepackage{colortbl}

\usepackage{caption}
\usepackage{subcaption}
\usepackage{multirow}
\usepackage{graphicx}
\usepackage{listings}

\title{Dysca: A Dynamic and Scalable Benchmark for Evaluating Perception Ability of LVLMs}

\author{Jie Zhang$^{12}$, Zhongqi Wang$^{12}$, Mengqi Lei$^{3}$, Zheng Yuan$^{12}$,\\
\textbf{Bei Yan}$^{12}$\textbf{,}  \textbf{Shiguang Shan}$^{12}$\textbf{,} \textbf{Xilin Chen}$^{12}$ \\
$^{1}$ Key Laboratory of AI Safety of CAS, Institute of Computing Technology,\\ Chinese Academy of Sciences (CAS), Beijing, China \\
$^2$ University of Chinese Academy of Sciences, Beijing, China \\
$^3$ China University of Geosciences\\
}

\iclrfinalcopy 
\begin{document}

\maketitle

\begin{abstract}
Currently many benchmarks have been proposed to evaluate the perception ability of the Large Vision-Language Models (LVLMs).
However, most benchmarks conduct questions by selecting images from existing datasets, resulting in the potential data leakage.  
Besides, these benchmarks merely focus on evaluating LVLMs on the realistic style images and clean scenarios, leaving the multi-stylized images and noisy scenarios unexplored. 
In response to these challenges, we propose a dynamic and scalable benchmark named Dysca for evaluating LVLMs by leveraging synthesis images. 
Specifically, we leverage Stable Diffusion and design a rule-based method to dynamically generate novel images, questions and the corresponding answers. 
We consider 51 kinds of image styles and evaluate the perception capability in 20 subtasks.
Moreover, we conduct evaluations under 4 scenarios (i.e., Clean, Corruption, Print Attacking and Adversarial Attacking) and 3 question types (i.e., Multi-choices, True-or-false and Free-form). Thanks to the generative paradigm, Dysca serves as a scalable benchmark for easily adding new subtasks and scenarios.
A total of 24 advanced open-source LVLMs and 2 close-source LVLMs are evaluated on Dysca, revealing the drawbacks of current LVLMs. 
The benchmark is released at \url{https://github.com/Robin-WZQ/Dysca}.
\end{abstract}

\section{Introduction}

Recent years have witnessed the great success of the Large Vision-Language Models (LVLMs)~\citep{li2023blip2,zhu2023minigpt4,dai2023instructblip,liu2023visual_llava,li2023otter,chen2023shikra,internlmxcomposer,su2023pandagpt, gong2023multimodalgpt,Emu2}. 
These models leverage the powerful Large Language Models (LLMs)~\citep{chung2022scaling_flant5,ChatGPT,touvron2023llama,openai2023gpt4,vicuna} as their brain and incorporate the state-of-the-art visual encoders~\citep{radford2021learning_clip,EVA,dosovitskiy2020image——} as their eyes. 
Thanks to the alignment of visual feature with textual space and the development of visual instruction tuning techniques~\citep{liu2023visual_llava}, LVLMs showcase the impressive capability in terms of visual scene comprehension and multimodal instruction-following.

In order to comprehensively evaluate the capabilities of LVLMs, many benchmarks have been purposed~\citep{Agrawal2016VQA,TowardsVQA,xu2023lvlmehub,shao2023tiny,li2023seedbench,li2023seedbench2,fu2023mme,bai2023touchstone,yu2023mmvet,yang2023dawn,chen2024right}, where we categorize the current benchmarks into three types~\citep{fu2023mme}. 
The first type is the classical benchmarks, such as COCO Caption~\citep{chen2015microsoft} and VQA~\citep{Agrawal2016VQA,VQAmatter2017,qkvqa2019}.  Although these benchmarks provide high-quality evaluation data, they also have notable limitations.
On the one hand, they are inadequate for measuring the fine-grained capabilities of current LVLMs, offering the limited insightful feedback for the future improvement.
On the other hand, since these classical benchmarks have been available as the open-source test data for a long time, it is hard to prevent the data leakage problem.  
The second type of benchmarks evaluate the LVLMs through a subjective manner~\citep{yang2023dawn,wu2023early}. 
Although the benchmarks reveal the insightful drawbacks of current models, their data scale is limited (i.e., less than 200 annotations) and they require manual evaluation by experts.
The third type is built for objectively evaluating current LVLMs and the comparison between them are shown in Tab. \ref{tab:Comparisons}. 
They provide an objective and automatic evaluation manner, giving the fine-grained evaluation for the LVLMs.
However, these benchmarks conduct Vision-language QAs by selecting images from existing dataset and annotate the textual questions.
Although they claim that the questions are re-annotated, the previous work~\citep{chen2024right} has demonstrated that these benchmarks have Models unintentionally leaked into the training data of LLMs and LVLMs. 
Besides, most benchmarks focus on evaluating LVLMs in the realistic images and clean scenarios, leaving the multi-stylized images and noisy scenarios unexplored. 
While some works like MMCBench~\citep{zhang2024benchmarking} and Typographic Dataset~\citep{Cheng2024UnveilingTD} have investigated the robustness of LVLMs with corrupted and print-attacked images, respectively, they have not explored the effect of these noisy images on various  perceptual tasks.

\begin{figure}[tb]
  \centering
  \includegraphics[height=7.2cm]{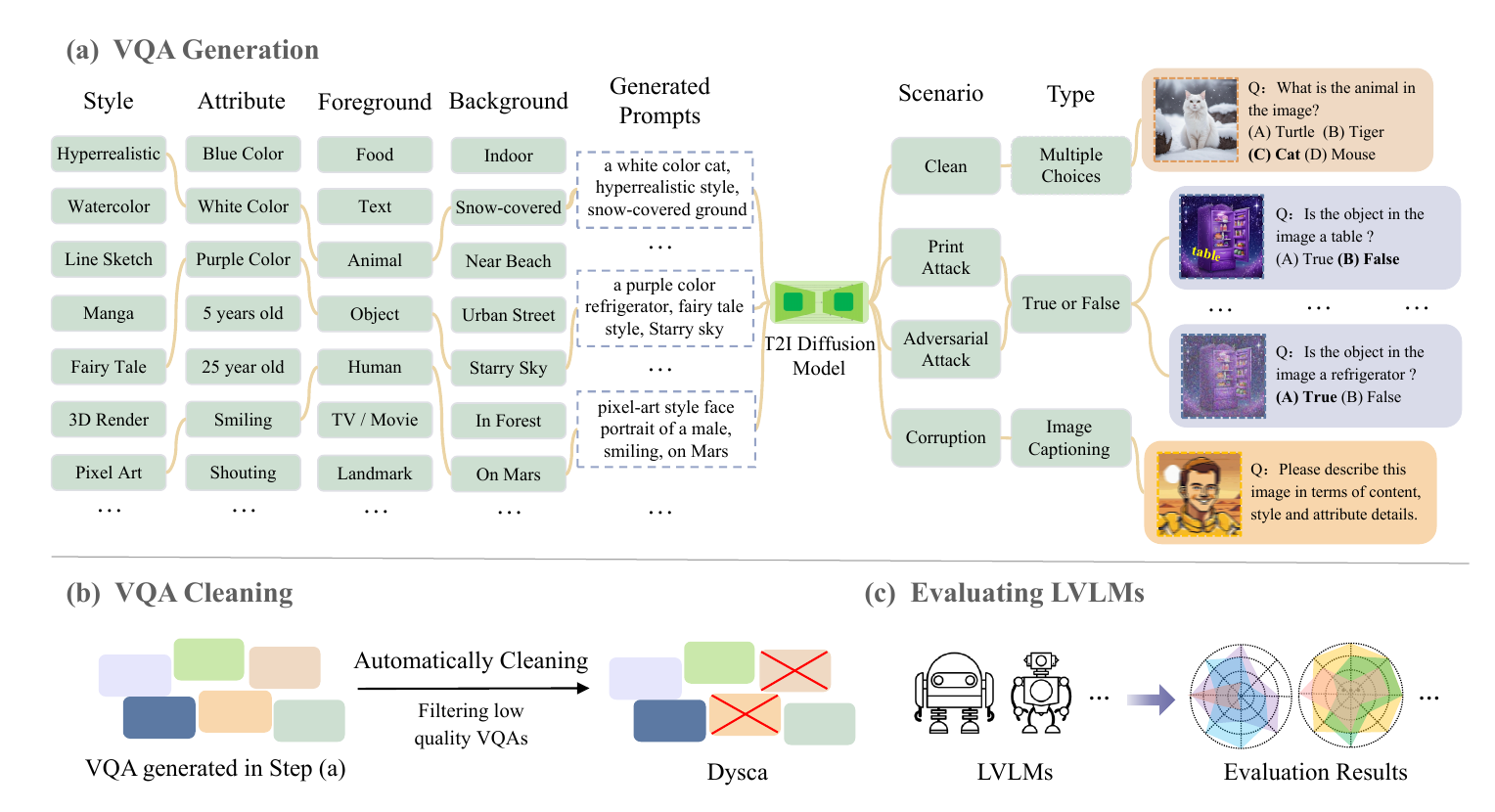}
  \caption{Overview of the automatic pipeline for generating Vision-language QAs, cleaning Vision-language QAs and evaluating LVLMs. \textbf{(\textit{a})} We first constructs prompts in terms of content, style and background, leveraging the Text-to-Image (T2I) diffusion model (e.g., SDXL~\citep{podell2023sdxl}) to synthesis images to be asked. Then based on the scenarios and the question type, we post-process the synthesis images and generate the specific textual questions, respectively. \textbf{(\textit{b})} We further filter out low quality Vision-language QAs by utilizing trained models to form the final Dysca. \textbf{(\textit{c})} Finally, we evaluate LVLMs on our Dysca and feedback the fine-grained evaluation results.
  }
  \label{fig:framework}
  \vspace{-0.6cm}
\end{figure}

In this paper, aiming to address these challenges above, we propose Dysca which is a dynamic and scalable benchmark for evaluating the perception ability of LVLMs via various subtasks and scenarios. 
Inspired by the prior evaluation works for LLMs~\citep{liang2023holistic}, we investigate on whether we could leverage the large-scale synthesized images for evaluating LVLMs. 
We display the overview of our pipeline in Fig. \ref{fig:framework}.
Specifically, we leverage Stable Diffusion and design a rule-based method to dynamically generate novel images, questions and corresponding answers. 
We decouple the prompt into 4 part, i.e., attribute, foreground, style and background, and design pre-defined templates to dynamically generate prompts, as displayed in Fig. \ref{fig:mpiq}. 
Then we utilize the state-of-the-art text-to-image diffusion models (e.g., SDXL~\citep{podell2023sdxl}) to generate the corresponding images. 
Since we already know the main information of the images through prompts, we easily generate question-answer textual pairs by the rule-based method.
After that, in order to obtain the high quality Vision-language QAs, we employ CLIP~\citep{radford2021learning_clip} to perform data cleaning on the generated Vision-language QA pairs.
Dysca focuses on assessing the fine-grained perception abilities, including recognizing human, animal, object, landmark, etc. Dysca evaluates LVLMs with \textbf{20 perceptual subtasks}, containing a total number of \textbf{51 different artistic styles}.  
Besides, to evaluate the robustness of the models across different scenarios and question types, we construct \textbf{4 testing scenarios} (clean, corruption, print attacking and adversarial attacking) and \textbf{3 question types} (multi-choices, true-or-false and free-form questions).

Compared to previous works in Tab. \ref{tab:Comparisons}, we provide an end-to-end process from image to Vision-QA generation. This approach significantly reduces annotation costs compared to manually labeling images (e.g., MME~\citep{fu2023mme}) while achieving the correctness for evaluating LVLMs. It also avoids the risk of hallucinate annotations that may occur when using ChatGPT for labeling based on image prompts (e.g., JourneyDB~\citep{NEURIPS2023_9bc59aff}). This novel pipeline enables us to create a benchmark that is easily scalable and adaptable for incorporating new subtasks and scenarios. Thanks to the generative paradigm, Dysca can be customized to meet the specific requirements of the evaluator for testing purposes.

In summary, our work makes the following key contributions:

\begin{itemize}
\item \textbf{Dynamic and Scalable Benchmark:} We propose Dysca, a benchmark 
 that is able to dynamically generate the test data that users need and is easily to scale up to to new subtasks and scenarios. 
\item \textbf{Multi-grained Perceptual Subtasks and Multi-scenarios:} Dysca aims to testing LVLMs’ performance on diverse styles, 4 image scenarios (i.e., clean, corruption, print attacking and adversarial attacking) and 3 question types (i.e., multi-choices, true-or-false and free-form questions), reporting the 20 perceptual subtasks performance of 26 mainstream LVLMs, including GPT-4o~\citep{OpenAI} and Gemini-1.5-Pro~\citep{geminiteam2024geminifamilyhighlycapable}.
\item  \textbf{Analysis and Observations: } We demonstrate for the first time that evaluating LVLMs using large-scale synthetic data is valid. Experiments show the strong correlation coefficient between our evaluation rankings and the rankings obtained from non-synthetic benchmarks. The evaluation results also reveal the weakness of current LVLMs when facing different question types, image styles and image scenarios.
\end{itemize}


\begin{table}[tb]
\renewcommand{\arraystretch}{1.2}
\centering
\caption{Comparisons between existing LVLM benchmarks. '\Checkmark\kern-1.2ex\raisebox{1ex}{\rotatebox[origin=c]{125}{\textbf{--}}}' indicates that the benchmarks include both newly collected images / annotations and images / annotations gathered from existing datasets. '*' The scale of our released benchmark is 617K, however Dysca is able to generate unlimited data to be tested.}
\label{tab:Comparisons}
\scalebox{0.78}{
\begin{tabular}{ccccccc}
\toprule
\textbf{Benchmark} & \textbf{\makecell{\#Evaluation \\ Data Scale}} & \textbf{\makecell{\#Perceptual \\Tasks}} & \textbf{\begin{tabular}[c]{@{}c@{}}Automatic\\ Annotation\end{tabular}} & \textbf{\makecell{Novel Images $\&$\\Novel Questions}} & \textbf{\makecell{Question \\Type}} & \textbf{\begin{tabular}[c]{@{}c@{}}Automatic\\ Evaluation\end{tabular}} \\ \midrule
LLaVA-Bench & 0.15K & - & \ding{55} & \Checkmark\kern-1.2ex\raisebox{1ex}{\rotatebox[origin=c]{125}{\textbf{--}}}  & Free-form & \ding{51} \\
MME  & 2.3K & 10 & \ding{55} & \Checkmark\kern-1.2ex\raisebox{1ex}{\rotatebox[origin=c]{125}{\textbf{--}}}  & True-or-false & \ding{51}\\
LVLM-eHub & - & 3 & \ding{51} & \ding{55} & Free-form & \ding{55} \\
tiny-LVLM-eHub & 2.1K & 3 & \ding{51} & \ding{55} & Free-form & \ding{51} \\
SEED-Bench & 19K & 8 & \Checkmark\kern-1.2ex\raisebox{1ex}{\rotatebox[origin=c]{125}{\textbf{--}}}  & \ding{55} & Multi-choices & \ding{51}  \\
MMBench & 2.9K & 12 & \ding{55} & \Checkmark\kern-1.2ex\raisebox{1ex}{\rotatebox[origin=c]{125}{\textbf{--}}}  & Multi-choices & \ding{51}  \\
TouchStone & 0.9K & 10 & \ding{55} & {\ding{51}} & Free-form & \ding{51} \\
REFORM-EVAL & 50K & 7 & \ding{51} & \ding{55} & Multi-choices & \ding{51} \\
MM-BigBench & 30K & 6 & \ding{51} & \ding{55} & Multi-choices & \ding{51}  \\
MM-VET & 0.2K & 4 & \Checkmark\kern-1.2ex\raisebox{1ex}{\rotatebox[origin=c]{125}{\textbf{--}}}  & \Checkmark\kern-1.2ex\raisebox{1ex}{\rotatebox[origin=c]{125}{\textbf{--}}}  & Free-form & \ding{51} \\
MLLM-Bench & 0.42K & 7 & \ding{55} & \Checkmark\kern-1.2ex\raisebox{1ex}{\rotatebox[origin=c]{125}{\textbf{--}}}  & Free-form & \ding{51}  \\
SEED-Bench2 & 24K & 10 & \Checkmark\kern-1.2ex\raisebox{1ex}{\rotatebox[origin=c]{125}{\textbf{--}}}  & \ding{55} & Multi-choices & \ding{51}  \\
BenchLMM & 2.4K & 15 & \ding{55} & \ding{55} & Free-form & \ding{51} \\ 
JourneyDB & 5.4K & 2 & \ding{51} & \ding{51} & \begin{tabular}[c]{@{}c@{}}Free-form\\ Multi-choices \end{tabular} & \ding{51}  \\ \cdashline{1-7} 
Dysca (Ours) & 617K* & 20 & \ding{51} & \ding{51} & \begin{tabular}[c]{@{}c@{}}  Free-form\\ Multi-choices\\ True-or-false\end{tabular} & \ding{51} \\ \bottomrule
\end{tabular}
}
\vspace{-0.5cm}
\end{table}

\vspace{-0.2cm}
\section{Related Works}
\vspace{-0.2cm}
\subsection{Large Vision-Language Models}

The landscape of Large Vision-Language Models (LVLMs) has been significantly shaped by the pioneering success of Large Language Models (LLMs) such as GPTs~\citep{radford2019language_gpt2,brown2020gpt3,ouyang2022GPT} and LLaMA~\citep{touvron2023llama}, catalyzing advancements in multimodal content understanding and generation~\citep{zhang2024mm_survey}, including intricate tasks like image-text comprehension.
At the forefront of these developments, BLIP-2~\citep{li2023blip2} introduces a lightweight Q-Former~\citep{li2023blip2} that facilitates alignment between textual and visual representations through a cross-attention mechanism~\citep{li2023blip2}. InstructBLIP~\citep{dai2023instructblip} takes a step further by incorporating textual instructions into the Q-Former, which significantly improves the zero-shot performance.  LLAVA~\citep{liu2023visual_llava} employs GPT-4~\citep{openai2023gpt4} to transform data into multimodal instruction-following data and uses CLIP~\citep{radford2021learning_clip} and LLAMA~\citep{touvron2023llama} for fine-tuning instructions, achieving advanced multimodal chat abilities. LLAVA-1.5~\citep{liu2023llava_1-5} extends this paradigm by integrating MLP projection and introducing academic task-specific Vision-language QA data. Recently, models like Otter~\citep{li2023otter}, MiniGPT-4~\citep{zhu2023minigpt4}, Qwen-VL-Chat~\citep{bai2023qwen-vl} and XComposer-VL~\citep{internlmxcomposer} further unleash the cross-modal understanding capabilities of LVLMs. Besides, many powerful closed-source LVLMs,
including Gemini-1.5-Pro~\citep{geminiteam2024geminifamilyhighlycapable} and GPT-4o~\citep{OpenAI}, have publicly released
their APIs, promoting the development of downstream applications.

\vspace{-0.2cm}
\subsection{Benchmarks for LVLMs}

The great progress of LVLMs triggers the development of benchmarks for evaluating these models, where we divide previous benchmarks into three categories. 
The first type is the classical benchmarks which focuses on evaluating LVLMs abilities via image caption~\citep{mscoco} and VQA~\citep{antol2015vqa,Agrawal2016VQA}.  
However, these benchmarks cannot provide the fine-grained feedback on how to improve the models. Besides, since these benchmarks have been the public resources for a long time, it is hardly to guarantee that the LVLMs have not use them for training. 
The second type subjectively evaluates LVLMs by experts~\citep{yang2023dawn,wu2023early}. Although these benchmarks reveal the insightful feedback of the LVLMs, their scale is limited (i.e., less than 200 annotations). The subjective manner also makes the evaluation expensive and hardly to expand the scale.  

The third type~\citep{liu2023visual_llava,fu2023mme,xu2023lvlmehub,shao2023tiny,li2023seedbench,li2023seedbench2,liu2023mmbench,bai2023touchstone,li2023reformeval,yang2023mmbigbench,yu2023mmvet,ge2023mllmbench,cai2023benchlmm,chen2024right,liu2024hidden} focuses on evaluating LVLMs in an objective and large-scaled manner, where we list the detailed information of them in the Tab. \ref{tab:Comparisons}. Some of them have been adopted by the community~\citep{2023opencompass} as the standard benchmarks for evaluating LVLMs~\citep{ChatGPT,internlmxcomposer}, like MME~\citep{fu2023mme} and MMBench~\citep{liu2023mmbench}.
These benchmarks evaluate models through the objective answer types and most of them leverage the automatic annotation and evaluation manner for revealing the fine-grained drawbacks of current LVLMs. 
However, the previous benchmarks primarily concentrate on evaluating LVLMs using realistic images under clean scenario, leaving multi-stylized images and noisy scenarios unexplored. Moreover, many of them conduct QA by selecting images from publicly available datasets (e.g.,~\citep{mscoco,Russakovsky2014ImageNetLS}). While they state that the questions have been re-annotated, they cannot guarantee that the LVLMs have not seen the image during training stage. The previous work~\citep{chen2024right} has proved that these benchmarks have unintentionally leaked into the training data of LLMs and LVLMs. One possible way to solve data leakage is using novel but synthesis images, where JourneyDB~\citep{NEURIPS2023_9bc59aff} is the first work aiming to leverage synthesis images to evaluate current LVLMs. The prompts together with the corresponding images are downloaded from Midjourney~\citep{midjourney} and ChatGPT~\citep{ChatGPT} is leveraged to label the images. However, JourneyDB is a top-down framework where the number of images is fixed. Besides, the ChatGPT labeling may cause hallucinate annotations, leading to the unreliable evaluation results. In contrast, Dysca serves as a bottom-up framework which enables dynamic and scalable generation of both images and evaluation questions, while also supporting varied evaluation tasks. The rule-based question generation method also makes the annotations more accuracy. 

\vspace{-0.3cm}
\section{Dysca}
\vspace{-0.2cm}
\subsection{Overview of Our Pipeline}
\vspace{-0.1cm}
The overview of our pipeline is shown in Fig. \ref{fig:framework}, containing data generation, data cleaning and LVLMs evaluation.
For the data generation, our Dysca benchmark consists of four dimensions, i.e., $(M,P,I,Q)$, where $M$ means ``Metadata'', $P$ means ``Prompt'', $I$ means ``Image'' and $Q$ means ``Question-answer pair''. 
We further decouple the metadata $M$ into 4 parts, i.e., ``style'', ``attribute'', ``foreground'' and ``background'', and the combination of the four parts constitute the image prompts $P$. 
Then, given the prompt $P$ and the selected scenario, we leverage the Text-to-Image (T2I) diffusion model (e.g., SDXL~\citep{podell2023sdxl}) to synthesis image $I$ and add the specific perturbation to the image $I$. 
After that, since the prompt already includes the question angle and the corresponding answer, we construct a rule-based approach to generate the $Q$. Three types of questions are considered, i.e., multi-choice, true-or-false and free-form. Multi-choice and true-or-false questions utilize a closed-ended manner to assess LVLMs, while free-form questions employ an open-ended manner through image captioning for evaluation. For the data cleaning, considering that the T2I diffusion model may generate unsuccessful outcomes, we then use CLIP~\citep{radford2021learning_clip} and PP-OCRv3~\citep{li2022ppocrv3} to automatically clean the whole dataset to obtain the final Dysca. Finally, we evaluate 14 open-sourced LVLMs and 2 closed-source LVLMs on our proposed Dysca.

\begin{figure}[t!]
 \begin{minipage}{0.42\textwidth} 
\captionof{table}{Key statistics of Dysca.}
 \centering
 \fontsize{8.2pt}{\baselineskip}\selectfont 
 \renewcommand\tabcolsep{1.0pt} 
 \renewcommand\arraystretch{0.8} 
 \begin{tabular}{lc}
 \toprule
 \textbf{Statistic} & \textbf{\#Number} \\
 \midrule
  Total questions & 617K \\
  ~- Clean & 156K (25.2\%) \\
  ~- Print attacking & 149K (24.1\%) \\
  ~- Adversarial attacking & 156K (25.2\%) \\
  ~- Corruption & 156K (25.2\%) \\
 \midrule
 Question type \\
 ~- Multi-choices & 251K (40.6\%) \\
 ~- True-or-false & 250K (40.5\%) \\
 ~- Free-form & 116K (18.8\%)  \\
 \midrule
 Image resolution & 1024*1024 \\
 \midrule
 Unique number of images & 289K \\
 Unique number of questions & 162K \\
 Unique number of answers & 31K \\
 \midrule
 Average question length & 37.8 \\
 Average answer length & 2.7 \\
 Average choice number & 3.0 \\
 \bottomrule
 \end{tabular}

 \label{tab:statistics}
 \end{minipage} 
 \hfill
 \begin{minipage}{0.56\textwidth}
 \centering
 \vspace{-1mm}
\includegraphics[width=0.8\linewidth]{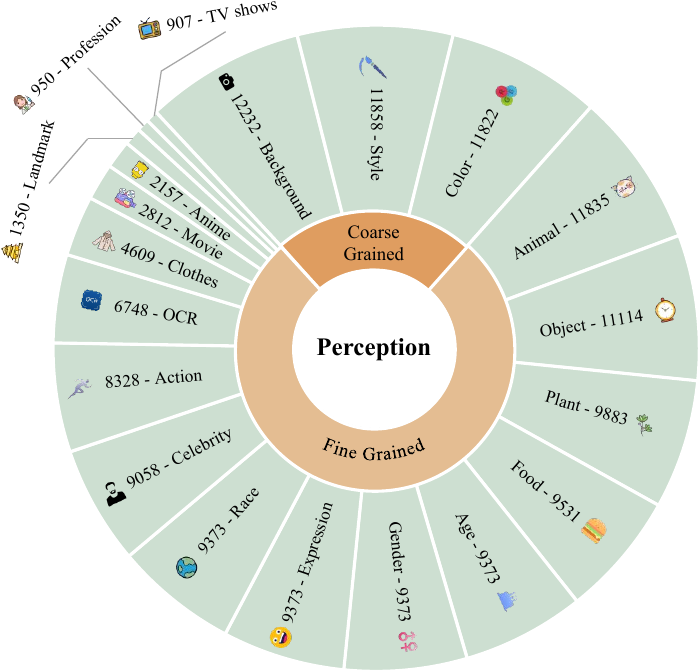}
\vspace{-1mm}
 \caption{Overview of the dataset distribution of 20 perceptual tasks. The number in each subtask shows the corresponding amount of their annotation.}
 \label{fig:subtasks}
 \end{minipage}
\vspace{-0.6cm}
\end{figure}

\vspace{-0.2cm}
\subsection{Perceptual Tasks}
\vspace{-0.2cm}
\textbf{Evaluation dimensions.} Perception is one of the most fundamental capabilities of LVLMs and previous works~\citep{fu2023mme} have shown that the lack of perceptual ability may result in hallucination~\citep{li-etal-2023-evaluating}. In order to comprehensively evaluate LVLMs' perception capability, we collect and organize existing sub-dimensions from current benchmarks, resulting in 20 assessment dimensions where we show all the subtasks and the corresponding amount of their annotation in the Fig. \ref{fig:subtasks}. We investigate on two types of perception dimensions, i.e., coarse-grained and fine-grained perception. Coarse-grained perception involves recognizing the style, background and color of images. Fine-grained perception involves recognizing the animal, object, plant, food, age, gender, expression, race, celebrity, action, text, clothes, movie, anime, landmark, profession and TV shows.

\textbf{Data sources.} For each perceptual subtask, we collect the textual data first to construct the metadata $M$. For the TV shows, Anime and Movie, we select the titles from the rating list of IMDb\footnote{https://www.imdb.com/} based on the number of user reviews. For the styles, we utilize the style lists collected from the community\footnote{https://stable-diffusion-art.com/sdxl-styles/} and remove those which have strong reflect on the image content like ``architectural style'' and ``Pokemon style''. Note that the style list does not include the style prompt associated with a particular artist's name. Besides, for the remaining contents, we select them from the label of current dataset (e.g., ImageNet~\citep{Russakovsky2014ImageNetLS}). All the selected textual data above constitute the metadata $M$. We provide the detailed information of the metadata in the Appendix \ref{Metadata}.
\vspace{-0.2cm}
\subsection{Construction of Questions \& Answers}
\begin{figure}
    \centering
    \includegraphics[width=0.95\linewidth]{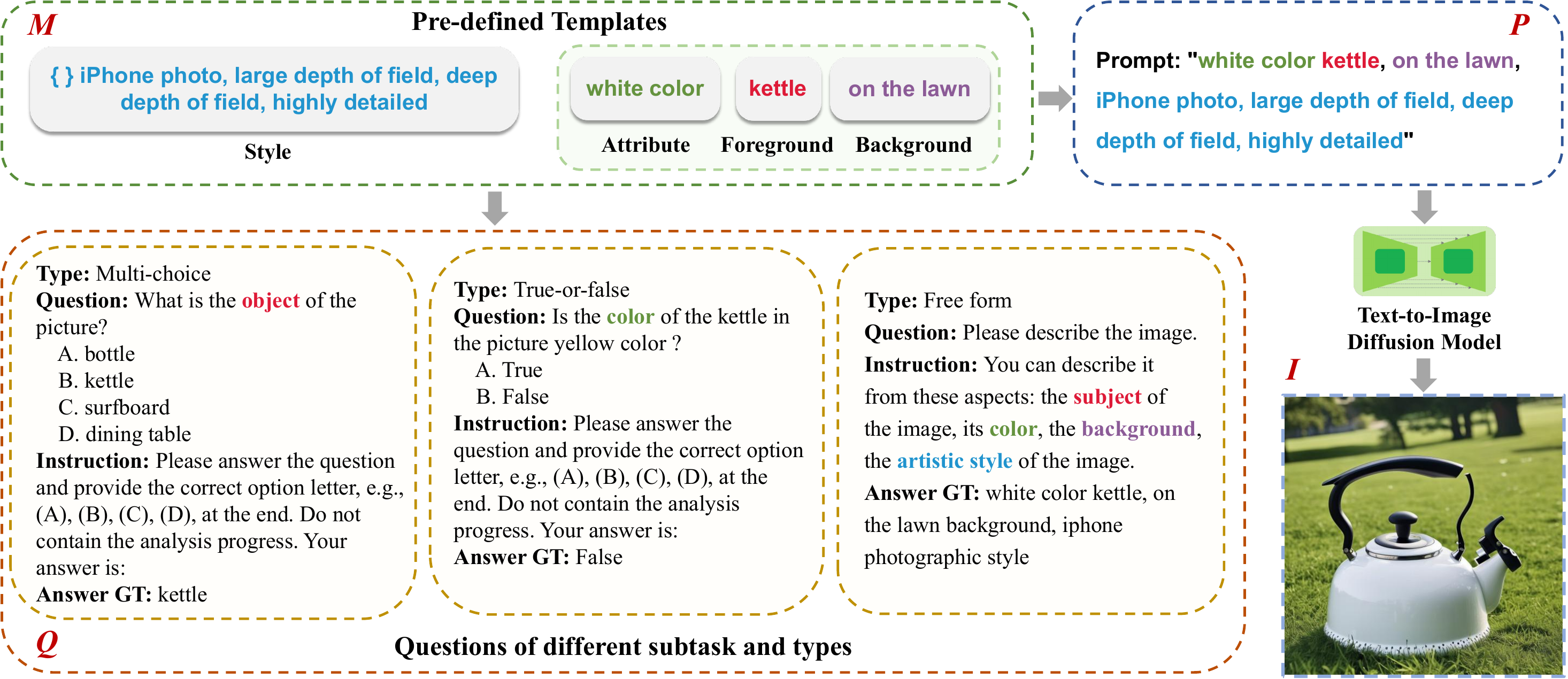}
    \caption{The process of generating the prompt (P), image (I) and QA pairs (Q) from metadata (M).}
    \label{fig:mpiq}
    \vspace{-0.6cm}
\end{figure}

\vspace{-0.2cm}
Recall that the data generation for Dysca benchmark consists of four dimensions, i.e., \((M,P,I,Q)\), denoting the metadata ($M$), prompt ($P$), image ($I$) and question-answer pairs ($Q$), respectively. The relationships between these parts and the process of constructing Dysca are shown in Fig. \ref{fig:mpiq}. The metadata M is the core of the whole Dysca, containing all the information for generating \(P\), \(I\) and \(Q\). The metadata \(M\) consists of foreground, attribute, background and style, which guides the generation of the prompt (\(P\)) through pre-designed templates. Then, we utilize the T2I diffusion model to generate the corresponding image using the prompt \(P\). For generating the image with a specific text on it for the OCR subtask, we leverage TextDiffusion2~\citep{chen2023textdiffuser}, which is the state-of-the-art text rendering method. For the rest of images, we leverage Stable Diffusion XL~\citep{podell2023sdxl}.  Subsequently, based on the different question types we select, i.e., multi-choices, true-or-false and free-form, we generate the corresponding VQA pairs in Dysca.

Besides, in order to evaluate the model performance under various scenarios, we conduct experiments on 4 scenarios, i.e., clean, corruption, print attacking and adversarial attacking.
For the print attacking, followed by~\citep{Cheng2024UnveilingTD}, we add the deceptive text on the image, where the text is a wrong option. 
Besides, to comprehensively evaluate the performance of LVLMs under corruption scenario, we add more typographic factors to original settings (i.e., different font orientations and font positions). 
For the adversarial attacking, we leverage PGD~\citep{Madry2017TowardsDL} to generate the adversarial image.
We use InstructBLIP~\citep{dai2023instructblip} as the proxy model and regard others as the black box models. 
The reason why we choose InstructBLIP is that it has shown superior performance in clean scenario. 
Besides, the black-box setting better reflects the robustness of the models when they face the real-world adversarial attacks. 
For the corruption, we leverage the image corruption methods collected from~\citep{zhang2024benchmarking}. 
We remove some hard corruptions as they significantly impact the quality of the image, leading to human failure in judging the style and content of the image. The detailed  examples are shown in Appendix \ref{Corrupted Scenarios}.

\textbf{Data Clean.} To ensure the quality of Dysca, four steps are adopt: 1) First, we manually remove difficult-to-generate foregrounds and attributes, along with backgrounds and styles that could heavily affect image content. We believe this process can serve as a coarse-grained method to eliminate samples that are highly likely to be generated incorrectly. 2) Then, we leverage the off-the-shelf models, i.e., PP-OCRv3~\citep{li2022ppocrv3} and CLIP-L-14~\citep{radford2021learning_clip}, to clean the data.
PP-OCRv3~\citep{li2022ppocrv3} is leveraged as the filter to exclude the failure image that TextDiffusion2~\citep{chen2023textdiffuser} generates the wrong text on the image. 
For the other images, we use CLIP-L-14~\citep{radford2021learning_clip} with a threshold of 0.75 to filter out the images with low text-image consistency. We find that using 0.75 as the threshold achieves a good balance between image correctness and data scale. 
3) After that,  We select the top six performing models and eliminate any question-answer pairs where the models either answer incorrectly or indicate that the answer was not included among the options. We observe that nearly 100\% of the samples filtered out by these models are incorrect. 4) Finally, we analyze the patterns in these incorrect samples, removing the associated vocabulary from our metadata and discarding all related samples. By meticulously refining the metadata manually and utilizing automated tools to assist in question filtering, Dysca ensures high-quality data synthesis.
In the end, we filter out nearly 40\% of low quality samples. The final statistics of our released Dysca are shown in Tab. \ref{tab:statistics}. Note that the OCR subtask does not involve print attacking scenario as misidentifying adversarial text does not indicate poor OCR robustness of the LVLMs. Therefore, there are 7K fewer questions in the print attacking scenario. Besides, for the free-form question type, since it allows to assess the model's perception abilities across multiple subtasks at the same time, we reduce the number of free-form questions for achieving a balanced data distribution. 

\vspace{-0.2cm}
\subsection{Evaluation Strategy}
\vspace{-0.2cm}
\textbf{Instruction Design.} We design two types of instructions to improve the instruction-following result of LVLMs. For the multi-choices and true-or-false questions, we design the questions followed by the description ``Please answer the question and provide the correct option letter, e.g., (A), (B), (C), (D), at the end. Do not contain the analysis progress. Your answer is: ''. For the free-form questions, recalling that the prompt $P$ contains four part, i.e., the style, attribute, foreground and background, we instruct the model to caption these four dimensions by ``Please describe the image. You can describe it from these aspects: $\{\}$'', where ``$\{\}$'' includes the specific template we design for each part. We display the sample in the Fig. \ref{fig:mpiq} and more examples can be found in the Appendix \ref{Examples of Dysca}.

\begin{figure}
    \centering
    \includegraphics[width=0.9\linewidth]{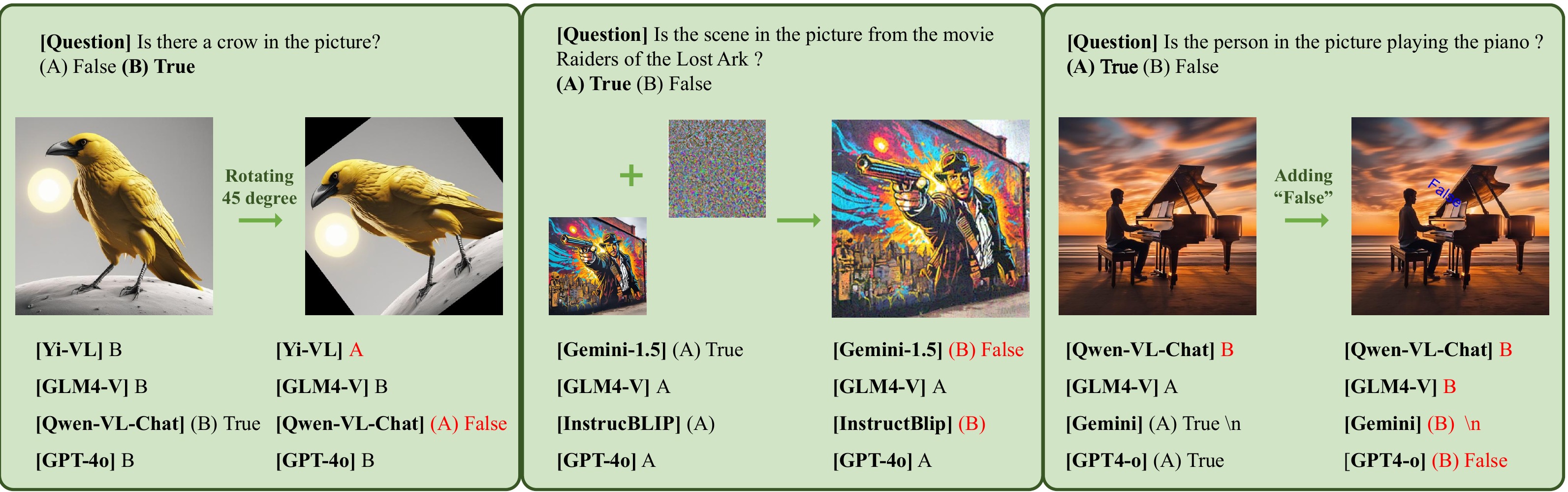}
    \caption{The failure cases for the noisy scenarios. From left to right are: corruption scenario, adversarial attacking scenario, and print attacking scenario.}
    \label{fig:other_results}
    \vspace{-0.4cm}
\end{figure}

\begin{figure}
    \centering
    \includegraphics[width=0.9\linewidth]{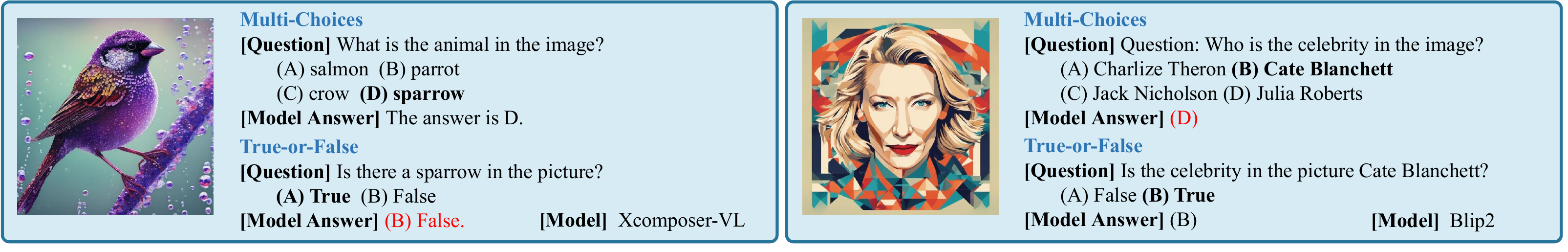}
    \caption{Models exhibit different performance when facing the same image but different question types.}
    \label{fig:failure_case}
    \vspace{-0.4cm}
\end{figure}

\textbf{Evaluation Metrics.} For the multi-choices and true-or-false questions, we use accuracy as the evaluation metric. We randomly shuffle the order of choices to prevent evaluation results from being influenced by the model's tendency towards specific choices~\citep{zong2023fool}. The random accuracy of the two types are equal to 25$\%$ and 50$\%$, respectively. We use regular expressions to extract the model's answer choices. For cases where the extraction is fail, we calculate the Levenshtein distance between the answer string and the choice string, and select the option with the minimum distance as the model's answer. For the free-form questions, we test the model's image caption capability where the ground truth is the prompt of the image. Followed by~\citep{xu2023lvlmehub}, we use SentenceTransformer~\citep{thakur-etal-2021-augmented} to compute the text similarity with prompt $P$ and the caption output of the LVLMs. The final score of each question type is the average score of subtasks.

Besides, thanks to generative evaluation framework of Dysca, we are able to effectively control variables and conduct a detailed analysis of the model's fine-grained capabilities. Specifically, we introduce two novel metrics to measure the sensitivity of LVLMs on question types and covariate shift (i.e., image style). 
The sensitivity to question type aims to evaluate whether LVLMs exhibit inconsistent performance when facing different types of question types (i.e., multi-choice vs. true-or-false). 
We first normalize the score of true-or-false by $TFSQ = \frac{S-50}{100-50}*100\%$ and multi-choices by $MCSQ = \frac{S-25}{100-25} * 100\%$, where $S$ denotes the score of the models in each question type. The sensitivity of LVLMs to question types is defined by:
\begin{equation}
   SQ = \frac{(TFSQ-SQ_{Avg})^2+(MCSQ-SQ_{Avg})^2}{2}, 
\end{equation}
where $SQ_{Avg} = \frac{TFSQ+MCSQ}{2}$. 

The sensitivity to covariate shift aims to evaluate whether LVLMs exhibit inconsistent performance when facing the same content and question format, but with variations in image covariates. It is defined by:
\begin{equation}
    SC = \frac{\sum_{i=1}^{N} (S_i-SC_{Avg})^2}{N}, 
\end{equation}
where $SC_{Avg}=\frac{\sum_{i=1}^{N}S_i}{N}$ and $S_i$ denotes the score of the models in each style. Since there are 51 image styles in Dysca, we set $N=51$.

\section{Results and Analysis}
\vspace{-0.3cm}
In this section, we report the evaluation results and make insightful analysis. A total of 26 LVLMs are evaluated on Dysca, including BLIP2~\citep{li2023blip2}, InstructBLIP~\citep{dai2023instructblip}, LLavA~\citep{liu2023llava_1-5}, MiniGPT-4~\citep{zhu2023minigpt4}, Otter~\citep{li2023otter}, XComposer-VL~\citep{internlmxcomposer}, Qwen-VL-chat~\citep{bai2023qwen-vl}, Shikra~\citep{chen2023shikra}, Emu2-Chat~\citep{sun2024generativemultimodalmodelsincontext}, GLM-4V~\citep{glm2024chatglm}, MiniCPM-v2.5~\citep{yao2024minicpm}, Yi-VL~\citep{ai2024yi}, mPLUG-Owl-2~\citep{ye2023mplugowl2revolutionizingmultimodallarge}, Phi-3-Vision~\citep{abdin2024phi3technicalreporthighly}, GPT-4o~\citep{openai2023gpt4}, Gemini-1.5-pro~\citep{geminiteam2024geminifamilyhighlycapable}. Each model is evaluated with all the 20 perception subtasks under 4 scenarios. The detailed rankings for each subtask can be found in the Appendix \ref{leaderboard}. 
\subsection{Main Results}
\vspace{-0.1cm}
\textbf{Blind Setting.} We first evaluate LVLMs when only textual questions are provided. As shown in the ``Blind'' column of Tab. \ref{tab:Results_final}, all LVLMs yield consistent results on the Dysca and perform comparable to random guessing. This outcome demonstrates that the generated paradigm employed by Dysca effectively mitigates the potential impact of data leakage ~\citep{chen2024right}, thereby enhancing the fairness of the evaluation results. Additional comparisons can be found in Appendix \ref{blind}.

\textbf{Clean Scenario.} The evaluation results of various LVLMs in different perceptual subtasks under clean scenarios are presented in the ``clean'' column of Tab. \ref{tab:Results_final}. We calculate the average score of 3 question types. As can be seen, GLM-4V~\citep{glm2024chatglm} outperforms other LVLMs, achieving top-1 performance. MiniCPM-v2.5~\citep{yao2024minicpm}, Xcomposer2-VL~\citep{internlmxcomposer2} and Gemini-1.5-pro~\citep{geminiteam2024geminifamilyhighlycapable} also perform well. It is evident that for the latest large models, their scores remain below 85. The results highlight that all existing LVLMs still struggle to provide accurate responses to questions formulated by Dysca. 

\textbf{Noisy Scenarios.} The evaluation results of various LVLMs under noisy scenarios (i.e., corruption, print attacking and adversarial attacking) are presented in last 3 columns in Tab. \ref{tab:Results_final}. The value in the brackets shows the relative values with respect to the ones in the clean scenario. As can be seen, GLM-4V~\citep{glm2024chatglm} still takes a lead on corruption and adversarial attacking scenarios. For the print attacking scenario, mPLUG-Owl-2~\citep{ye2023mplugowl2revolutionizingmultimodallarge} performs the best. Here, we present a failure case sample for each of the three different scenarios on Fig. \ref{fig:other_results}.

\begin{table}[t]
\centering
\caption{Evaluation results of 26 LVLMs, where the darker colors represent better performance. The top 1 result on each column are \textbf{bolded} and the value in brackets is the relative values with respect to the ones in the clean scenario. ``PrintAtt'' and ``AdverAtt'' means ``Print Attacking'' and ``Adversarial Attacking'', respectively. ``*'': the model is under white-box setting.}
\label{tab:Results_final}
\scalebox{0.60}{
\begin{tabular}{cccccccc|cc}
\hline
 &  &  & & \multicolumn{4}{c|}{\textbf{Scenarios}} \\ \cline{5-8} 
\multirow{-2}{*}{\textbf{Model}} & \multirow{-2}{*}{\textbf{LLM}} & \multirow{-2}{*}{\textbf{Visual Encoder}}  & \multirow{-2}{*}{\textbf{Blind }} & \textbf{Clean $\uparrow$} & \textbf{Corruption $\uparrow$} & \textbf{AdverAtt $\uparrow$} & \textbf{PrintAtt $\uparrow$} &  \multirow{-2}{*}{\textbf{SQ $\downarrow$}}   &  \multirow{-2}{*}{\textbf{SC $\downarrow$}}    \\ \hline
MiniGPT-4                        & Vicuna-7B                      & EVA-CLIP ViT-G                            & \cellcolor[HTML]{F3F9F7}35.37    & \cellcolor[HTML]{E1F1E7}41.38 & \cellcolor[HTML]{D9EEE1}42.30 (+0.92)  & \cellcolor[HTML]{F0F8F5}34.42 (-6.96)   & \cellcolor[HTML]{D8EEE0}42.71 (+1.33)  & \cellcolor[HTML]{6DC284}193.6  & \cellcolor[HTML]{74C58A}1.7   \\
MiniGPT-4                        & Vicuna-13B                     & EVA-CLIP ViT-G                            & \cellcolor[HTML]{F4F9F8}35.21    & \cellcolor[HTML]{C6E6D0}50.17 & \cellcolor[HTML]{C3E5CE}49.63 (-0.54) & \cellcolor[HTML]{F8FBFC}31.77 (-18.40)  & \cellcolor[HTML]{CAE8D4}47.55 (-2.62)  & \cellcolor[HTML]{90D0A2}758.0    & \cellcolor[HTML]{84CB98}2.5   \\
MiniGPT-4                        & LLaMA2-7B                         & EVA-CLIP ViT-G                            & \cellcolor[HTML]{F5F9F9}34.77    & \cellcolor[HTML]{B2DEBF}56.61 & \cellcolor[HTML]{B1DEBF}55.70 (-0.91)  & \cellcolor[HTML]{F3F9F7}33.55 (-23.06)  & \cellcolor[HTML]{C3E5CE}49.78 (-6.83)  & \cellcolor[HTML]{B4DFC1}1344.5 & \cellcolor[HTML]{8ECFA0}3.0     \\
MiniGPT-2                        & LLaMA2-7B                         & EVA-CLIP ViT-G                            & \cellcolor[HTML]{F3F9F8}35.28    & \cellcolor[HTML]{ACDCBA}58.46 & \cellcolor[HTML]{AADBB9}58.06 (-0.40) & \cellcolor[HTML]{AFDDBC}56.62 (-1.84)   & \cellcolor[HTML]{BAE1C6}52.96 (-5.50)  & \cellcolor[HTML]{BFE3CA}1512.1 & \cellcolor[HTML]{A8DAB7}4.3   \\
BLIP2                            & Flan-T5-XL                     & EVA-CLIP ViT-G                            & \cellcolor[HTML]{F3F9F7}35.35    & \cellcolor[HTML]{97D3A8}65.30  & \cellcolor[HTML]{93D2A4}66.09 (+0.79) & \cellcolor[HTML]{F6FAFA}32.55 (-32.75)  & \cellcolor[HTML]{AEDDBB}57.01 (-8.29)  & \cellcolor[HTML]{6BC182}165.5  & \cellcolor[HTML]{FCFCFF}8.5   \\
BLIP2                            & OPT-3B                         & EVA-CLIP ViT-G                            & \cellcolor[HTML]{F4F9F8}34.99    & \cellcolor[HTML]{E6F4EC}39.54 & \cellcolor[HTML]{DFF1E6}40.29 (+0.75) & \cellcolor[HTML]{FBFCFF}30.62 (-8.92)   & \cellcolor[HTML]{E8F4EE}37.26 (-2.28)  & \cellcolor[HTML]{68C07F}110.7  & \cellcolor[HTML]{80CA94}2.3   \\
BLIP2                            & OPT-7B                         & EVA-CLIP ViT-G                            & \cellcolor[HTML]{F4F9F8}35.21    & \cellcolor[HTML]{E6F4EC}39.55 & \cellcolor[HTML]{DCF0E4}41.12 (+1.57) & \cellcolor[HTML]{F8FBFC}31.76 (-7.79)   & \cellcolor[HTML]{E3F2EA}38.82 (-0.73)  & \cellcolor[HTML]{64BE7C}50.2   & \cellcolor[HTML]{76C68C}1.8   \\
InstructBLIP                     & Vicuna-7B                      & EVA-CLIP ViT-G                            & \cellcolor[HTML]{F4F9F8}35.14    & \cellcolor[HTML]{90D1A2}67.54 & \cellcolor[HTML]{90D1A2}67.01 (-0.53) & \cellcolor[HTML]{F0F8F5}34.42 (-33.12)  & \cellcolor[HTML]{BBE2C7}52.58 (-14.96) & \cellcolor[HTML]{B1DDBE}1287.1 & \cellcolor[HTML]{8ACE9D}2.8   \\
InstructBLIP                     & Vicuna-13B                     & EVA-CLIP ViT-G                            & \cellcolor[HTML]{F6FAFA}34.37    & \cellcolor[HTML]{98D4A9}64.89 & \cellcolor[HTML]{97D3A8}64.68 (-0.21) & \cellcolor[HTML]{F8FBFC}31.77 (-33.12)  & \cellcolor[HTML]{B8E1C4}53.53 (-11.36) & \cellcolor[HTML]{AFDCBC}1252.4 & \cellcolor[HTML]{84CB98}2.5   \\
InstructBLIP                     & Flan-T5-XL                     & EVA-CLIP ViT-G                            & \cellcolor[HTML]{F6FAFA}34.51    & \cellcolor[HTML]{93D2A5}66.54 & \cellcolor[HTML]{8ED0A0}67.58 (+1.04) & \cellcolor[HTML]{F5F9F9}32.95 (-33.59)* & \cellcolor[HTML]{BCE2C8}52.09 (-14.45) & \cellcolor[HTML]{72C488}271.3  & \cellcolor[HTML]{D0EAD9}6.3   \\
InstructBLIP                     & Flan-T5-XXL                    & EVA-CLIP ViT-G                            & \cellcolor[HTML]{F5F9F9}34.82    & \cellcolor[HTML]{8DCF9F}68.65 & \cellcolor[HTML]{88CD9B}69.79 (+1.14) & \cellcolor[HTML]{F5F9F9}32.95 (-35.70)  & \cellcolor[HTML]{ABDCBA}57.73 (-10.92) & \cellcolor[HTML]{6EC285}215.0    & \cellcolor[HTML]{DAEEE1}6.8   \\
LLava-1.5                        & Vicuna-7B                      & CLIP ViT-L                                & \cellcolor[HTML]{F5FAF9}34.63    & \cellcolor[HTML]{C2E5CD}51.27 & \cellcolor[HTML]{BDE3C9}51.70 (+0.43)  & \cellcolor[HTML]{C3E5CE}49.62 (-1.65)   & \cellcolor[HTML]{CAE8D4}47.27 (-4.00)  & \cellcolor[HTML]{92D1A3}788.1  & \cellcolor[HTML]{92D1A4}3.2   \\
LLava-1.5                        & Vicuna-13B                     & CLIP ViT-L                                & \cellcolor[HTML]{F4F9F8}35.21    & \cellcolor[HTML]{AADBB8}59.23 & \cellcolor[HTML]{A6D9B5}59.58 (+0.35) & \cellcolor[HTML]{AEDDBC}56.87 (-2.36)   & \cellcolor[HTML]{BDE3C9}51.69 (-7.54)  & \cellcolor[HTML]{9AD4AA}912.5  & \cellcolor[HTML]{ECF5F1}7.7   \\
Otter                            & LLaMA-7B                       & CLIP ViT-L                                & \cellcolor[HTML]{F4F9F8}35.19    & \cellcolor[HTML]{B7E0C4}54.90  & \cellcolor[HTML]{B0DEBE}56.02 (+1.12) & \cellcolor[HTML]{BEE3CA}51.42 (-3.48)   & \cellcolor[HTML]{E6F4EC}37.78 (-17.12) & \cellcolor[HTML]{7CC890}427.4  & \cellcolor[HTML]{CAE7D4}6.0     \\
Shikra                           & LLaMA-7B                       & CLIP ViT-L                                & \cellcolor[HTML]{F4F9F9}34.96    & \cellcolor[HTML]{A1D7B0}62.24 & \cellcolor[HTML]{9CD5AC}63.06 (+0.82) & \cellcolor[HTML]{A8DAB7}58.78 (-3.46)   & \cellcolor[HTML]{C4E5CE}49.56 (-12.68) & \cellcolor[HTML]{7CC891}440.6  & \cellcolor[HTML]{B6DFC3}5.0     \\
Xcomposer-VL                     & InternLM-7B                    & EVA-CLIP ViT-G                            & \cellcolor[HTML]{FCFCFF}32.33    & \cellcolor[HTML]{84CC98}71.40  & \cellcolor[HTML]{81CA95}72.08 (+0.68) & \cellcolor[HTML]{FCFCFF}30.28 (-41.12)  & \cellcolor[HTML]{97D3A8}64.71 (-6.69)  & \cellcolor[HTML]{6AC181}147.9  & \cellcolor[HTML]{CCE8D5}6.1   \\
Xcomposer2-VL                    & InternLM2-7B                   & CLIP ViT-L                                & \cellcolor[HTML]{FBFCFE}32.76    & \cellcolor[HTML]{6DC283}79.13 & \cellcolor[HTML]{6EC384}78.64 (-0.49) & \cellcolor[HTML]{74C589}76.60 (-2.53)    & \cellcolor[HTML]{92D1A4}66.34 (-12.79) & \cellcolor[HTML]{63BE7B}\textbf{20.9}   & \cellcolor[HTML]{78C68D}1.9   \\
Qwen-VL-Chat                     & Qwen-7B                        & OpenClip ViT-bigG                         & \cellcolor[HTML]{FAFCFE}33.06    & \cellcolor[HTML]{A1D7B0}62.18 & \cellcolor[HTML]{A2D8B1}61.05 (-1.13) & \cellcolor[HTML]{A5D9B4}59.85 (-2.33)   & \cellcolor[HTML]{BDE3C8}51.94 (-10.24) & \cellcolor[HTML]{D6ECDE}1885.4 & \cellcolor[HTML]{D8EDE0}6.7   \\
Emu2-Chat                        & LLaMA-33B                      & EVA2-CLIP-E                               & \cellcolor[HTML]{F4F9F8}35.14    & \cellcolor[HTML]{9CD5AC}63.64 & \cellcolor[HTML]{9CD6AD}62.81 (-0.83) & \cellcolor[HTML]{9FD7AF}61.90 (-1.74)    & \cellcolor[HTML]{B4DFC1}54.82 (-8.82)  & \cellcolor[HTML]{FCFCFF}2497.9 & \cellcolor[HTML]{72C488}1.6   \\
GLM-4V                           & GLM-4-9B-Chat                  & EVA2-CLIP-E                               & \cellcolor[HTML]{F4F9F8}35.08    & \cellcolor[HTML]{63BE7B}\textbf{82.09} & \cellcolor[HTML]{64BF7C}\textbf{81.95} (-0.14) & \cellcolor[HTML]{68C07F}\textbf{80.72} (-1.37)   & \cellcolor[HTML]{BCE2C8}52.09 (-30.00) & \cellcolor[HTML]{63BE7B}25.5   & \cellcolor[HTML]{63BE7B}\textbf{0.8}   \\
MiniCPM-V2.5                     & Llama3-Instruct 8B             & SigLIP SoViT-400m                         & \cellcolor[HTML]{F4F9F8}34.99    & \cellcolor[HTML]{6EC384}78.75 & \cellcolor[HTML]{71C487}77.41 (-1.34) & \cellcolor[HTML]{77C68C}75.44 (-3.31)   & \cellcolor[HTML]{A2D8B2}60.77 (-17.98) & \cellcolor[HTML]{64BE7B}38.3   & \cellcolor[HTML]{74C58A}1.7   \\
Yi-VL                            & Yi-6B-Chat                     & OpenClip ViT-H                            & \cellcolor[HTML]{F4F9F8}35.01    & \cellcolor[HTML]{77C68C}75.71 & \cellcolor[HTML]{79C78E}74.94 (-0.77) & \cellcolor[HTML]{80CA94}72.53 (-3.18)   & \cellcolor[HTML]{96D3A7}64.97 (-10.74) & \cellcolor[HTML]{70C386}233.1  & \cellcolor[HTML]{78C68D}1.9   \\
mPLUG-Owl-2                      & LLaMA2-7B                       & CLIP ViT-L                                & \cellcolor[HTML]{F4F9F8}35.03    & \cellcolor[HTML]{7CC891}74.09 & \cellcolor[HTML]{7FCA93}72.85 (-1.24) & \cellcolor[HTML]{88CD9B}69.76 (-4.33)   & \cellcolor[HTML]{7FCA93}\textbf{72.85} (-1.24)  & \cellcolor[HTML]{6CC283}180.9  & \cellcolor[HTML]{7EC993}2.2   \\
Phi-3-Vision                     & Phi-3                          & CLIP ViT-L                                & \cellcolor[HTML]{F5F9F9}34.74    & \cellcolor[HTML]{7FCA93}73.23 & \cellcolor[HTML]{81CA95}72.11 (-1.12) & \cellcolor[HTML]{88CD9B}69.66 (-3.57)   & \cellcolor[HTML]{ABDCB9}57.78 (-15.45) & \cellcolor[HTML]{73C489}292.7  & \cellcolor[HTML]{70C386}1.5   \\
GPT-4o                           & /                              & /                                         & \cellcolor[HTML]{F4F9F8}35.02    & \cellcolor[HTML]{77C68C}75.69 & \cellcolor[HTML]{77C68C}75.52 (-0.17) & \cellcolor[HTML]{7DC991}73.47 (-2.22)   & \cellcolor[HTML]{B0DDBD}56.34 (-19.35) & \cellcolor[HTML]{65BF7D}67.4   & \cellcolor[HTML]{78C68D}1.9   \\
Gemini-1.5-Pro                   & /                              & /                                         & \cellcolor[HTML]{F6FAFA}34.55    & \cellcolor[HTML]{71C487}77.79 & \cellcolor[HTML]{72C488}77.12 (-0.67) & \cellcolor[HTML]{76C68B}75.89 (-1.90)   & \cellcolor[HTML]{A2D8B1}61.05 (-16.74) & \cellcolor[HTML]{7CC891}439.6  & \cellcolor[HTML]{72C488}1.6  
\\ \hline
\end{tabular}
}
\vspace{-0.2cm}
\end{table}

\subsection{Key Observations}
\noindent\textbf{(1) For LVLMs, the capacity of the language model plays a crucial role.} When using the same visual encoder, models that utilize a language model with a larger parameter sizes (e.g., MiniGPT-4 achieves an improvement score of 8.79 in the clean scenario when using Vicuna-13B compared to Vicuna-7B) or with a stronger capability (e.g., GLM-4V that uses the GLM-4-9B-Chat language model shows an improvement score of 18.45 in the clean scenario compared to Emu2, which uses the LLaMA-33B language model) tend to achieve a better performance.

\noindent \textbf{(2) Each model shows robustness in the corruption scenario, but experiences significant degradation in both attack scenarios.} In the image corruption scenario, all models demonstrate minimal score variations, i.e., less than 1\%. However, under print attacks, performance drops are notable. For instance, two closed-source models exhibit significant performance degradation: Gemini-1.5-pro declines by 21.5\%, resulting in a score reduction from 77.79 to 61.05, while GPT-4o shows a 25.8\% decrease, dropping its score from 75.69 to 56.10. Among the leading open-source models, GLM-4V experiences a sharp 36.5\% performance drop, lowering its score from 82.09 to 52.09, and Phi-3-Vision records a 21.39\% decline, reducing its score from 73.23 to 57.78. Notably, mPLUG-Owl-2 demonstrates the highest robustness, with only a 1.7\% reduction. The XComposer-VL series also exhibits strong resilience against print-based attacks. In the adversarial attack scenario, where the attack algorithm directly targets the image encoder, LVLMs employing a shared encoder architecture (e.g., Blip2, InstructBLIP, and XComposer-VL, all of which utilize EVA-CLIP~\citep{Fang2022EVAET} as their image encoder) suffer substantial performance declines, with some models performing even worse than random chance. For example, XComposer-VL experiences a 57.6\% drop, reducing its score from 71.40 to 30.28. Models with different image encoders also experience performance degradation ranging from 1\% to 5\%, showing a higher impact than that of corruption noise. More detailed results are available in Appendix \ref{Corrupted Scenarios}.


\noindent \textbf{(3) Models exhibit varying sensitivity to different question types and covariate shifts.} For the sensitivity to question types, we present two examples in Fig. \ref{fig:failure_case}. As can be seen, the XComposer-VL~\citep{internlmxcomposer} recognizes the sparrow in the image under a multiple-choice setting but fails to identify the sparrow in the same image under a true-or-false setting. The quantity results are shown in Tab. \ref{tab:Results_final}. Xcomposer2-VL achieves the best result with a score of 20.9. However, we observe that the perception ability of LVLM does not show a positive correlation with sensitivity to question types. For instance, while the GLM-4V achieves the highest performance in evaluation tasks, it exhibits higher sensitivity to question types than Xcomposer2-VL. One of the factors influencing the sensitivity to question types may be the inherent biases of the language model. Using the same base language model may result in similar outcomes for this metric. it is also noted that Gemini-1.5-pro performs bad in this metric, revealing its preference for certain question types. For the sensitivity to covariate shifts, as shown in the last column of Tab. \ref{tab:Results_final}, GLM-4V achieves the best result with a score of 0.8. However, we also observe that the perception ability of LVLM does not show a positive correlation with covariate shifts. For example, InstructBLIP-Flan-T5-XXL outperforms InstructBLIP-Flan-T5-XL in terms of performance but shows higher sensitivity to covariate shifts. 
\vspace{-0.3cm}
\subsection{Analysis on inter-task and intra-task}
\vspace{-0.2cm}

\textbf{Analysis on Inter-task.} In order to investigate the inter-relationships across evaluation dimensions, we conduct hierarchical clustering based on the Euclidean distance of the scores across 20 dimensions for 26 models. We observe that models show varied consistency of performance across the dimensions. LVLMs tend to perform better in dimensions that involve well-defined image perception, such as landmark recognition and object recognition. However, dimensions likes style recognition and movie recognition exhibit greater challenges for LVLMs, which may be attributed to the limited training resources in these specific domains.
Besides, commercial models exhibit poor performance in tasks related to ``people'' (i.e., ``race'', ``age'', and ``gender''). This is likely due to the additional safety training incorporated into closed-source models, which results in more conservative responses on related questions.
Detailed results are provided in Appendix \ref{leaderboard}.

\textbf{Analysis on Intra-task.}
Thanks to the diverse metadata design, Dysca enables highly granular analysis within a single task. Taking animal categories as an example, we analyze the performance of LVLMs on different animals. Here, we perform the same hierarchical clustering methods across 51 animal categories. LVLMs performance vary significantly across these categories. We find that models tend to perform poorly when applied to marine life. This could be caused by the challenges in collecting ocean-related data. This observation highlights the need to direct the model's focus to oceanic domains.

\subsection{The Validity of Dysca}
\begin{wraptable}{r}{7cm}
    \vspace{-0.4cm}
	\centering
    \caption{The correlation results on three benchmarks, where $\rho \in [-1,1]$ and $\tau \in [-1,1]$.}
    \scalebox{0.7}
    {
    \label{tab:correlation}
    \begin{tabular}{ccccc}
    \hline
    \textbf{Style} & \textbf{Method} & \textbf{MMBench} & \textbf{OCRBench} & \textbf{SeedBench-2 }\\ \hline
    \multirow{2}{*}{All} & $\rho$ & 0.70 & 0.90 & 0.46 \\
     & $\tau$ & 0.60 & 0.80 & 0.43 \\ \hline
    \multirow{2}{*}{Realistic} & $\rho$ & 0.70 & 1.00 & 0.64 \\
     & $\tau$ & 0.60 & 1.00 & 0.62 \\ \hline
    \end{tabular}
    }
\end{wraptable}

In this section, we investigate on the evaluation gap between Dysca and non-synthesis benchmarks. We calculate the Spearman's rank correlation coefficient~\citep{ca468a70-0be4-389a-b0b9-5dd1ff52b33f} $\rho$  and the Kendall rank correlation coefficient~\citep{kendalltau} $\tau$  between the evaluation ranking of Dysca under clean scenario with the non-synthesis benchmark's evaluation ranking, i.e., MMBench~\citep{liu2023mmbench}, OCRBench~\citep{liu2024hidden} and SeedBench-2~\citep{li2023seedbench2}. Both coefficient generate a score in the range of [-1,1], where 1 represents a perfect positive correlation, -1 represents a perfect negative correlation, and 0 represents no correlation. These coefficients are typical tools for measuring the correlation between variables in statistics. When the absolute value of either coefficient exceeds 0.6, it is considered to indicate a significant correlation ~\citep{AKOGLU201891}. Specifically, we intersect our Dysca with current benchmarks based on the perceptual subtasks, evaluation models and evaluation question types. We then calculate the correlation of model evaluation rankings within this intersection. The results are shown in the first row of Tab. \ref{tab:correlation}. For the MMbench and OCRBench, both metrics show the high correlation, with $\rho$ and $\tau$ higher than 0.6. However, the correlation for SeedBench-2 is not as strong. Considering that SeedBench-2 only contains realistic images, we conduct additional experiments using the evaluation ranks on our realistic style images only. As shown in the second row of Tab. \ref{tab:correlation}, the correlation results of SeedBench-2 significantly improve (i.e., 0.46 vs. 0.64 for $\rho$ and 0.43 vs. 0.62 for $\tau$). The correlation with OCRBench also improves to 1, demonstrating the validity of using synthetic datasets for evaluation LVLMs.


\begin{wrapfigure}{r}{0.33\textwidth}
\vspace{-0.4cm}
        \includegraphics[width=0.2\textheight]{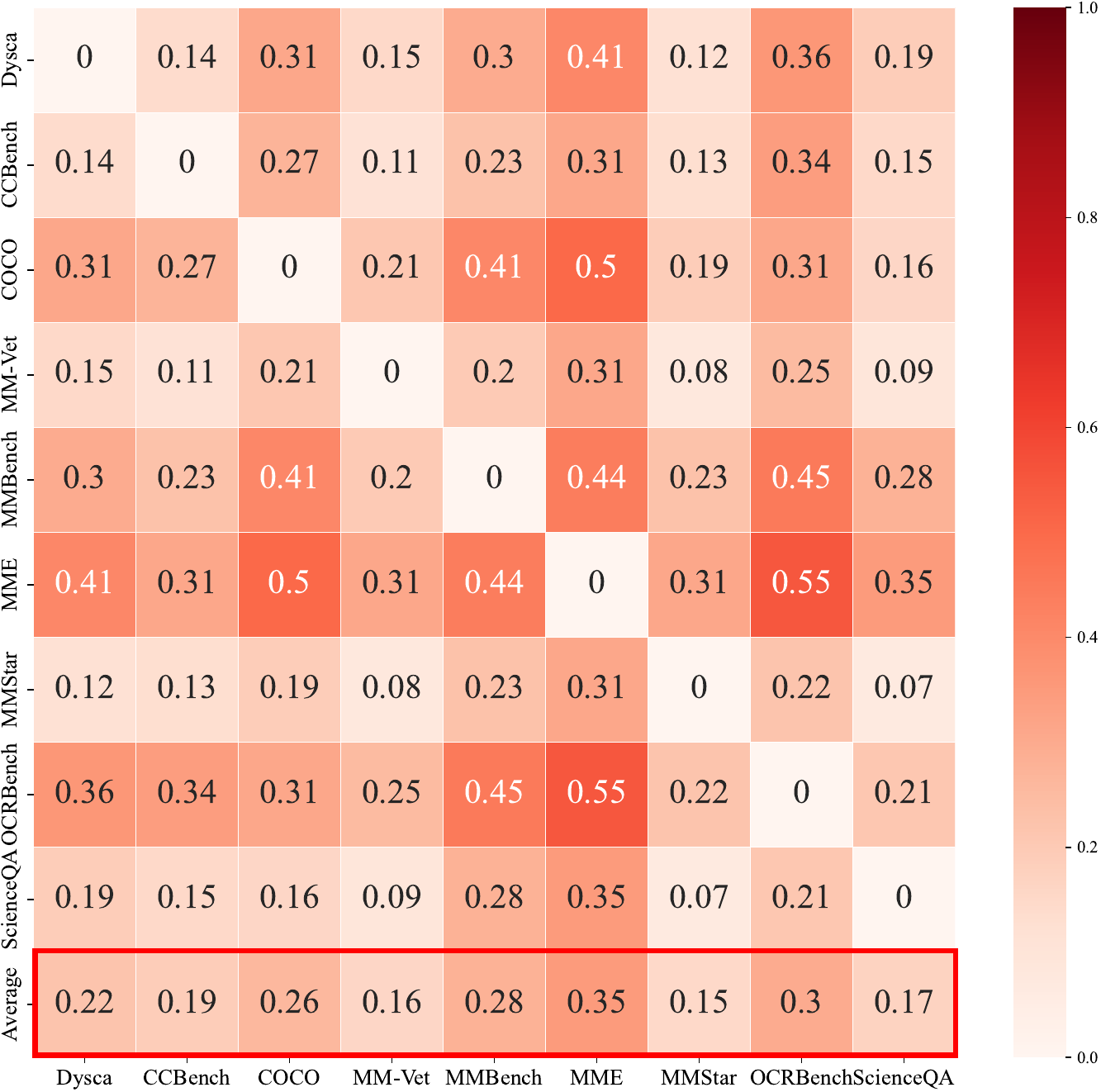}
        \caption{The KMMD distance between each benchmarks, with darker colors indicating larger distances.}
        \label{fig:KMMD}
\end{wrapfigure}

Besides, we calculate the data distribution distance between each benchmark to prove the low distance distribution between Dysca and non-synthesis benchmarks. We select CCBench~\citep{liu2023mmbench}, COCO-Val ~\citep{mscoco}, MMVet~\citep{yu2023mmvet}, MMBench~\citep{liu2023mmbench}, MME~\citep{fu2023mme}, MMStar~\citep{chen2024right}, OCRBench~\citep{liu2024hidden} and ScienceQA~\citep{lu2022learn_scienceQA}. The reason why we choose these  benchmark is that they have been widely used in evaluating LVLMs. We use Kernel Maximum Mean Discrepancy (KMMD)~\citep{6287330}  to measure the distribution distance. Specifically, we randomly sample 3,000 images from each benchmark (if the scale of the benchmark less than 3000, we use all that data) and utilize CLIP ~\citep{radford2021learning_clip} to encode these images. Then, we calculate the KMMD value using an RBF kernel between each pair. The results are shown in Fig. \ref{fig:KMMD}. Each row represents the value of KMMD between two benchmarks. The value of last row denotes the average value of KMMD. As can be seen, the distribution distance between Dysca and real-image benchmarks ranks in the middle compared to all other benchmarks, indicating that the evaluation results can effectively reflect the model's performance in real-world scenarios.


\section{Conclusion}
In this paper, we purpose Dysca, a dynamic and scalable benchmark for evaluating perception ability of Large Vision Language Models (LVLMs).  Dysca consists of 617K Vision-language QA pairs, covering 20 perceptual subtasks, 4 image scenarios and 3 question types. We conduct the experiment on 24 advanced open-source LVLMs and 2 closed-source LVLMs, revealing the insightful weakness of current LVLMs when facing different question types, image styles and image conditions. Experiments demonstrate the validity on evaluating LVLMs by using synthesis images.
\section*{Acknowledgement}
This work is partially supported by Strategic Priority Research Program of the Chinese Academy of Sciences (No. XDB0680202),
Beijing Nova Program (20230484368), Suzhou Frontier
Technology Research Project (No. SYG202325),
and Youth Innovation Promotion Association CAS.

\bibliography{iclr2025_conference}
\bibliographystyle{iclr2025_conference}

\clearpage

\appendix
\section*{Appendix}

\section{The Leaderboards} \label{leaderboard}

The model performance results for each subtask under clean scenario are shown in Tab. \ref{tab:leaderboard_1}, Tab. \ref{tab:leaderboard_2}, Tab. \ref{tab:leaderboard_3}, and Tab. \ref{tab:leaderboard_4}. Since the free-form question allows to assess the model's perception abilities across multiple subtasks at the same time, the  results of free-form are not taken into account. 

\begin{table}[h]
\centering
\caption{Evaluation results on the 5 perceptual subtasks. The top two results on each subtask are \textbf{bolded} and \underline{underlined}, respectively. ``MC'' and ``TF'' indicate the accuracy (\%) of ``Multi-choices'' and ``True-or-false'', respectively.}
\label{tab:leaderboard_1}
\scalebox{0.64}{
\begin{tabular}{cccllllllllll}
\hline
\multirow{2}{*}{\textbf{Model}} & \multirow{2}{*}{\textbf{LLM}} & \multirow{2}{*}{\textbf{Visual Encoder}} & \multicolumn{2}{c}{\textbf{Movie}} & \multicolumn{2}{c}{\textbf{Action}} & \multicolumn{2}{c}{\textbf{Tv Show}} & \multicolumn{2}{c}{\textbf{Profession}} & \multicolumn{2}{c}{\textbf{Landmark}} \\ \cline{4-13} 
 &  &  & \multicolumn{1}{c}{\textbf{MC}} & \multicolumn{1}{c}{\textbf{TF}} & \multicolumn{1}{c}{\textbf{MC}} & \multicolumn{1}{c}{\textbf{TF}} & \multicolumn{1}{c}{\textbf{MC}} & \multicolumn{1}{c}{\textbf{TF}} & \multicolumn{1}{c}{\textbf{MC}} & \multicolumn{1}{c}{\textbf{TF}} & \multicolumn{1}{c}{\textbf{MC}} & \multicolumn{1}{c}{\textbf{TF}} \\ \hline
MiniGPT-4 & Vicuna-7B & EVA-CLIP ViT-G & 29.53 & 51.03 & 45.06 & 50.67 & 30.43 & 47.09 & 35.85 & 46.88 & 52.37 & 51.15 \\
MiniGPT-4 & Vicuna-13B & EVA-CLIP ViT-G & 49.76 & 49.20 & 68.99 & 50.16 & 38.24 & 39.47 & 51.85 & 49.02 & 68.67 & 51.69 \\
MiniGPT-4 & LLaMA2-7B & EVA-CLIP ViT-G & 61.35 & 52.94 & 91.38 & 57.72 & 47.06 & 55.26 & 81.48 & 49.02 & 87.95 & 58.43 \\
MiniGPT-2 & LLaMA2-7B & EVA-CLIP ViT-G & 72.95 & 50.27 & 95.31 & 62.22 & 73.53 & 52.63 & 79.63 & 62.75 & 87.95 & 50.56 \\
BLIP2 & Flan-T5-XL & EVA-CLIP ViT-G & 72.45 & 67.72 & 97.16 & 93.02 & 57.97 & 57.00 & 81.60 & 75.45 & 98.11 & 93.44 \\
BLIP2 & OPT-3B & EVA-CLIP ViT-G & 31.88 & 41.18 & 34.49 & 48.23 & 32.35 & 47.37 & 25.93 & 54.90 & 32.53 & 43.82 \\
BLIP2 & OPT-7B & EVA-CLIP ViT-G & 19.90 & 42.78 & 24.05 & 47.91 & 23.53 & 47.37 & 18.52 & 50.98 & 24.39 & 39.33 \\
InstructBLIP & Vicuna-7B & EVA-CLIP ViT-G & 83.57 & 56.68 & 97.88 & 71.38 & 88.24 & 55.26 & 81.48 & 58.82 & 97.59 & 71.91 \\
InstructBLIP & Vicuna-13B & EVA-CLIP ViT-G & 83.57 & 59.36 & 98.79 & 64.63 & 88.24 & 63.16 & 77.78 & 60.78 & 97.59 & 66.29 \\
InstructBLIP & Flan-T5-XL & EVA-CLIP ViT-G & 77.93 & 67.24 & 97.60 & 93.98 & 68.12 & 56.04 & 82.08 & 75.00 & 97.16 & 95.08 \\
InstructBLIP & Flan-T5-XXL & EVA-CLIP ViT-G & 75.80 & 62.83 & 97.75 & 93.54 & 66.67 & 61.84 & 84.91 & 78.57 & 96.85 & 90.16 \\
LLava-1.5 & Vicuna-7B & CLIP ViT-L & 55.71 & 53.54 & 74.10 & 50.89 & 53.14 & 57.49 & 57.08 & 58.93 & 83.28 & 55.08 \\
LLava-1.5 & Vicuna-13B & CLIP ViT-L & 65.45 & 55.91 & 79.04 & 59.58 & 58.45 & 58.45 & 67.45 & 62.95 & 94.01 & 57.05 \\
Otter & LLaMA-7B & CLIP ViT-L & 66.01 & 59.53 & 66.62 & 72.44 & 70.05 & 57.00 & 68.87 & 55.36 & 59.31 & 66.56 \\
Shikra & LLaMA-7B & CLIP ViT-L & 68.34 & 61.26 & 78.44 & 78.01 & 60.87 & 57.97 & 82.08 & 68.30 & 88.96 & 70.82 \\
Xcomposer-VL & InternLM-7B & EVA-CLIP ViT-G & 80.82 & 77.64 & 97.01 & 94.80 & 78.26 & 70.53 & 84.43 & 76.34 & 97.16 & 95.41 \\
Xcomposer2-VL & InternLM2-7B & CLIP ViT-L & 91.79 & 88.77 & 98.18 & 95.34 & 85.29 & 84.21 & 88.89 & \underline{90.20} & 98.80 & \underline{100.00} \\
Qwen-VL-Chat & Qwen-7B & OpenClip ViT-bigG & 71.08 & 49.61 & 95.96 & 63.52 & 68.12 & 42.51 & 80.66 & 51.34 & 95.90 & 55.41 \\
Emu2-Chat & LLaMA-33B & EVA2-CLIP-E & 91.30 & 54.55 & 98.79 & 55.14 & 85.29 & 50.00 & 90.74 & 41.18 & \underline{100.00} & 69.66 \\
GLM-4V & GLM-4-9B-Chat & EVA2-CLIP-E & \textbf{93.72} & \textbf{91.98} & \textbf{99.55} & \textbf{98.39} & \textbf{97.06} & \textbf{89.47} & 88.89 & \textbf{94.12} & \textbf{100.00} & \textbf{100.00} \\
MiniCPM-V2.5 & Llama3-Instruct 8B & SigLIP SoViT-400m & \underline{93.24} & \underline{89.84} & 98.18 & 92.77 & \underline{94.12} & \underline{89.47} & 90.74 & 88.24 & 100.00 & 98.88 \\
Yi-VL & Yi-6B-Chat & OpenClip ViT-H & 89.37 & 82.89 & 97.88 & 85.85 & 88.24 & 89.47 & \textbf{94.44} & 74.51 & 97.59 & 93.26 \\
mPLUG-Owl-2 & LLaMA2-7B & CLIP ViT-L & 90.34 & 76.47 & 96.82 & 85.37 & 82.35 & 81.58 & 87.04 & 82.35 & 98.80 & 86.52 \\
Phi-3-Vision & Phi-3 & CLIP ViT-L & 87.44 & 70.05 & 97.13 & 90.03 & 85.29 & 65.79 & 88.89 & 84.31 & 100.00 & 96.63 \\  \hdashline
GPT-4o & / & / & 86.47 & 74.87 & 96.96 & \underline{94.81} & 79.41 & 86.84 & 83.33 & 80.39 & 96.39 & 98.88 \\
Gemini-1.5-Pro & / & / & 92.72 & 63.64 & \underline{99.24} & 89.55 & 93.55 & 68.42 & \underline{90.74} & 84.31 & 100.00 & 94.38 \\ \hline
\end{tabular}
}
\end{table}

\begin{table}[h]
\centering
\caption{Evaluation results on the 5 perceptual subtasks. The top two results on each subtask are \textbf{bolded} and \underline{underlined}, respectively. ``MC'' and ``TF'' indicate the accuracy (\%) of ``Multi-choices'' and ``True-or-false'', respectively.}
\label{tab:leaderboard_2}
\scalebox{0.64}{
\begin{tabular}{ccccccccccccc}
\hline
\multirow{2}{*}{\textbf{Model}} & \multirow{2}{*}{\textbf{LLM}} & \multirow{2}{*}{\textbf{Visual Encoder}} & \multicolumn{2}{c}{\textbf{Anime}} & \multicolumn{2}{c}{\textbf{Clothes}} & \multicolumn{2}{c}{\textbf{Celebrity}} & \multicolumn{2}{c}{\textbf{Food}} & \multicolumn{2}{c}{\textbf{Plant}} \\ \cline{4-13} 
 &  &  & MC & TF & MC & TF & MC & TF & MC & TF & MC & TF \\ \hline
MiniGPT-4 & Vicuna-7B & EVA-CLIP ViT-G & 27.44 & 47.27 & 30.61 & 49.49 & 27.26 & 47.87 & 43.96 & 50.00 & 47.07 & 49.94 \\
MiniGPT-4 & Vicuna-13B & EVA-CLIP ViT-G & 44.25 & 45.14 & 47.76 & 53.57 & 31.06 & 51.05 & 73.66 & 51.21 & 68.35 & 49.94 \\
MiniGPT-4 & LLaMA2-7B & EVA-CLIP ViT-G & 40.71 & 53.47 & 66.94 & 51.43 & 54.49 & 54.46 & 81.61 & 52.90 & 84.97 & 58.14 \\
MiniGPT-2 & LLaMA2-7B & EVA-CLIP ViT-G & 50.44 & 47.92 & 65.71 & 47.50 & 57.81 & 53.00 & 86.58 & 56.76 & 89.76 & 59.38 \\
BLIP2 & Flan-T5-XL & EVA-CLIP ViT-G & 56.91 & 61.77 & 84.08 & 75.26 & 80.78 & 76.45 & 93.24 & 90.56 & 92.53 & 90.15 \\
BLIP2 & OPT-3B & EVA-CLIP ViT-G & 26.55 & 45.14 & 28.57 & 45.00 & 31.89 & 48.70 & 32.42 & 48.31 & 35.11 & 48.82 \\
BLIP2 & OPT-7B & EVA-CLIP ViT-G & 24.78 & 46.53 & 29.39 & 45.00 & 23.29 & 51.86 & 26.96 & 49.15 & 26.73 & 50.25 \\
InstructBLIP & Vicuna-7B & EVA-CLIP ViT-G & 83.19 & 70.14 & 89.80 & 58.21 & 80.40 & 58.67 & 93.66 & 68.12 & 96.28 & 73.29 \\
InstructBLIP & Vicuna-13B & EVA-CLIP ViT-G & 76.11 & 62.50 & 93.06 & 60.71 & 77.91 & 54.13 & 95.03 & 65.46 & 95.61 & 72.30 \\
InstructBLIP & Flan-T5-XL & EVA-CLIP ViT-G & 60.77 & 63.65 & 88.27 & 82.72 & 83.11 & 75.75 & 93.96 & 91.36 & 92.98 & 91.36 \\
InstructBLIP & Flan-T5-XXL & EVA-CLIP ViT-G & 62.80 & 67.42 & 87.24 & 87.32 & 82.11 & 82.93 & 94.28 & 91.91 & 94.01 & 93.90 \\
LLava-1.5 & Vicuna-7B & CLIP ViT-L & 48.37 & 48.02 & 47.86 & 50.92 & 56.38 & 54.56 & 56.68 & 50.83 & 50.80 & 50.06 \\
LLava-1.5 & Vicuna-13B & CLIP ViT-L & 58.54 & 49.34 & 65.41 & 59.00 & 61.04 & 57.14 & 82.51 & 56.15 & 79.59 & 55.91 \\
Otter & LLaMA-7B & CLIP ViT-L & 62.80 & 54.99 & 47.24 & 71.37 & 44.15 & 63.76 & 44.44 & 79.46 & 69.29 & 80.56 \\
Shikra & LLaMA-7B & CLIP ViT-L & 47.15 & 58.19 & 76.63 & 59.71 & 63.90 & 62.65 & 84.26 & 67.72 & 88.73 & 66.39 \\
Xcomposer-VL & InternLM-7B & EVA-CLIP ViT-G & 74.80 & 74.01 & 86.84 & 87.22 & 88.23 & 87.25 & 93.72 & 90.96 & 92.34 & 91.55 \\
Xcomposer2-VL & InternLM2-7B & CLIP ViT-L & 88.50 & 84.03 & 95.10 & 92.50 & 84.72 & 88.82 & 96.89 & \textbf{94.08} & 97.74 & 93.42 \\
Qwen-VL-Chat & Qwen-7B & OpenClip ViT-bigG & 73.37 & 54.61 & 80.71 & 62.17 & 87.17 & 50.03 & 92.05 & 54.24 & 90.99 & 59.21 \\
Emu2-Chat & LLaMA-33B & EVA2-CLIP-E & 86.73 & 66.67 & 94.69 & 49.64 & 94.02 & 47.16 & 97.76 & 49.15 & \underline{98.54} & 54.66 \\
GLM-4V & GLM-4-9B-Chat & EVA2-CLIP-E & \textbf{92.92} & \textbf{96.53} & \underline{95.51} & 87.86 & \textbf{98.67} & \textbf{96.11} & \underline{98.26} & 92.63 & \textbf{99.07} & \textbf{96.89} \\
MiniCPM-V2.5 & Llama3-Instruct 8B & SigLIP SoViT-400m & 82.30 & \underline{89.58} & 91.84 & \underline{92.50} & 88.04 & 80.71 & 96.02 & \underline{93.48} & 98.14 & \underline{95.03} \\
Yi-VL & Yi-6B-Chat & OpenClip ViT-H & 77.88 & 81.94 & 92.24 & 76.07 & 87.21 & 86.71 & 95.28 & 79.59 & 96.14 & 79.63 \\
mPLUG-Owl-2 & LLaMA2-7B & CLIP ViT-L & 83.19 & 70.14 & 89.80 & 81.79 & 86.05 & 72.61 & 93.79 & 84.18 & 94.95 & 84.47 \\
Phi-3-Vision & Phi-3 & CLIP ViT-L & 76.11 & 72.22 & 88.57 & 90.00 & 83.89 & 62.07 & 94.16 & 91.79 & 96.01 & 93.66 \\  \hdashline
GPT-4o & / & / & 72.57 & 88.11 & 89.34 & 88.89 & 70.72 & 55.92 & \textbf{98.26} & 92.75 & 97.47 & 94.90 \\
Gemini-1.5-Pro & / & / & \underline{91.82} & 76.39 & \textbf{97.55} & \textbf{93.57} & \underline{94.85} & \underline{91.57} & 98.01 & 92.75 & 97.87 & 93.54 \\ \hline
\end{tabular}
}
\end{table}

\clearpage

\begin{table}[h]
\centering
\caption{Evaluation results on the 5 perceptual subtasks. The top two results on each subtask are \textbf{bolded} and \underline{underlined}, respectively. ``MC'' and ``TF'' indicate the accuracy (\%) of ``Multi-choices'' and ``True-or-false'', respectively.}
\label{tab:leaderboard_3}
\scalebox{0.64}{
\begin{tabular}{ccccccccccccc}
\hline
\multirow{2}{*}{\textbf{Model}} & \multirow{2}{*}{\textbf{LLM}} & \multirow{2}{*}{\textbf{Visual Encoder}} & \multicolumn{2}{c}{\textbf{Age}} & \multicolumn{2}{c}{\textbf{Gender}} & \multicolumn{2}{c}{\textbf{Expression}} & \multicolumn{2}{c}{\textbf{Race}} & \multicolumn{2}{c}{\textbf{Animal}} \\ \cline{4-13} 
 &  &  & MC & TF & MC & TF & MC & TF & MC & TF & MC & TF \\ \hline
MiniGPT-4 & Vicuna-7B & EVA-CLIP ViT-G & 30.02 & 48.28 & 52.00 & 46.75 & 42.29 & 49.82 & 25.10 & 48.42 & 45.54 & 49.39 \\
MiniGPT-4 & Vicuna-13B & EVA-CLIP ViT-G & 29.32 & 50.06 & 58.30 & 48.60 & 59.51 & 53.58 & 33.08 & 51.51 & 60.39 & 52.37 \\
MiniGPT-4 & LLaMA2-7B & EVA-CLIP ViT-G & 49.48 & 49.74 & 84.34 & 56.14 & 57.26 & 57.74 & 28.35 & 52.96 & 89.84 & 54.49 \\
MiniGPT-2 & LLaMA2-7B & EVA-CLIP ViT-G & 41.62 & 52.84 & 86.13 & 61.60 & 65.69 & 63.85 & 51.34 & 53.56 & 92.53 & 60.20 \\
BLIP2 & Flan-T5-XL & EVA-CLIP ViT-G & 62.72 & 58.71 & 99.32 & 94.12 & 88.41 & 75.30 & 75.76 & 71.87 & 96.62 & 95.59 \\
BLIP2 & OPT-3B & EVA-CLIP ViT-G & 24.61 & 43.28 & 61.41 & 48.60 & 34.53 & 53.84 & 31.55 & 51.51 & 36.92 & 48.78 \\
BLIP2 & OPT-7B & EVA-CLIP ViT-G & 29.84 & 42.76 & 53.13 & 47.81 & 30.93 & 53.78 & 26.55 & 49.82 & 28.40 & 49.43 \\
InstructBLIP & Vicuna-7B & EVA-CLIP ViT-G & 70.16 & 57.36 & 99.66 & 74.00 & 83.13 & 67.23 & 84.16 & 67.55 & 97.46 & 69.17 \\
InstructBLIP & Vicuna-13B & EVA-CLIP ViT-G & 67.28 & 63.44 & 99.33 & 65.49 & 86.28 & 74.38 & 81.23 & 56.45 & 97.01 & 71.45 \\
InstructBLIP & Flan-T5-XL & EVA-CLIP ViT-G & 64.06 & 58.48 & 99.58 & 86.91 & 91.11 & 78.06 & 78.28 & 73.98 & 97.08 & 95.52 \\
InstructBLIP & Flan-T5-XXL & EVA-CLIP ViT-G & 67.84 & 81.16 & 99.58 & 98.50 & 82.56 & 80.11 & 82.46 & 83.55 & 97.32 & 95.71 \\
LLava-1.5 & Vicuna-7B & CLIP ViT-L & 37.74 & 55.40 & 53.62 & 49.89 & 63.49 & 51.34 & 42.55 & 51.34 & 49.45 & 52.18 \\
LLava-1.5 & Vicuna-13B & CLIP ViT-L & 49.29 & 59.78 & 98.60 & 83.59 & 70.12 & 58.87 & 70.81 & 63.45 & 86.12 & 58.74 \\
Otter & LLaMA-7B & CLIP ViT-L & 36.78 & 50.92 & 78.99 & 77.84 & 72.50 & 61.02 & 42.38 & 58.62 & 81.57 & 83.57 \\
Shikra & LLaMA-7B & CLIP ViT-L & 65.07 & 54.96 & 98.14 & 73.29 & 90.08 & 70.52 & 75.20 & 54.19 & 90.20 & 70.32 \\
Xcomposer-VL & InternLM-7B & EVA-CLIP ViT-G & 67.09 & 79.02 & 99.53 & 97.57 & 90.25 & 81.34 & 79.47 & 76.26 & 97.43 & \underline{96.32} \\
Xcomposer2-VL & InternLM2-7B & CLIP ViT-L & \underline{86.65} & \underline{91.73} & 99.11 & 99.03 & \underline{96.74} & \textbf{94.41} & 88.38 & 89.02 & 98.06 & 89.56 \\
Qwen-VL-Chat & Qwen-7B & OpenClip ViT-bigG & 53.19 & 48.71 & 97.50 & 59.18 & 85.55 & 64.04 & 74.43 & 55.95 & 95.45 & 63.89 \\
Emu2-Chat & LLaMA-33B & EVA2-CLIP-E & 84.95 & 49.35 & \underline{99.89} & 50.91 & 92.58 & 53.06 & \underline{91.95} & 51.39 & 98.21 & 53.51 \\
GLM-4V & GLM-4-9B-Chat & EVA2-CLIP-E & 86.13 & \textbf{93.80} & \textbf{100.00} & \textbf{100.00} & \textbf{98.09} & \underline{94.28} & \textbf{94.51} & \textbf{94.69} & \textbf{99.40} & \textbf{99.02} \\
MiniCPM-V2.5 & Llama3-Instruct 8B & SigLIP SoViT-400m & 77.23 & 68.86 & 99.55 & \underline{99.76} & 95.05 & 92.07 & 81.61 & 88.66 & \underline{98.65} & 94.62 \\
Yi-VL & Yi-6B-Chat & OpenClip ViT-H & 71.86 & 86.69 & 98.88 & 98.66 & 91.11 & 88.30 & 84.55 & 74.31 & 97.31 & 83.85 \\
mPLUG-Owl-2 & LLaMA2-7B & CLIP ViT-L & 73.43 & 81.91 & 98.21 & 96.60 & 88.30 & 84.27 & 80.84 & 84.44 & 95.52 & 83.85 \\
Phi-3-Vision & Phi-3 & CLIP ViT-L & \textbf{86.65} & 70.03 & 98.99 & 99.51 & 96.40 & 81.66 & 91.19 & \underline{89.51} & 96.86 & 88.42 \\  \hdashline
GPT-4o & / & / & 75.79 & 86.43 & 78.61 & 96.71 & 95.50 & 91.03 & 37.60 & 51.75 & 95.21 & 95.27 \\
Gemini-1.5-Pro & / & / & 82.07 & 77.26 & 99.55 & 98.91 & 96.40 & 87.65 & 31.42 & 63.57 & 97.91 & 89.72 \\ \hline
\end{tabular}
}
\end{table}

\begin{table}[h]
\centering
\caption{Evaluation results on the 5 perceptual subtasks. The top two results on each subtask are \textbf{bolded} and \underline{underlined}, respectively. ``MC'' and ``TF'' indicate the accuracy (\%) of ``Multi-choices'' and ``True-or-false'', respectively.}
\label{tab:leaderboard_4}
\scalebox{0.64}{
\begin{tabular}{cccllllllllll}
\hline
\multirow{2}{*}{\textbf{Model}} & \multirow{2}{*}{\textbf{LLM}} & \multirow{2}{*}{\textbf{Visual Encoder}} & \multicolumn{2}{c}{\textbf{Object}} & \multicolumn{2}{c}{\textbf{Text}} & \multicolumn{2}{c}{\textbf{Style}} & \multicolumn{2}{c}{\textbf{Background}} & \multicolumn{2}{c}{\textbf{Color}} \\ \cline{4-13} 
 &  &  & \multicolumn{1}{c}{MC} & \multicolumn{1}{c}{TF} & \multicolumn{1}{c}{MC} & \multicolumn{1}{c}{TF} & \multicolumn{1}{c}{MC} & \multicolumn{1}{c}{TF} & \multicolumn{1}{c}{MC} & \multicolumn{1}{c}{TF} & \multicolumn{1}{c}{MC} & \multicolumn{1}{c}{TF} \\ \hline
MiniGPT-4 & Vicuna-7B & EVA-CLIP ViT-G & 52.54 & 50.66 & 29.68 & 50.25 & 35.68 & 18.89 & 31.20 & 48.98 & 35.55 & 48.86 \\
MiniGPT-4 & Vicuna-13B & EVA-CLIP ViT-G & 61.27 & 51.89 & 36.55 & 51.79 & 56.00 & 44.16 & 48.15 & 50.99 & 54.25 & 50.75 \\
MiniGPT-4 & LLaMA2-7B & EVA-CLIP ViT-G & 92.33 & 55.07 & 49.67 & 57.84 & 73.19 & 48.73 & 58.10 & 49.50 & 69.79 & 54.14 \\
MiniGPT-2 & LLaMA2-7B & EVA-CLIP ViT-G & 96.67 & 60.97 & 43.60 & 57.54 & 79.35 & 48.22 & 56.19 & 50.20 & 69.02 & 59.79 \\
BLIP2 & Flan-T5-XL & EVA-CLIP ViT-G & 90.26 & 89.86 & 74.33 & 62.32 & 83.87 & 35.45 & 76.23 & 72.02 & 88.58 & 86.64 \\
BLIP2 & OPT-3B & EVA-CLIP ViT-G & 37.83 & 44.63 & 26.68 & 53.08 & 27.90 & 44.67 & 29.21 & 45.03 & 32.34 & 46.33 \\
BLIP2 & OPT-7B & EVA-CLIP ViT-G & 26.83 & 48.33 & 26.25 & 51.98 & 29.35 & 47.21 & 25.21 & 46.42 & 23.00 & 47.46 \\
InstructBLIP & Vicuna-7B & EVA-CLIP ViT-G & 97.67 & 67.32 & 78.74 & 62.60 & 99.28 & 47.21 & 85.82 & 62.23 & 94.21 & 68.08 \\
InstructBLIP & Vicuna-13B & EVA-CLIP ViT-G & 98.17 & 71.10 & 81.34 & 57.04 & 99.82 & 48.73 & 90.16 & 56.46 & 96.33 & 59.13 \\
InstructBLIP & Flan-T5-XL & EVA-CLIP ViT-G & 90.70 & 91.10 & 75.14 & 60.76 & 83.94 & 33.05 & 77.08 & 73.41 & 90.26 & 86.45 \\
InstructBLIP & Flan-T5-XXL & EVA-CLIP ViT-G & 90.26 & 92.13 & 78.61 & 60.09 & 83.65 & 38.62 & 78.77 & 80.23 & 88.81 & 88.76 \\
LLava-1.5 & Vicuna-7B & CLIP ViT-L & 58.36 & 48.52 & 50.36 & 50.73 & 38.44 & 19.45 & 46.62 & 51.57 & 42.03 & 52.71 \\
LLava-1.5 & Vicuna-13B & CLIP ViT-L & 80.16 & 56.66 & 64.65 & 54.02 & 72.17 & 19.73 & 70.05 & 56.85 & 66.01 & 53.37 \\
Otter & LLaMA-7B & CLIP ViT-L & 48.62 & 83.92 & 67.20 & 62.46 & 82.27 & 23.54 & 71.49 & 67.59 & 49.89 & 57.76 \\
Shikra & LLaMA-7B & CLIP ViT-L & 74.35 & 70.26 & 64.99 & 53.78 & 79.65 & 22.97 & 79.58 & 64.28 & 83.06 & 60.57 \\
Xcomposer-VL & InternLM-7B & EVA-CLIP ViT-G & 90.19 & 91.24 & 72.46 & 79.01 & 79.65 & 29.18 & 81.35 & 81.01 & 86.60 & 89.37 \\
Xcomposer2-VL & InternLM2-7B & CLIP ViT-L & 98.83 & 97.13 & 89.26 & 88.79 & 100.00 & 94.42 & \underline{92.06} & \underline{92.94} & 94.11 & \textbf{95.95} \\
Qwen-VL-Chat & Qwen-7B & OpenClip ViT-bigG & 90.12 & 65.15 & 73.82 & 52.18 & 82.63 & 21.07 & 78.73 & 51.57 & 86.37 & 53.25 \\
Emu2-Chat & LLaMA-33B & EVA2-CLIP-E & 98.67 & 51.44 & 87.09 & 53.67 & 100.00 & 47.21 & 91.85 & 48.51 & \underline{95.46} & 52.07 \\
GLM-4V & GLM-4-9B-Chat & EVA2-CLIP-E & \textbf{99.00} & \textbf{98.49} & \textbf{93.60} & \textbf{87.20} & \textbf{100.00} & 94.92 & \textbf{94.39} & \textbf{95.33} & 94.69 & 94.54 \\
MiniCPM-V2.5 & Llama3-Instruct 8B & SigLIP SoViT-400m & 98.50 & 96.97 & 88.39 & \underline{84.13} & 100.00 & \textbf{98.98} & 90.69 & 88.87 & \textbf{96.33} & \underline{95.10} \\
Yi-VL & Yi-6B-Chat & OpenClip ViT-H & 98.17 & 75.04 & 84.49 & 75.60 & 100.00 & 84.26 & 90.05 & 75.94 & 94.79 & 92.09 \\
mPLUG-Owl-2 & LLaMA2-7B & CLIP ViT-L & 96.33 & 88.80 & 81.45 & 65.77 & 99.28 & 78.17 & 85.82 & 74.65 & 91.80 & 88.23 \\
Phi-3-Vision & Phi-3 & CLIP ViT-L & 98.17 & \underline{97.73} & 32.32 & 28.87 & 100.00 & 95.94 & 90.14 & 78.42 & 88.63 & 89.75 \\  \hdashline
GPT-4o & / & / & 95.17 & 94.25 & 82.97 & 82.59 & \underline{100.00} & \underline{95.94} & 91.19 & 89.46 & 83.11 & 89.36 \\
Gemini-1.5-Pro & / & / & \underline{98.83} & 95.76 & \underline{89.37} & 80.85 & 98.91 & 92.39 & 91.96 & 88.17 & 87.15 & 88.42 \\ \hline
\end{tabular}
}
\end{table}

It can be observed that certain models exhibit inconsistency when faced with different forms of questions. We have noted this phenomenon across several state-of-the-art models (e.g., Emu2~\citep{Emu2} and Qwen-VL-Chat~\citep{bai2023qwen-vl}).

\begin{figure}[htbp]
    \centering
    \begin{subfigure}{0.45\textwidth}
        \centering
        \includegraphics[width=\linewidth]{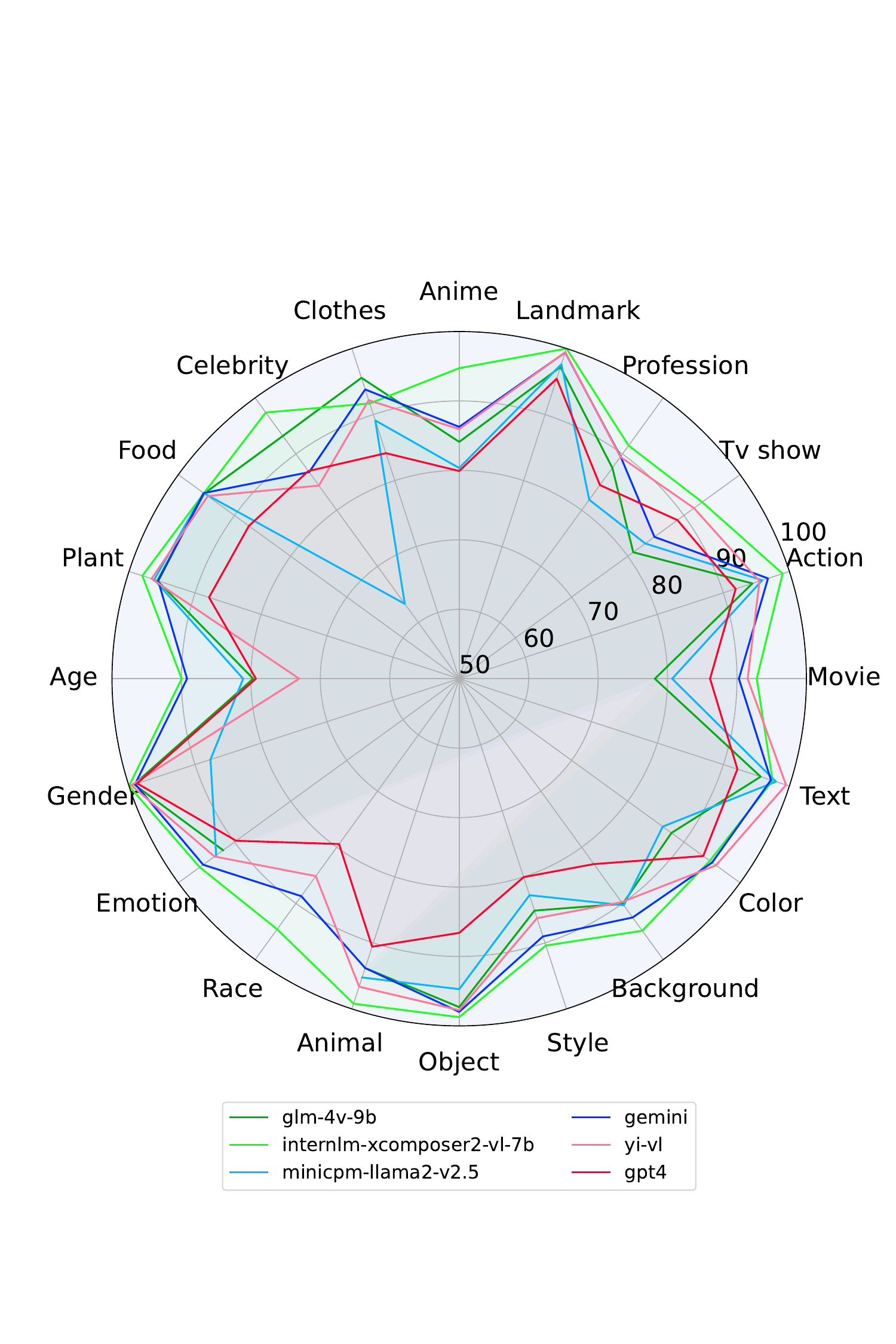}
        \caption{Comparison of 6 top LVLMs on 20 subtasks under clean scenario. The full score of each subtask is 100.}
        \label{fig:image1}
    \end{subfigure}
    \hfill
    \begin{subfigure}{0.45\textwidth}
        \centering
        \includegraphics[width=\linewidth]{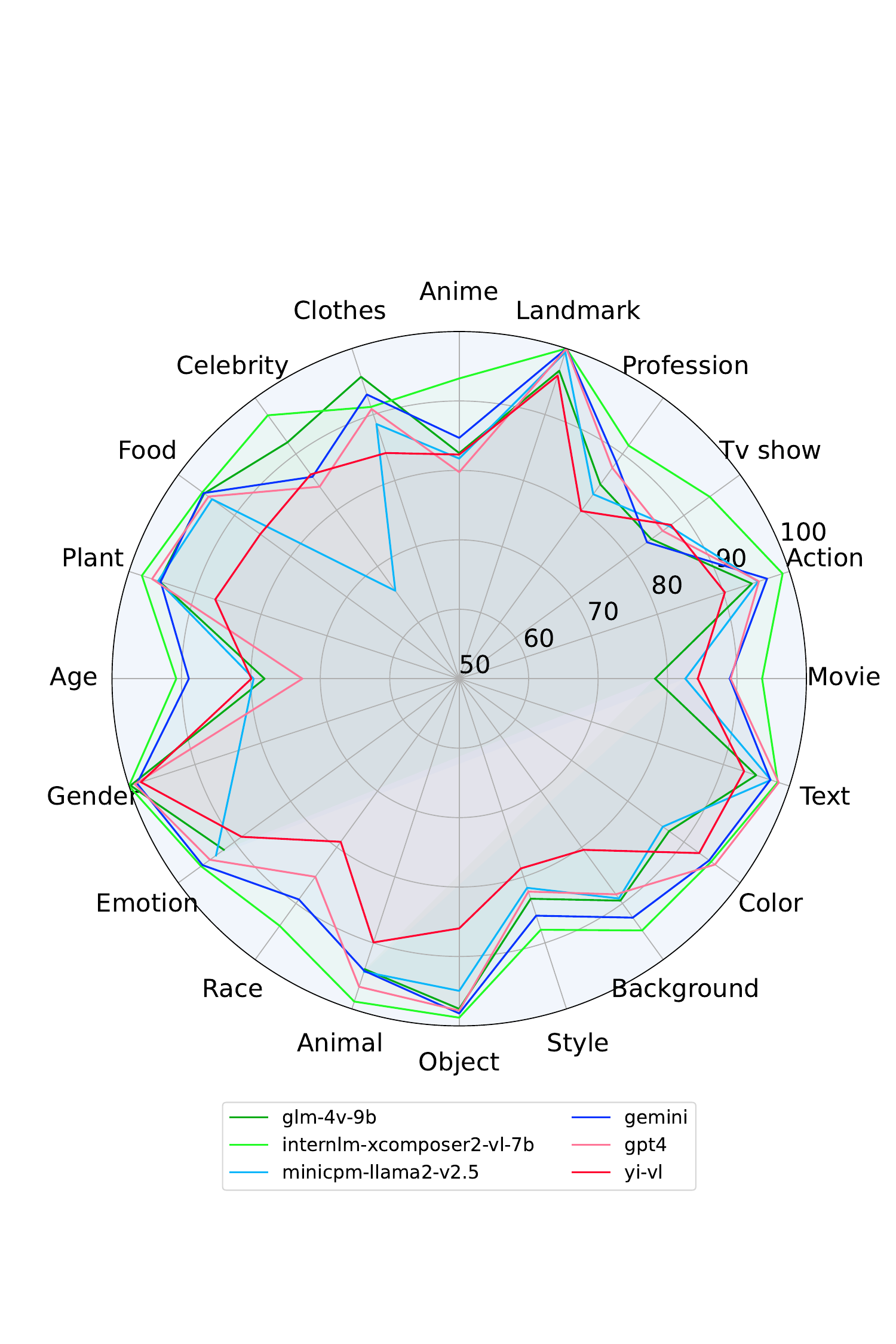}
        \caption{Comparison of 6 top LVLMs on 20 subtasks under corruption scenario. The full score of each subtask is 100.}
        \label{fig:image2}
    \end{subfigure}

    \begin{subfigure}{0.45\textwidth}
        \centering
        \includegraphics[width=\linewidth]{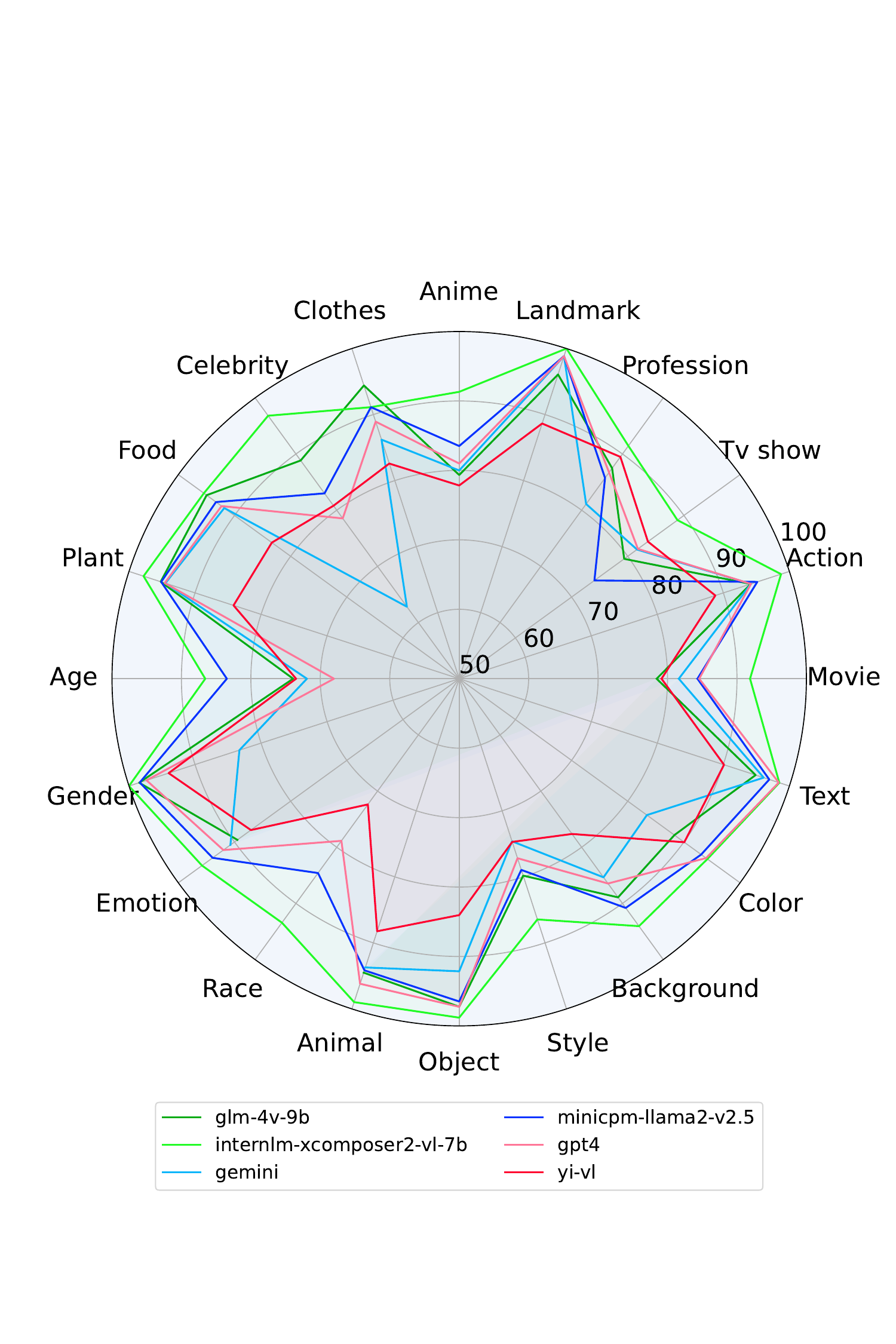}
        \caption{Comparison of 6 top LVLMs on 20 subtasks under adversarial attacking scenario. The full score of each subtask is 100.}
        \label{fig:image3}
    \end{subfigure}
    \hfill
    \begin{subfigure}{0.45\textwidth}
        \centering
        \includegraphics[width=\linewidth]{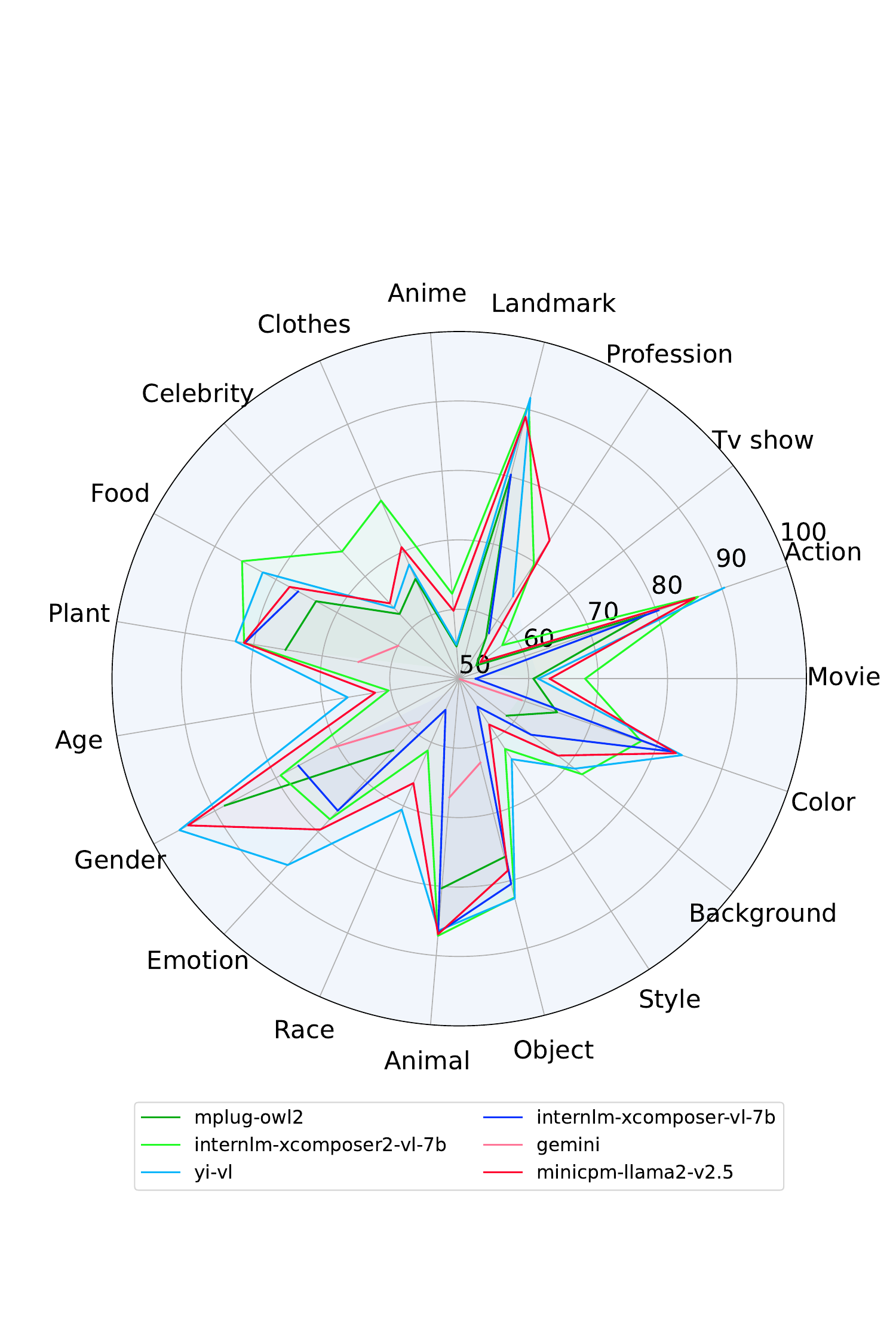}
        \caption{Comparison of 6 top LVLMs on 19 subtasks under print attacking scenario. The full score of each subtask is 100.}
        \label{fig:image4}
    \end{subfigure}

    \caption{Radar chart for the top 6 LVLMs in each scenario,}
    \label{fig:four_images}
\end{figure}

We also provide a visualization of the scores for the top six models across 20 subtasks in different scenarios, as shown in Fig. \ref{fig:four_images}. It can be observed that the same model exhibits performance variations across different dimensions. For instance, age perception is the most significant weakness for the Yi-VL~\citep{ai2024yi}. Additionally, by comparing the radar charts of different scenarios, we can see that corruption scenario has the least impact on the models, while print attacking is almost catastrophic. This highlights the need for future work to focus on improving model robustness against print attacking.

\clearpage


\vspace{-0.2cm}


\clearpage

\clearpage

\section{Discussion}

\subsection{General Discussion}

\textbf{Limitation.} Dysca is the dynamic and scalable benchmark, offering evaluation for 20 perceptual subtasks under 51 image styles and 4 scenarios.  However, generating data for evaluating cognition abilities (e.g., commonsense reasoning) presents challenge within the existing framework. This limitation arises from the reliance on predefined rules for prompt and question generation, which may not adequately capture the complexity of cognitive-level questions.

\textbf{Synthesis Data for Training / Fine-tuning.} The use of synthetic data for model training / fine-tuning has been adopted in the field of Natural Language Processing (NLP)~\citep{llama3}. In this work, we do not explore the possibility of utilizing our benchmark for model training. Our primary goal in this paper is to provide a large-scale evaluation benchmark that addresses the issue of data leakage in current multimodal evaluation benchmarks and offers evaluation results across multiple subtasks, scenarios, question types and styles. Nevertheless, considering that Dysca has the capability to synthesize high-resolution and unlimited amounts of annotated multimodal data, we believe that Dysca also holds potential as a training data synthesis tool for LVLMs.

\textbf{Reproducibility and License.} All the experiments are built on 8 * RTX 4090. All the data and the code for generation and evaluation are released at \url{https://github.com/Robin-WZQ/Dysca}. The license of Dysca is \href{https://huggingface.co/stabilityai/stable-diffusion-xl-base-1.0/blob/main/LICENSE.md}{``CreativeML Open RAIL++-M''}, which follows the license set by the Stable Diffusion XL.

\textbf{Ethical Concerns.} Our Dysca leverages the Stable Diffusion XL~\citep{podell2023sdxl} to generate images. In order to prevent the model generating unsafe images, e.g., NSFW and offensive images, lots of efforts have been made. First, we use the safety checker~\citep{Rando2022RedTeamingTS} to post filter the unsafe images. The safety checker is a post-processor deployed by model developers~\citep{podell2023sdxl} to prevent the generation of NSFW images. With the unsafe image that is recognized by the safety checker, the model's output will be a blank image. Besides, we manually exclude the specific styles or the word that may trigger the unsafe images generation from the Metadata $M$. To validate the safety of Dysca, we utilize the NudeNet detector~\citep{NudeNet} with with a threshold of 0.8 to identify images that contain NSFW content. NudeNet is the widely used automatic tool for detecting NSFW images which is adopted by~\citep{gandikota2023erasing,10657770}. With 77847 images (clean scenario) in our Dysca, only 5 images are classified as NSFW (0.006\%). We check the 5 images and find they are all false positive samples. This indicates that our Dysca contains a minimal amount of such unsafety images.

\subsection{The Stability of Dysca} \label{stability}
\begin{figure}[h]
    \centering
    \includegraphics[width=0.8\linewidth]{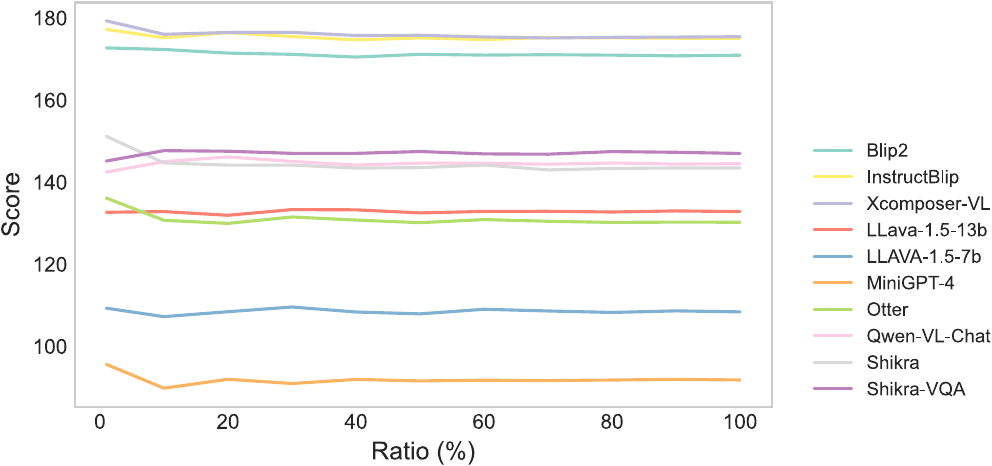}
    \caption{The tendency of 10 model's overall performance under clean scenario with different scale of evaluation data.}
    \label{fig:scale_result}
\end{figure}

In this section, we focus on examining the stability of Dysca. We partition Dysca into 11 different scales: 1\%, 10\%, 20\%, 30\%, 40\%, 50\%, 60\%, 70\%, 80\%, 90\% and 100\%. We randomly select 10 models and compute their evaluation scores using each of these data scales. The score is calculated as the sum of scores obtained from multiple-choice, true-or-false and free-form questions. As can be seen in Fig. \ref{fig:scale_result}, when the evaluation data scale is less than 30\% of Dysca (i.e., less than 46.8K samples), the evaluation score show significant fluctuations. When the data scale exceeds 40\%, we obtain the stable results, reflecting current scale of Dysca achieves the stable and reliable evaluation results. 
Although 40\% evaluation scale of Dysca has achieved stable scores, Dysca aims to provide more than just stable rankings, but also draws on massive amounts of data to provide in-depth feedback across different image styles and perceptual subtasks.

\section{The Correctness of Dysca}
Since our generative pipeline relies on text-to-image models, the quality of the Dysca dataset is inherently influenced by the generation performance of the model. To address this issue, we have employed extensive data cleaning techniques to ensure the correctness of the dataset. Furthermore, to validate this, we randomly sample 7.7K images (10\% of our data) and manually select images that contain wrong content or potentially lead to incorrect responses. In the end, a total of 167 images (2.2\%) are filtered. Compared to previous but wide spread benchmarks, the incorrect data ratio of Dysca is less~\citep{northcutt2021pervasive}. For example, the ImageNet dataset~\citep{deng2009imagenet} has at least 2,916 errors in its validation set, with an error rate of 6\%; the QuickDraw dataset~\citep{ha2018a} has at least 5 million errors, with an error rate of approximately 10\%. Subsequently, we compute the scores with and without these incorrect images, and the results are presented below.

\begin{table}[h]
\centering
\caption{The scores with and without these incorrect images.}
\label{tab:my-table}
\begin{tabular}{ccc}
\hline
\textbf{Model} & \textbf{Before} & \textbf{After} \\ \hline
minigpt4-vicuna-7b & 41.38 & {\color[HTML]{333333} 41.24} \\
minigpt4-vicuna-13b & 50.17 & {\color[HTML]{333333} 50.78} \\
minigpt4-llama2 & 56.61 & {\color[HTML]{333333} 56.82} \\
minigpt-v2 & 58.46 & {\color[HTML]{333333} 58.43} \\
blip2-flan-t5-xl & 65.30 & {\color[HTML]{333333} 65.32} \\
blip2-opt-3b & 39.54 & {\color[HTML]{333333} 39.80} \\
blip2-opt-7b & 39.55 & {\color[HTML]{333333} 39.58} \\
instructblip-vicuna-7b & 67.54 & {\color[HTML]{333333} 67.89} \\
instructblip-vicuna-13b & 64.89 & {\color[HTML]{333333} 65.02} \\
instructblip-flan-t5-xl & 66.54 & {\color[HTML]{333333} 66.73} \\
instructblip-flan-t5-xxl & 68.65 & {\color[HTML]{333333} 68.74} \\
llava-1.5-7b & 51.27 & {\color[HTML]{333333} 51.44} \\
llava-1.5-13b & 59.23 & {\color[HTML]{333333} 59.46} \\
otter & 54.90 & {\color[HTML]{333333} 54.75} \\
shikra-7b & 62.24 & {\color[HTML]{333333} 62.35} \\
internlm-xcomposer-vl-7b & 71.40 & {\color[HTML]{333333} 71.55} \\
internlm-xcomposer2-vl-7b & 79.13 & {\color[HTML]{333333} 79.23} \\
qwen-vl-chat & 62.18 & {\color[HTML]{333333} 62.26} \\
emu2-chat & 63.64 & {\color[HTML]{333333} 63.67} \\
glm-4v-9b & 82.09 & {\color[HTML]{333333} 82.13} \\
minicpm-llama2-v2.5 & 78.75 & 79.02 \\
yi-vl & 75.71 & 76.32 \\
mplug-owl2 & 74.09 & 74.43 \\
phi-3-vision & 73.23 & 73.57 \\ \hdashline
GPT-4o & 75.69 & 76.21 \\
Gemini-1.5-Pro-Vision-latest & 77.79 & 77.99 \\ \hline
\end{tabular}
\end{table}

As observed, the score difference is negligible, making the automatic evaluation result reliable. We believe that with the development of deep generative models, Dysca will serve as an evolving benchmark with improving quality.

\section{Blind Experiment} \label{blind}
To demonstrate that Dysca has minimal data leakage issues, we conduct the "Blind" experiment where only textual questions are provided to LVLMs. We present the evaluation results of LLaVa-1.5-7B ~\citep{liu2023visual_llava} on Dysca and the six benchmarks, including MMMU  ~\citep{10656299}, MMB ~\citep{liu2023mmbench}, ScienceQA ~\citep{lu2022learn_scienceQA}, AI2D ~\citep{10.1007/978-3-319-46493-0_15}, Seed ~\citep{li2023seedbench} and MathVista ~\citep{lu2024mathvista}.
\begin{table}[h]
\centering
\caption{Scores of LLaVa under the blind setting across 7 benchmarks. }
\label{tab:blind}
\scalebox{0.95}{
\begin{tabular}{cccccccc}
\hline
\textbf{Model} & \textbf{MMMU} & \textbf{MMB } & \textbf{ScienceQA } & \textbf{AI2D } & \textbf{SEED } & \textbf{MathVista } & \textbf{Dysca} \\ \hline
Random Choice  & 22.1          & 0            & 24.2               & 23.8          & 24.3          & 17.9               & 37.5           \\
LLaVA-1.5-7B   & 29.9          & 19.5         & 64.1               & 48.7          & 37.5          & 20.3               & 38.7\\          \hline
\end{tabular}
}
\end{table}

As shown in Tab. \ref{tab:blind}, LLAVA significantly outperforms random selection when only text questions are provided in the other 6 benchmarks. In contrast, Dysca achieves results closest to random selection, indicating that our work has a limited data leakage issue. 

\section{The Metadata ($M$) of Dysca} \label{Metadata}
Metadata (\(M\)) is the core of Dysca, which is randomly assembled from our collected source material and contains all the information needed to generate prompt (\(P\)), image (\(I\)), and question-answer pairs(\(Q\)). Specifically, the metadata is a data container that contains information in multiple dimensions about the foreground, the attributes corresponding to the foreground, the background, and the artistic style required to generate an image. Therefore, each instance of M is mapped one-to-one to a prompt, an image, and a set of question-answer pairs, respectively.

In order to ensure the quality and stability of the generated images, we carefully select the source material. First, for each perceptual subtask, we collect rich annotation material as described in Section 3.2. However, the metadata composed of these raw annotations is not always usable. On the one hand, some of the content is polysemous, which can easily be misinterpreted by the model's when generating images. On the other hand, there are backgrounds or artistic styles (e.g., ``Pokemon Style'', ``architectural style'', etc.) that negatively affect the quality of the image and do not accurately generate the desired content. In order to test the usability of these source materials, we went through several small-scale pre-generations covering all the source materials. After careful selection, we retain the clips that consistently produced high-quality images. The detailed information of the source materials are shown in Tab. \ref{tab:category_info}.

\begin{table}[htbp]
\centering
\setlength{\extrarowheight}{5pt}  
\caption{Detailed information of the source materials.}
\label{tab:category_info}
\begin{tabularx}{\textwidth}{c|X|c}
\hline
\textbf{Category} & \centering{\textbf{Data Description}} & \textbf{\#Numbers} \\
\hline
Style & We collected artistic styles from the \href{https://stable-diffusion-art.com/sdxl-styles/}{community}, which can be well rendered by Stable Diffusion model. We removed those which have strong reflect on the image content or may generate unsafe image. & 51 \\
\hline
Background & We have selected 20 rich backgrounds and they can be accurately generated by Stable Diffusion Model. & 20 \\
\hline
Age & We chose four well-characterized age nodes: 14, 25, 40, and 80. & 4 \\
\hline
Expression & We chose three characteristic expressions: smiling happily, shouting furiously, calm and placid. & 3 \\
\hline
Gender & Male and Female. & 2 \\
\hline
Race & We identified these five races based on the ethnicity that can be generated by Stable Diffusion: Caucasian, Asian, African, Indian, and Middle Eastern races & 5 \\
\hline
Profession & After pre-generation and careful selection, we chose 20 occupations with distinctive characteristics that are easy to generate. & 20 \\
\hline
Action & After pre-generation and careful selection, we chose 20 occupations with distinctive characteristics that are easy to generate. & 20 \\
\hline
Celebrity & After pre-generation and careful selection, we chose 50 well-known celebrities. & 50 \\
\hline
Animal & We selected a rich variety of animals, including mammals, birds, reptiles, insects, and aquatic animals, and can they be generated by the Stable Diffusion model accurately. & 67 \\
\hline
Plant & We selected a rich variety of plants, including flowers, trees, fruits, and vegetables, and they can be generated by the Stable Diffusion model accurately. & 37 \\
\hline
Clothes & We selected 16 common types of clothing that are highly distinguishable from each other. & 16 \\
\hline
Object & We took the annotations from the MSCOCO~\citep{mscoco} dataset after removing people, animals, and plants, and added some additional common objects. & 80 \\
\hline
Landmark & We chose 23 characteristic landmarks from around the globe and they can be generated by the Stable Diffusion model accurately. & 23 \\
\hline
Food & We collected 29 special dishes from around the globe and they can be generated by the Stable Diffusion model accurately. & 29 \\
\hline
Movie & We selected 106 movies titles from the rating list of \href{https://www.imdb.com/}{IMDb} based on the number of user reviews. & 106 \\
\hline
Anime & We selected 44 anime titles from the rating list of \href{https://www.imdb.com/}{IMDb} based on the number of user reviews. & 44 \\
\hline
TV shows & We selected 20 TV show titles from the rating list of \href{https://www.imdb.com/}{IMDb} based on the number of user reviews. & 20 \\
\hline
OCR & We randomly selected 5000 words from the IELTS vocabulary for the text material. Among them, words with length less than 3 were removed. & 5000 \\
\hline
Color & We selected 8 easily distinguishable colors: red, orange, yellow, green, blue, purple, white, black. & 8 \\
\hline
\end{tabularx}
\end{table}

\section{Scenarios Details} \label{Corrupted Scenarios}

\subsection{Print Attack Scenario}

\begin{figure}[h]
    \centering
    \includegraphics[width=0.8\linewidth]{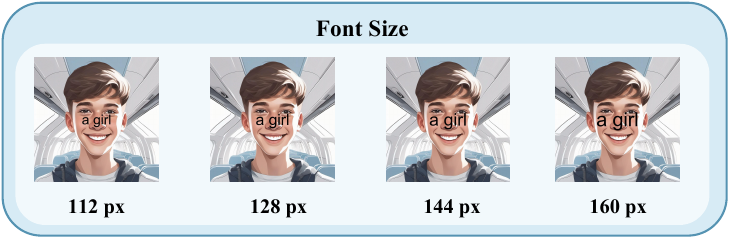}
    \caption{Images with different font size. '112px' means that the typos are 112 pixels in size.}
    \label{fig:font_size_example}
\end{figure}

\begin{figure}[h]
    \centering
    \includegraphics[width=0.8\linewidth]{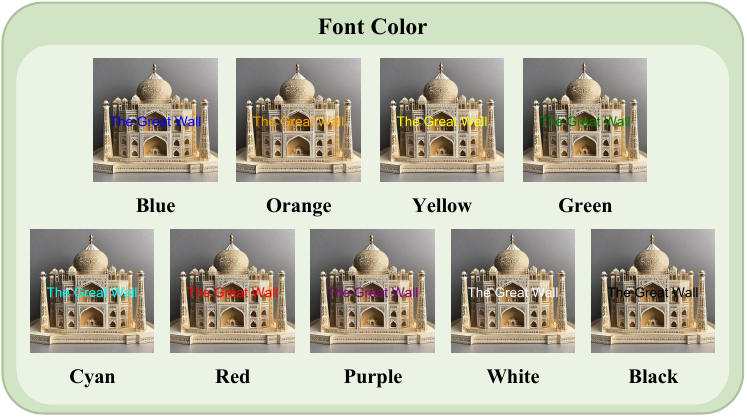}
    \caption{Images with different font color. 'Blue' means that the color of typos are in blue.}
    \label{fig:font_color_example}
\end{figure}

\begin{figure}[h]
    \centering
    \includegraphics[width=0.8\linewidth]{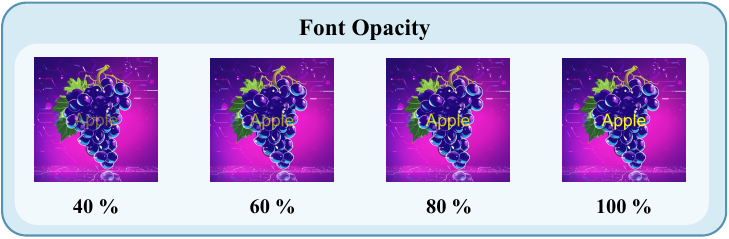}
    \caption{Images with different font opacity. '40\%' means that the transparency of typos are 40\% and '100\%' implies full opacity of typos. }
    \label{fig:font_opacity_example}
\end{figure}

\begin{figure}[h]
    \centering
    \includegraphics[width=0.8\linewidth]{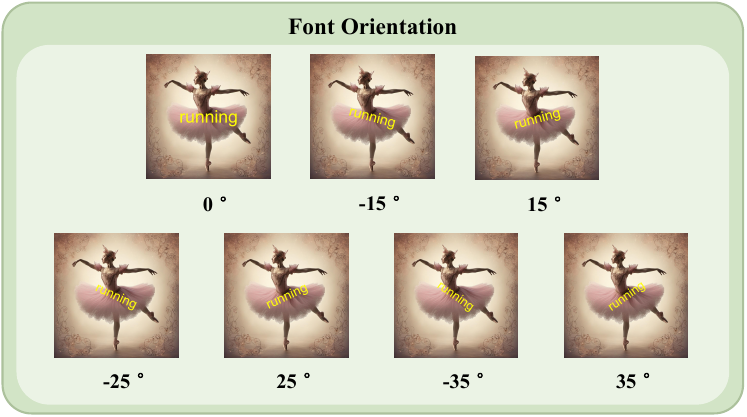}
    \caption{Images with different font orientation. '$15^{\circ}$' means that the orientation of typos are $15^{\circ}$ and $0^{\circ}$ implies the typos are horizontal.}
    \label{fig:font_orientation_example}
\end{figure}

\begin{figure}[h]
    \centering
    \includegraphics[width=0.8\linewidth]{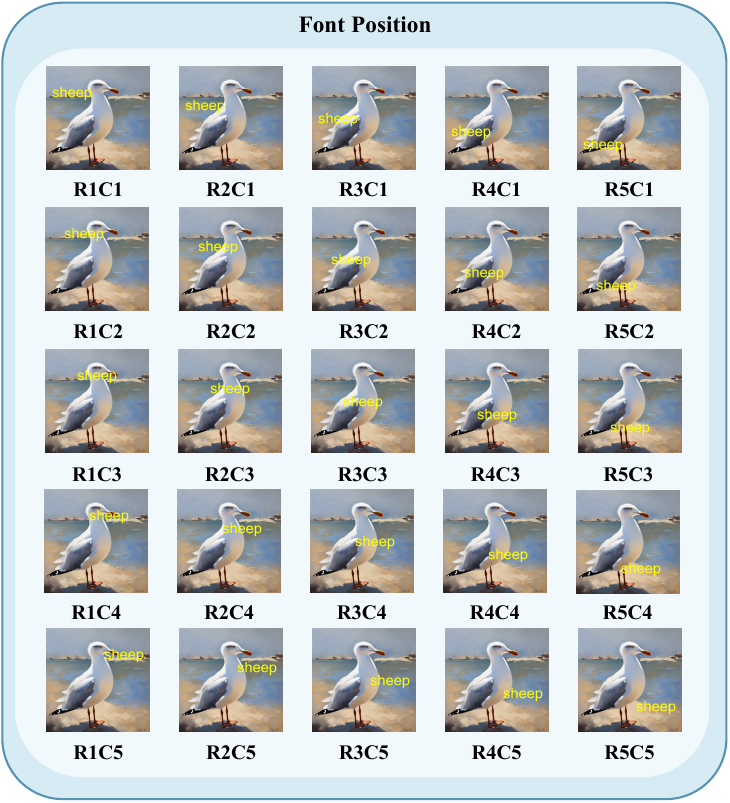}
    \caption{Images with different font position. An image is divided into a grid of 5 rows and 5 columns, leading to 25 sections. ’R1C1’ means the typo is located in row 1, column 1, which is the top left corner of the image.}
    \label{fig:font_position_example}
\end{figure}

Followed by the settings in~\citep{Cheng2024UnveilingTD}, we add the attack text on the images. Consider that the image resolution in our Dysca is much higher than the one in~\citep{Cheng2024UnveilingTD}, we increase more font form in terms of font position and font orientation. Fig. \ref{fig:font_size_example} to Fig. \ref{fig:font_position_example} shows the detailed examples.

\subsection{Corruption Scenario}
\begin{figure}[h]
    \centering
    \includegraphics[width=1\linewidth]{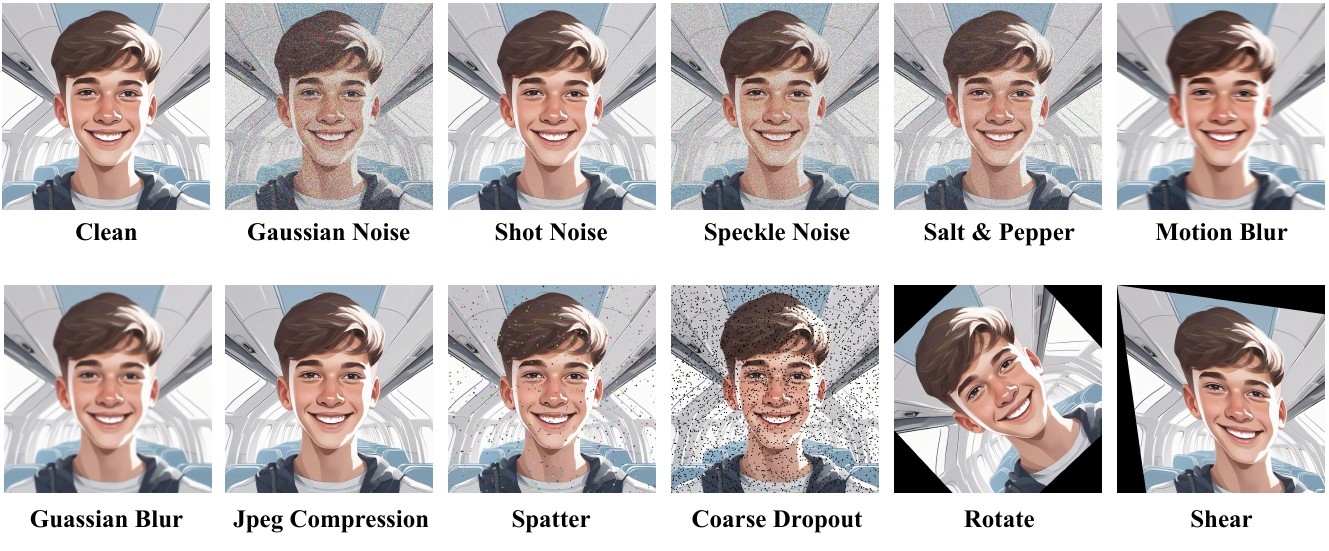}
    \caption{Examples of 11 image corruptions applied to a single clean image.}
    \label{fig:corrupted_examples}
\end{figure}

Examples of the 11 image corruptions are shown in Fig \ref{fig:corrupted_examples}.

\clearpage
\section{Hard Samples} \label{hard samples}
\begin{figure}[h]
    \centering
    \includegraphics[width=0.65\linewidth]{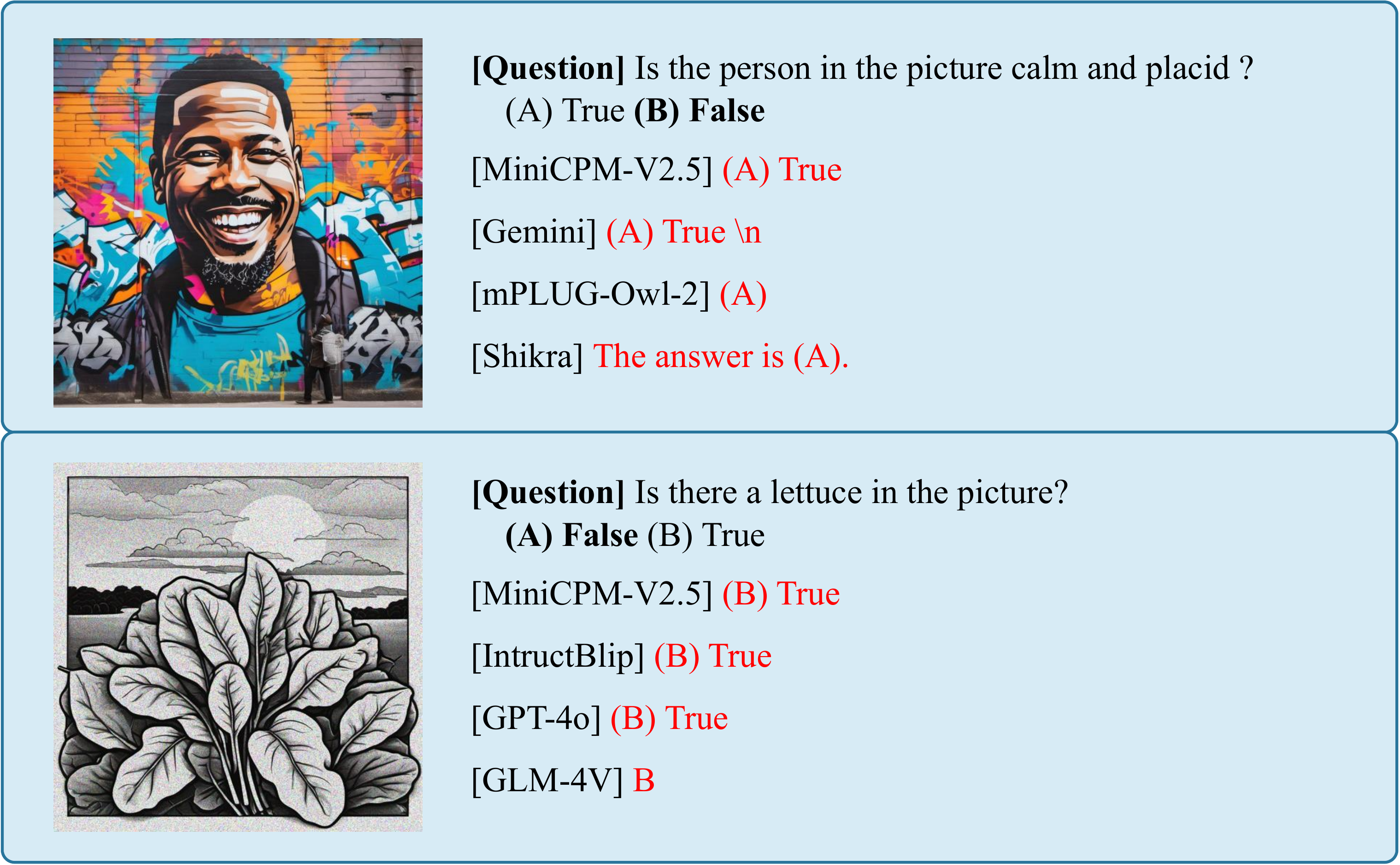}
    \caption{Hard samples.}
    \label{fig:hard_example_1}
\end{figure}

\begin{figure}[h]
    \centering
    \includegraphics[width=0.65\linewidth]{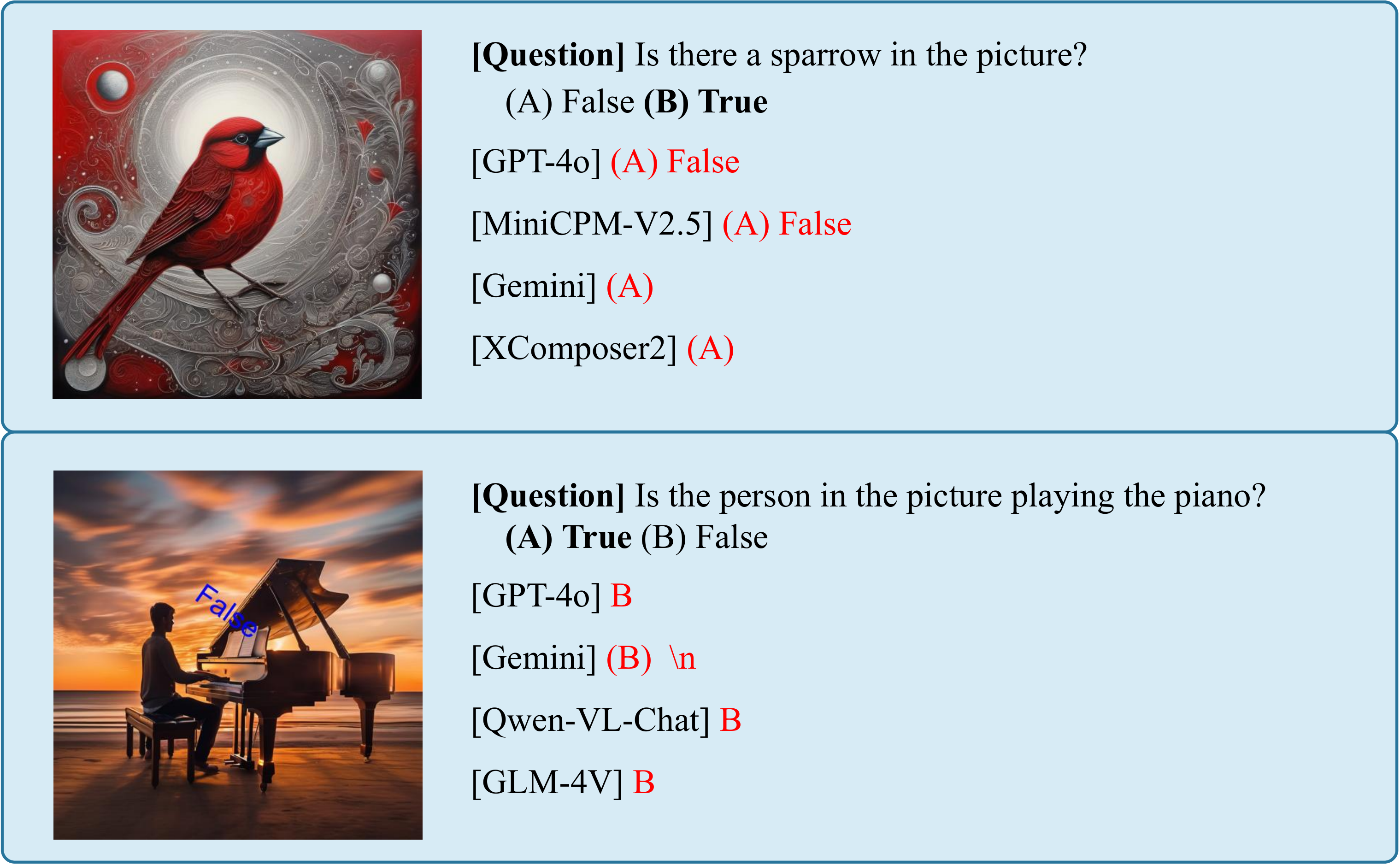}
    \caption{Hard samples.}
    \label{fig:hard_example_2}
\end{figure}

\begin{figure}[h]
    \centering
    \includegraphics[width=0.65\linewidth]{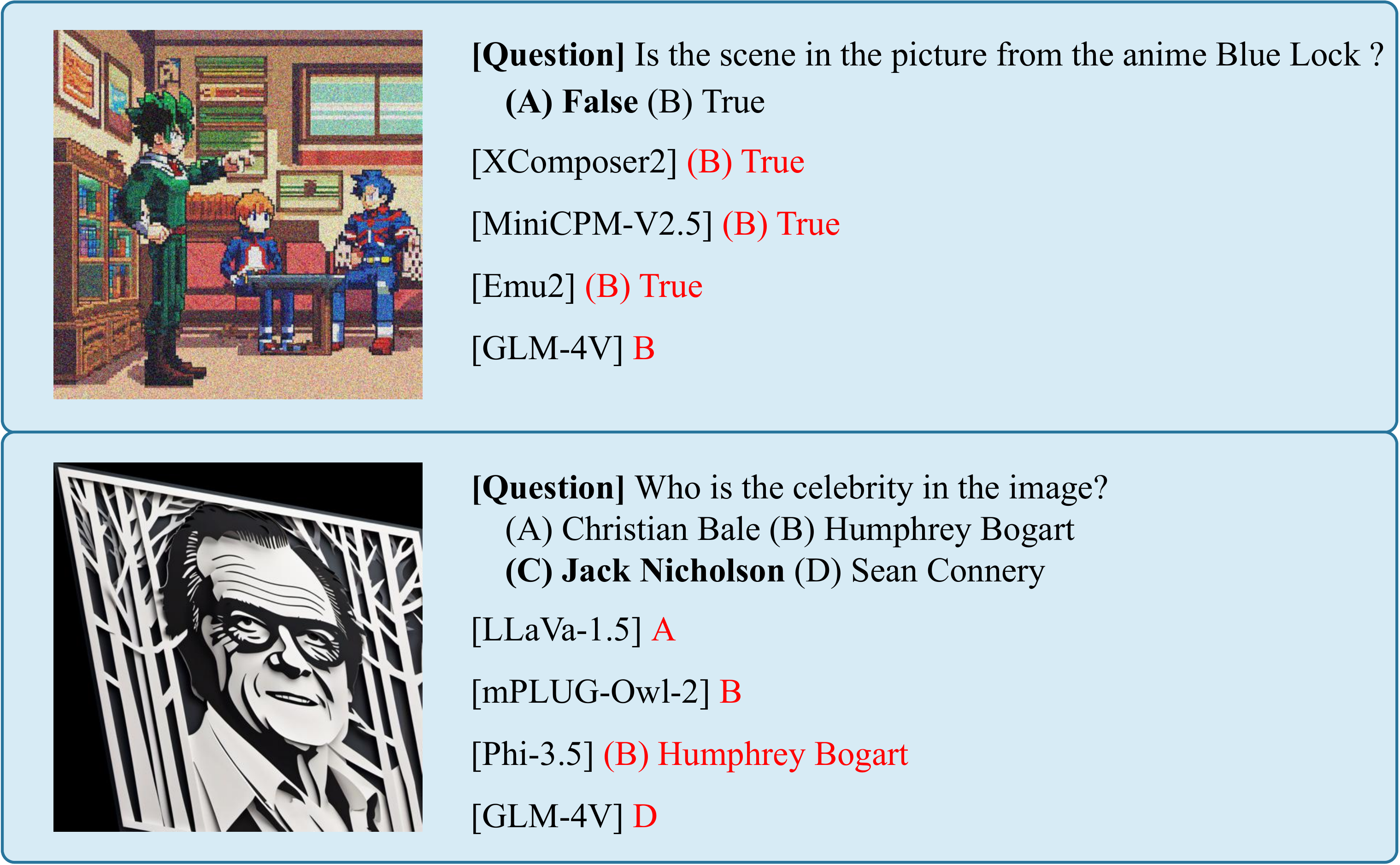}
    \caption{Hard samples.}
    \label{fig:hard_example_3}
\end{figure}

In this section, we provide hard samples that LVLMs are likely to answer incorrectly, as shown in Fig. \ref{fig:hard_example_1} to Fig. \ref{fig:hard_example_3}. 
\vspace{-0.5cm}
\vspace{-0.5cm}

\clearpage
\section{More Examples of Dysca} \label{Examples of Dysca}

For each subject we collected in Metadata ($M$), we display one example of their prompt ($P$), generated image ($I$) and corresponding question-answer pairs ($Q$).

\begin{figure}[h]
    \centering
    \includegraphics[width=1\linewidth]{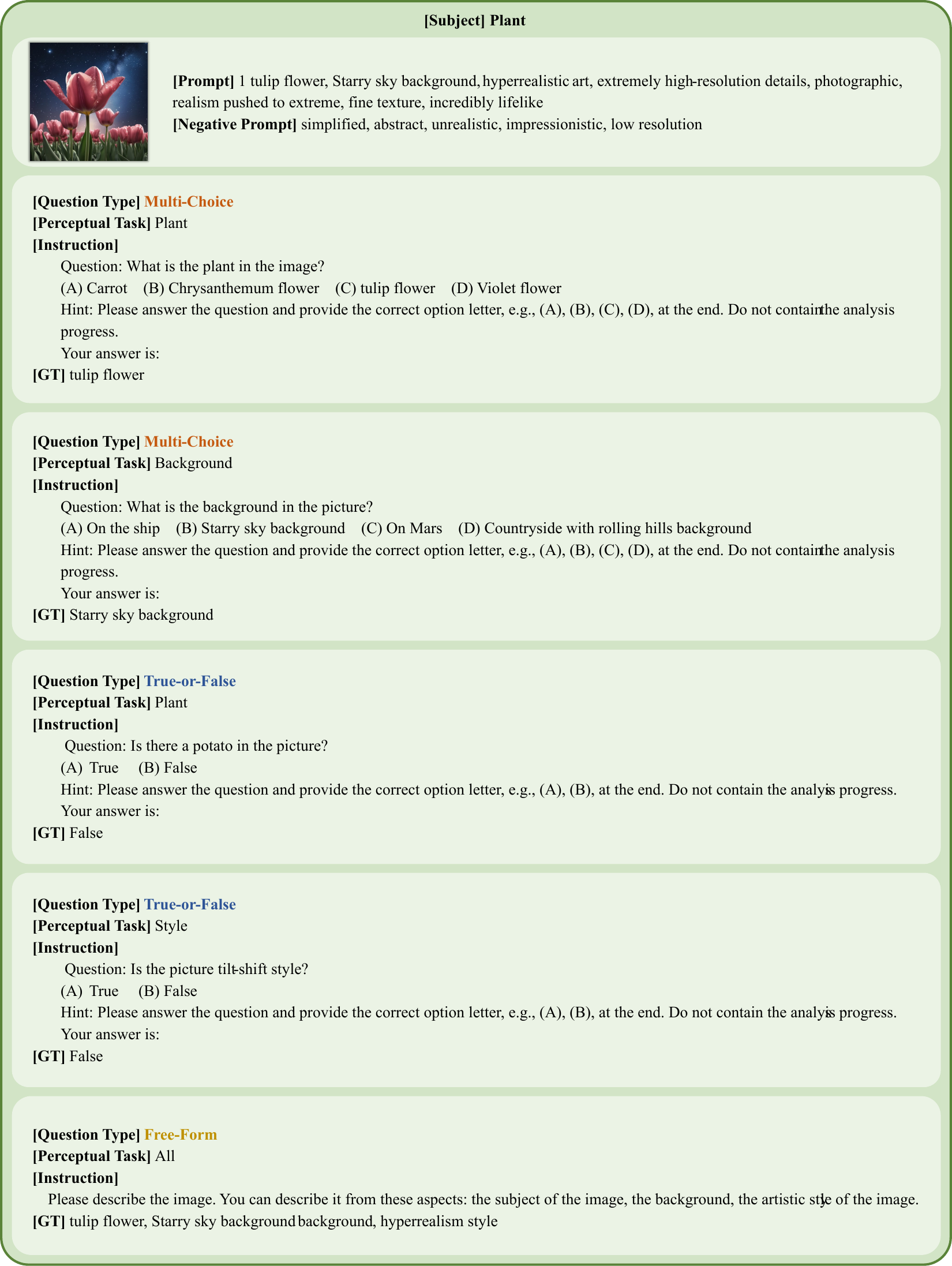}
    \caption{Plant}
    \label{fig:plant_example}
\end{figure}

\begin{figure}[h]
    \centering
    \includegraphics[width=1\linewidth]{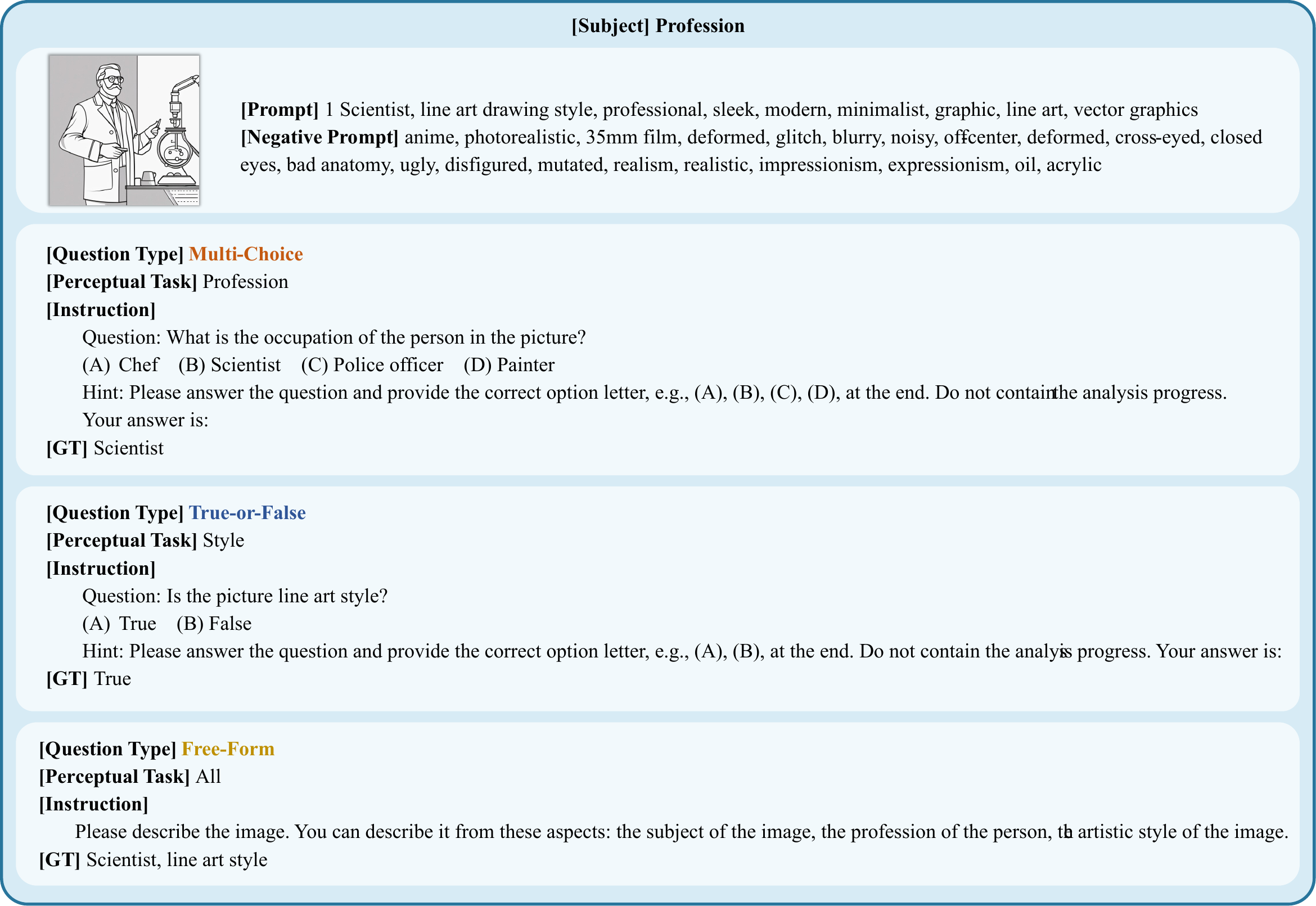}
    \caption{Profession}
    \label{fig:profession_examples}
\end{figure}

\begin{figure}[h]
    \centering
    \includegraphics[width=1\linewidth]{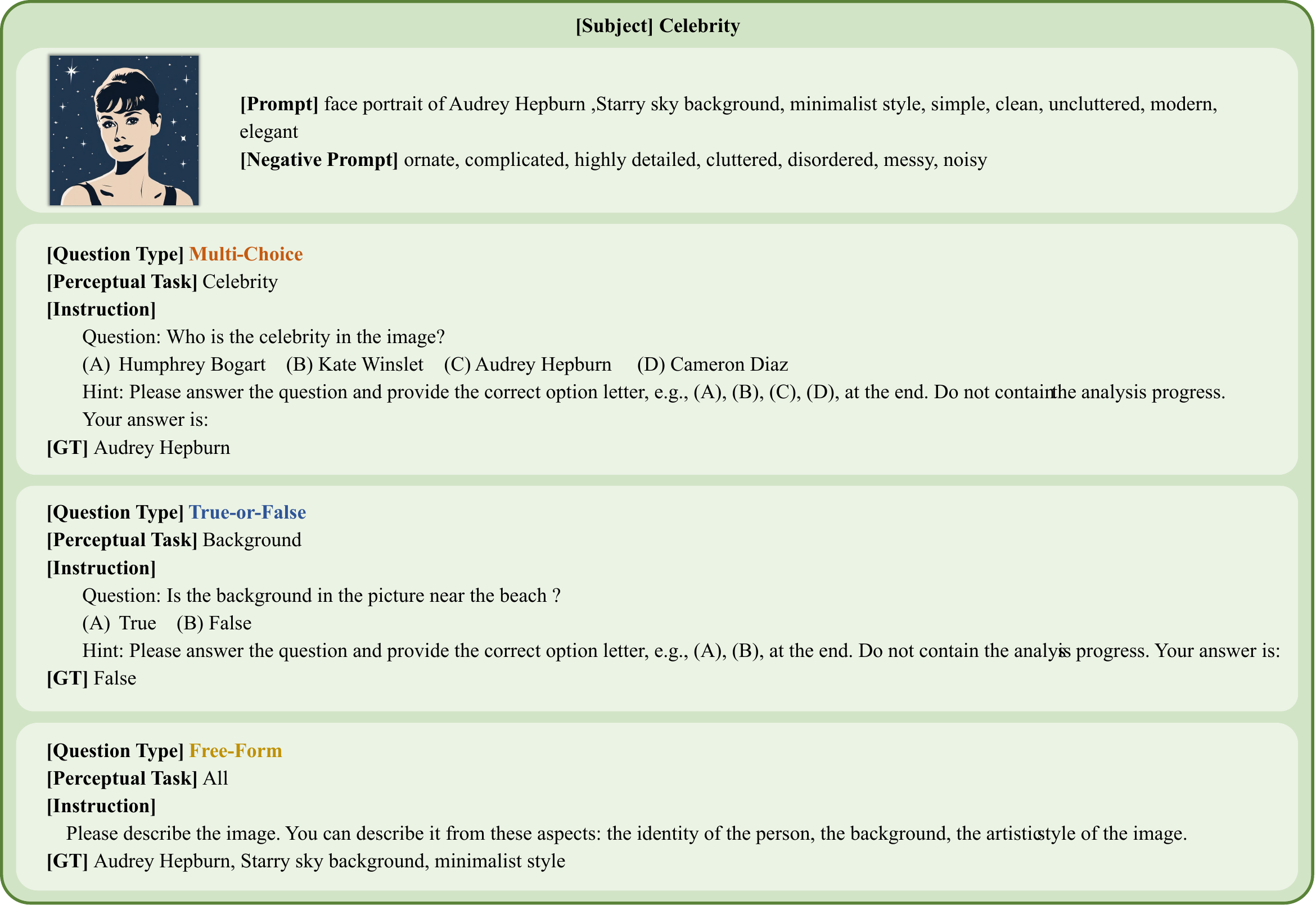}
    \caption{Celebrity}
    \label{fig:celebrity_example}
\end{figure}

\begin{figure}[h]
    \centering
    \includegraphics[width=1\linewidth]{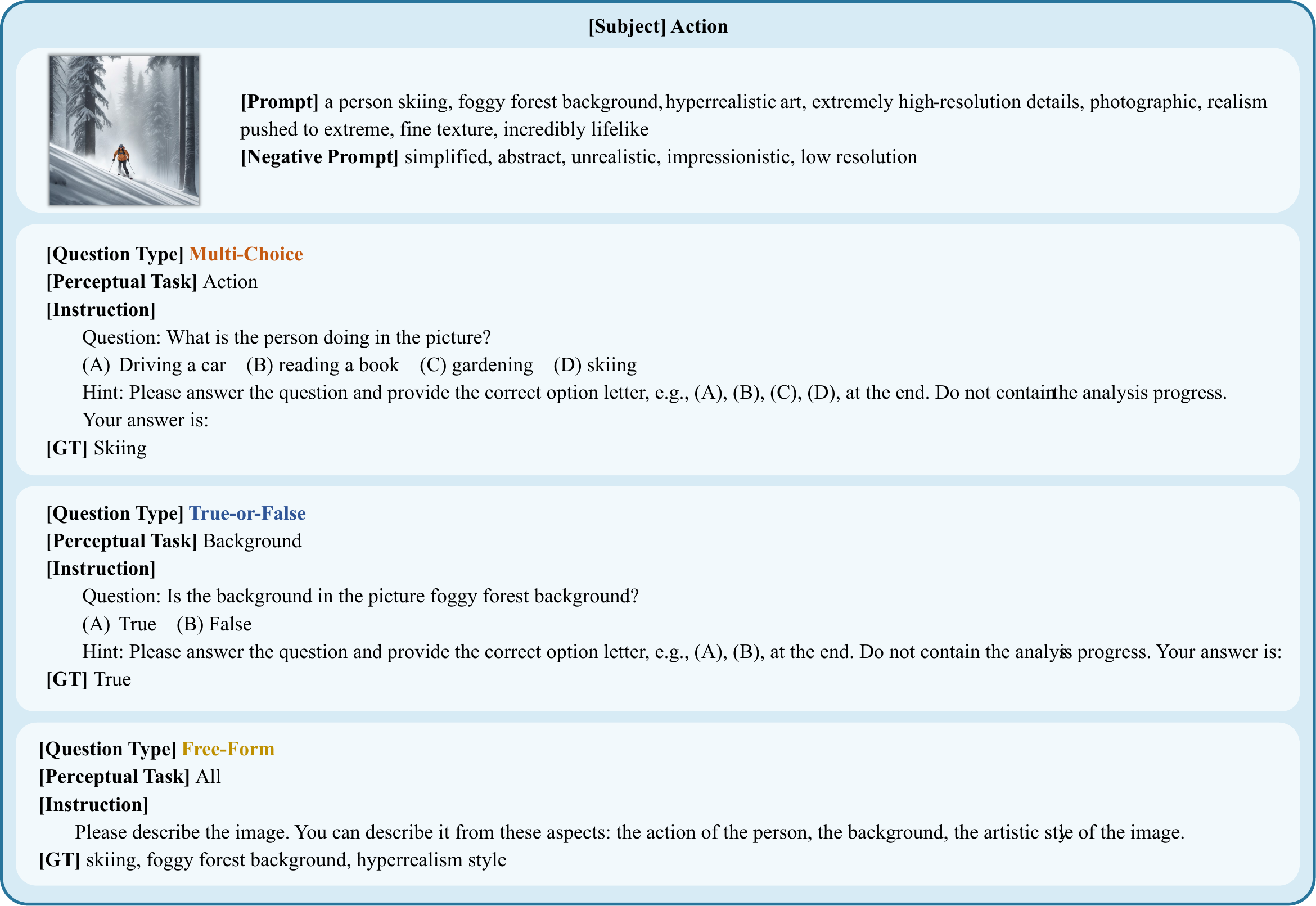}
    \caption{Action}
    \label{fig:action_example}
\end{figure}

\begin{figure}[h]
    \centering
    \includegraphics[width=1\linewidth]{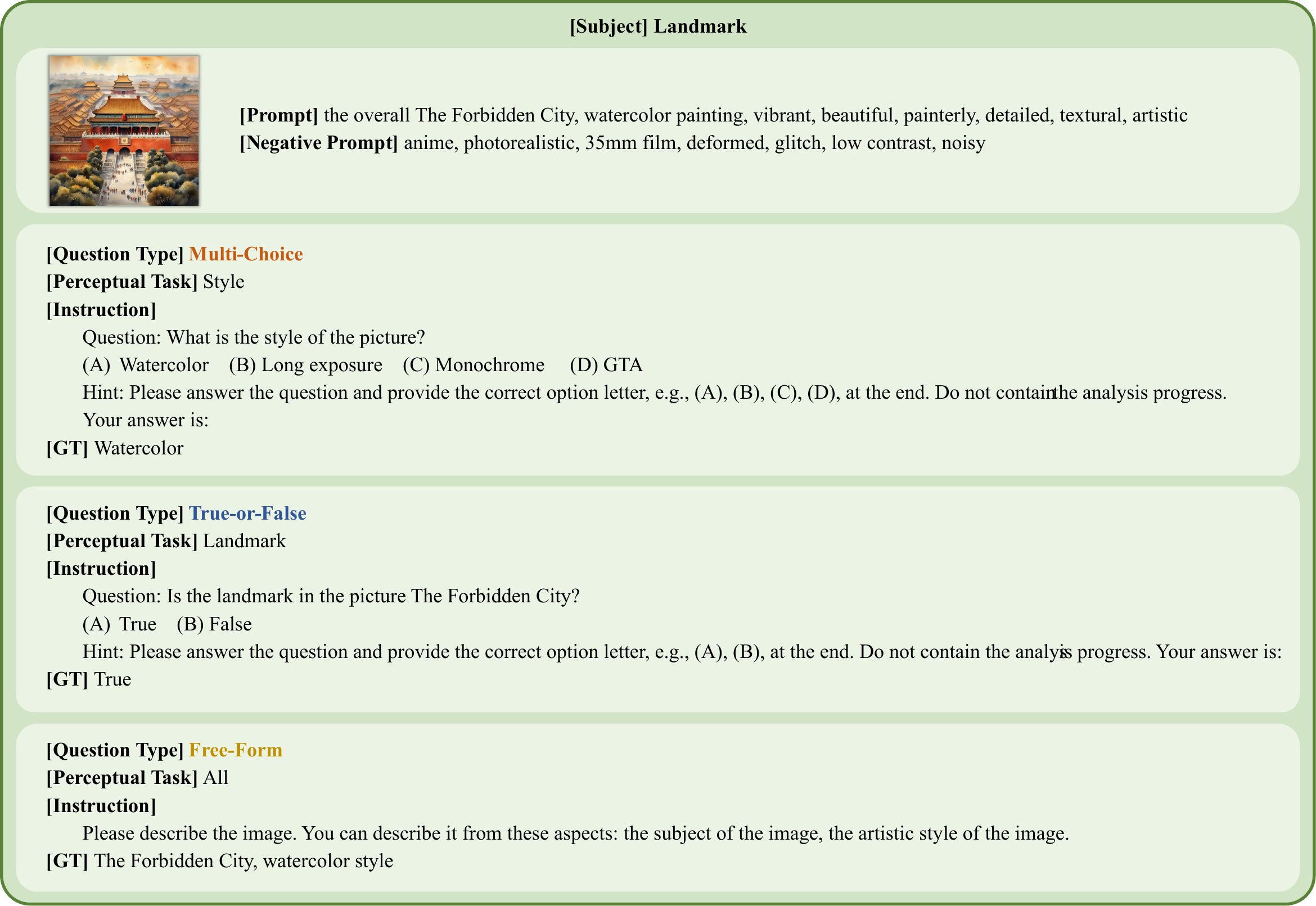}
    \caption{Landmark}
    \label{fig:landmark_example}
\end{figure}

\begin{figure}[h]
    \centering
    \includegraphics[width=1\linewidth]{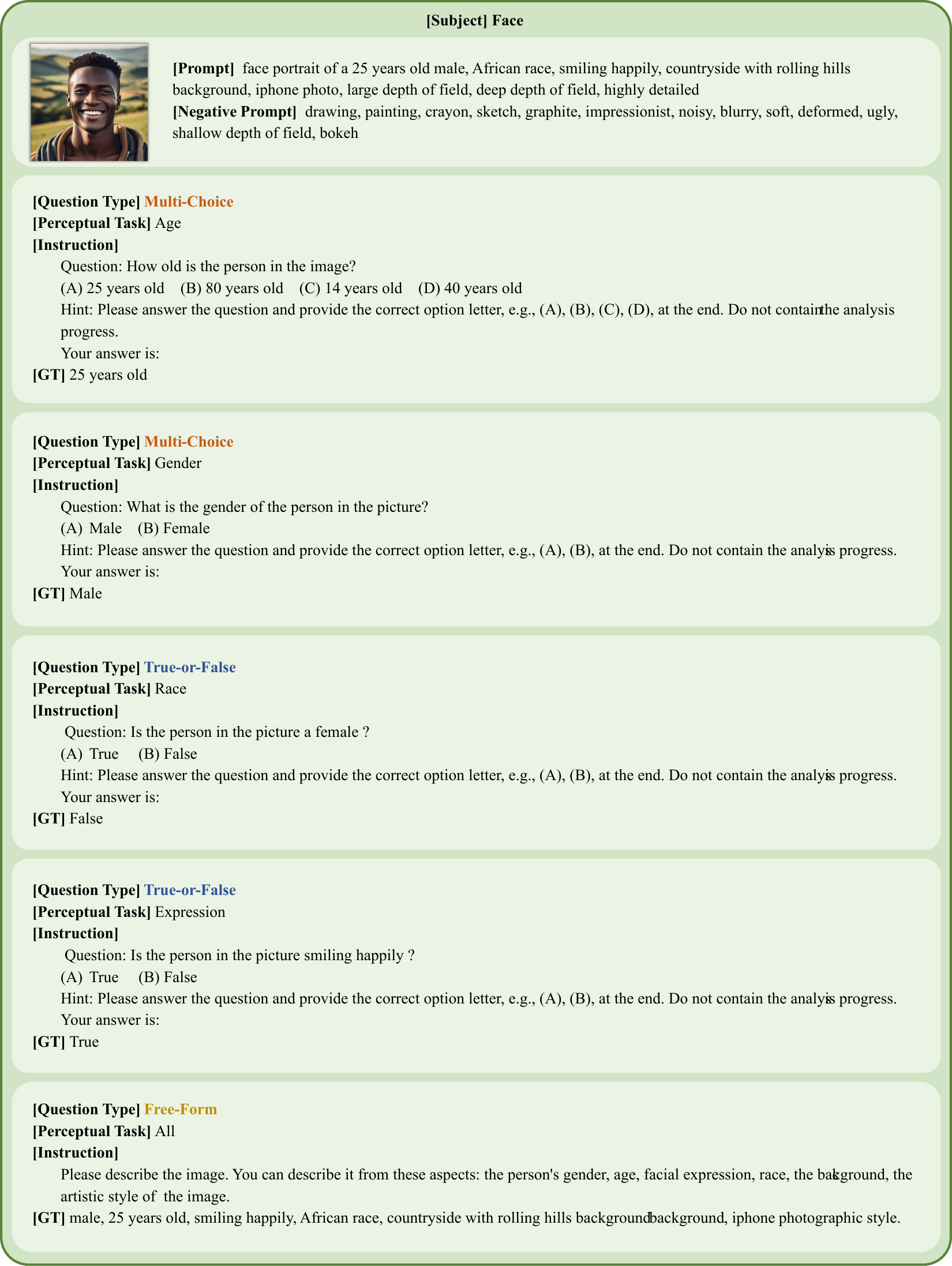}
    \caption{Face}
    \label{fig:face_example}
\end{figure}

\begin{figure}[h]
    \centering
    \includegraphics[width=1\linewidth]{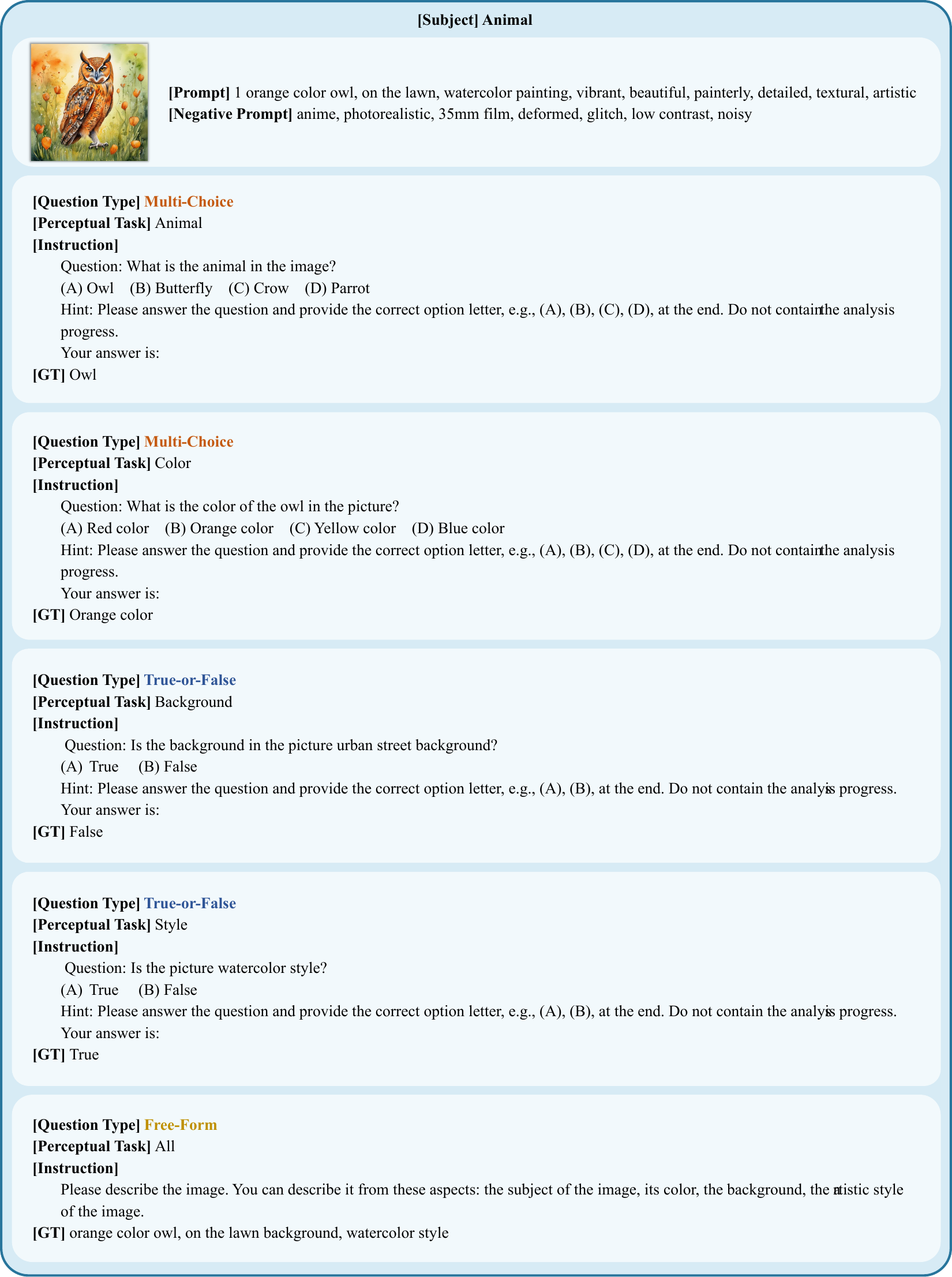}
    \caption{Animal}
    \label{fig:animal_example}
\end{figure}

\begin{figure}[h]
    \centering
    \includegraphics[width=1\linewidth]{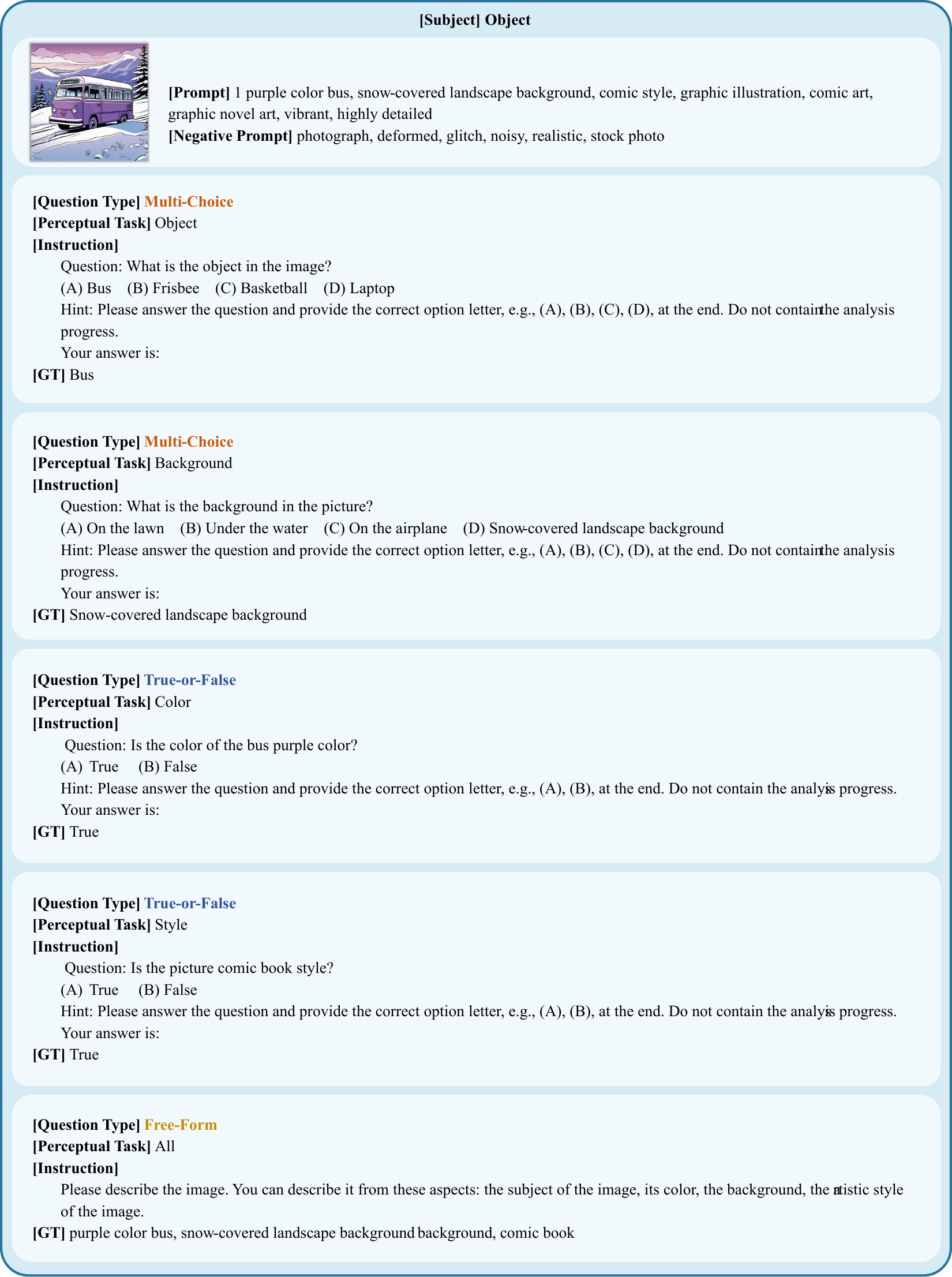}
    \caption{Object}
    \label{fig:object_example}
\end{figure}

\begin{figure}[h]
    \centering
    \includegraphics[width=1\linewidth]{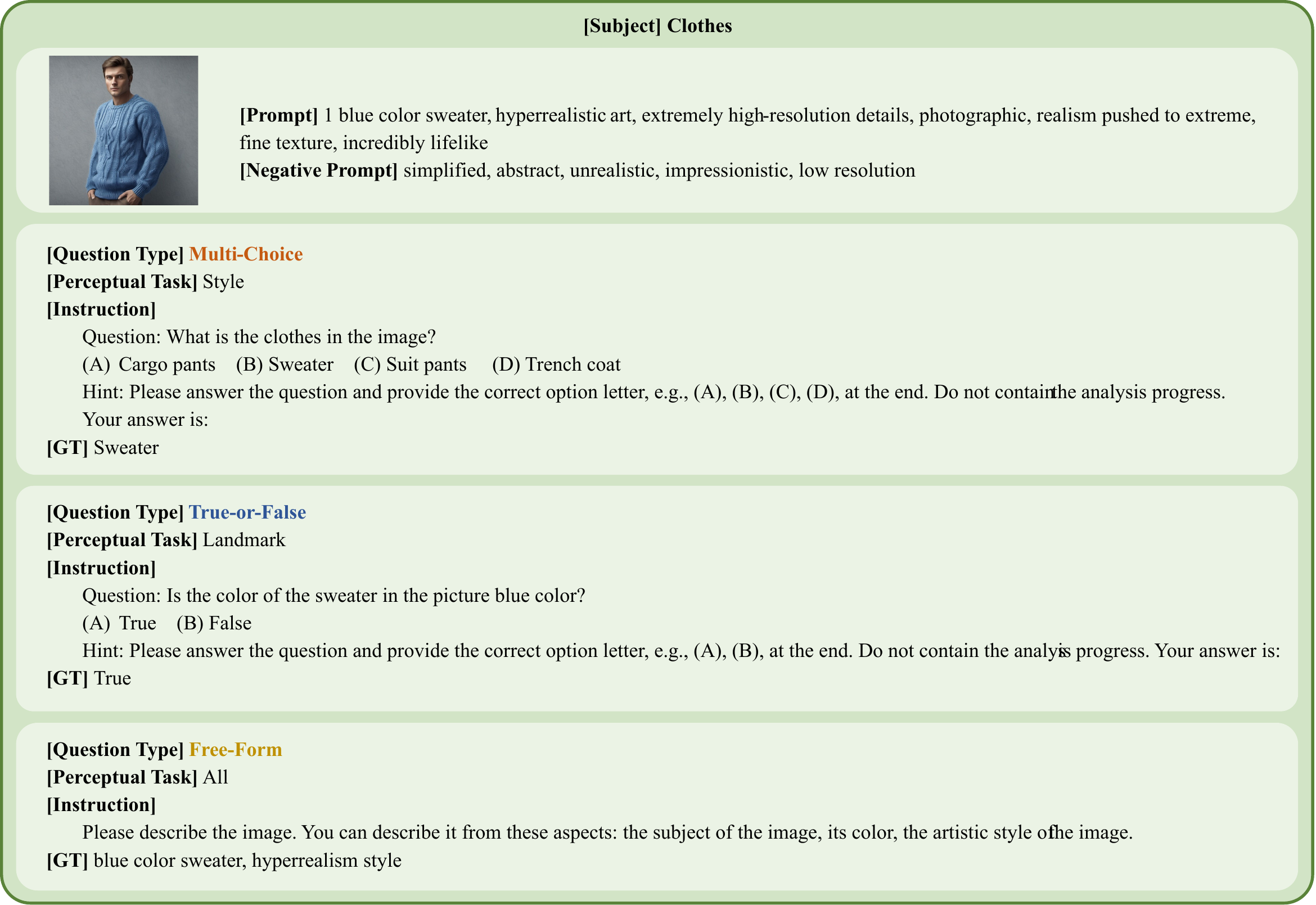}
    \caption{Clothes}
    \label{fig:clothes_example}
\end{figure}

\begin{figure}[h]
    \centering
    \includegraphics[width=1\linewidth]{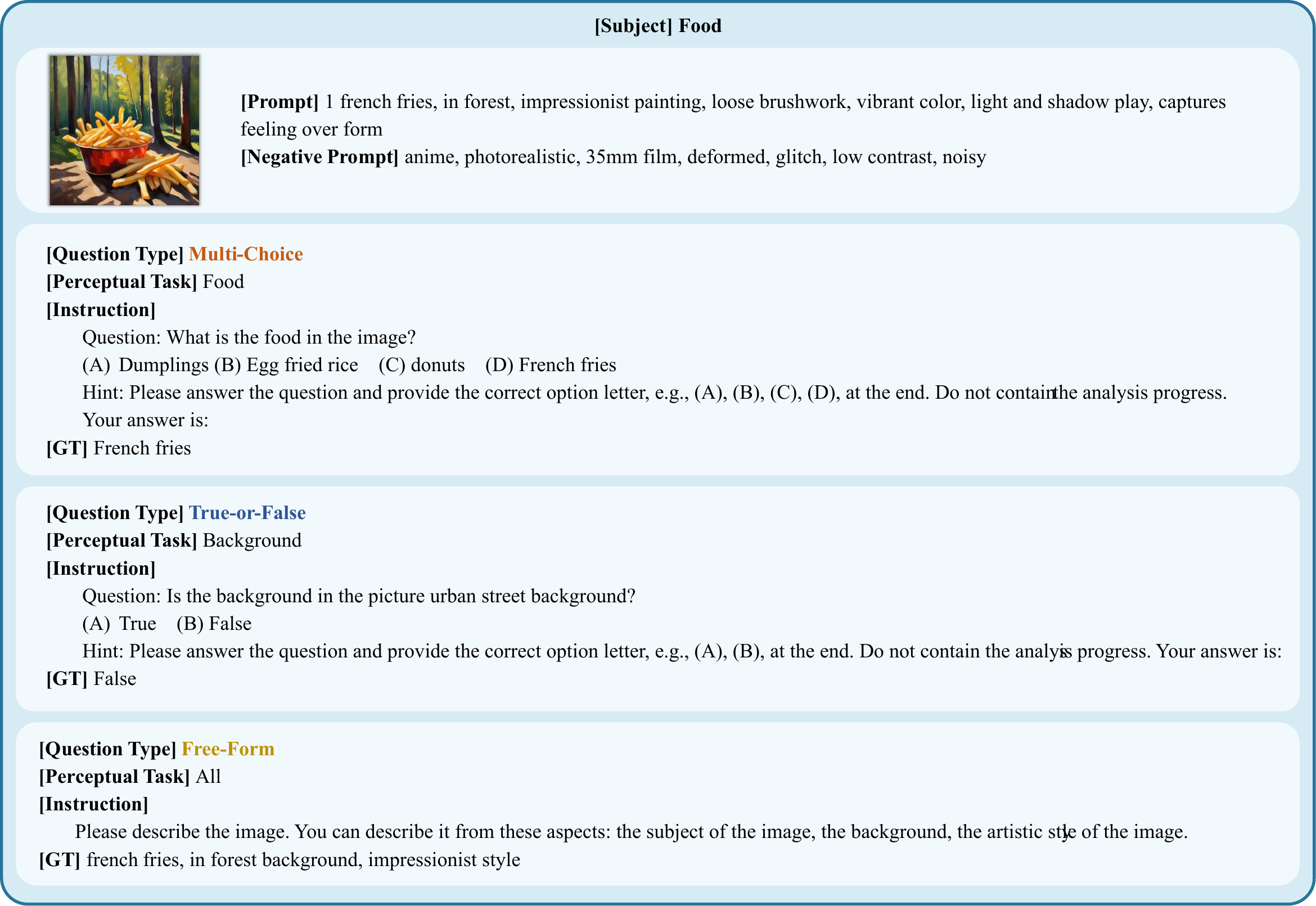}
    \caption{Food}
    \label{fig:food_example}
\end{figure}

\begin{figure}[h]
    \centering
    \includegraphics[width=1\linewidth]{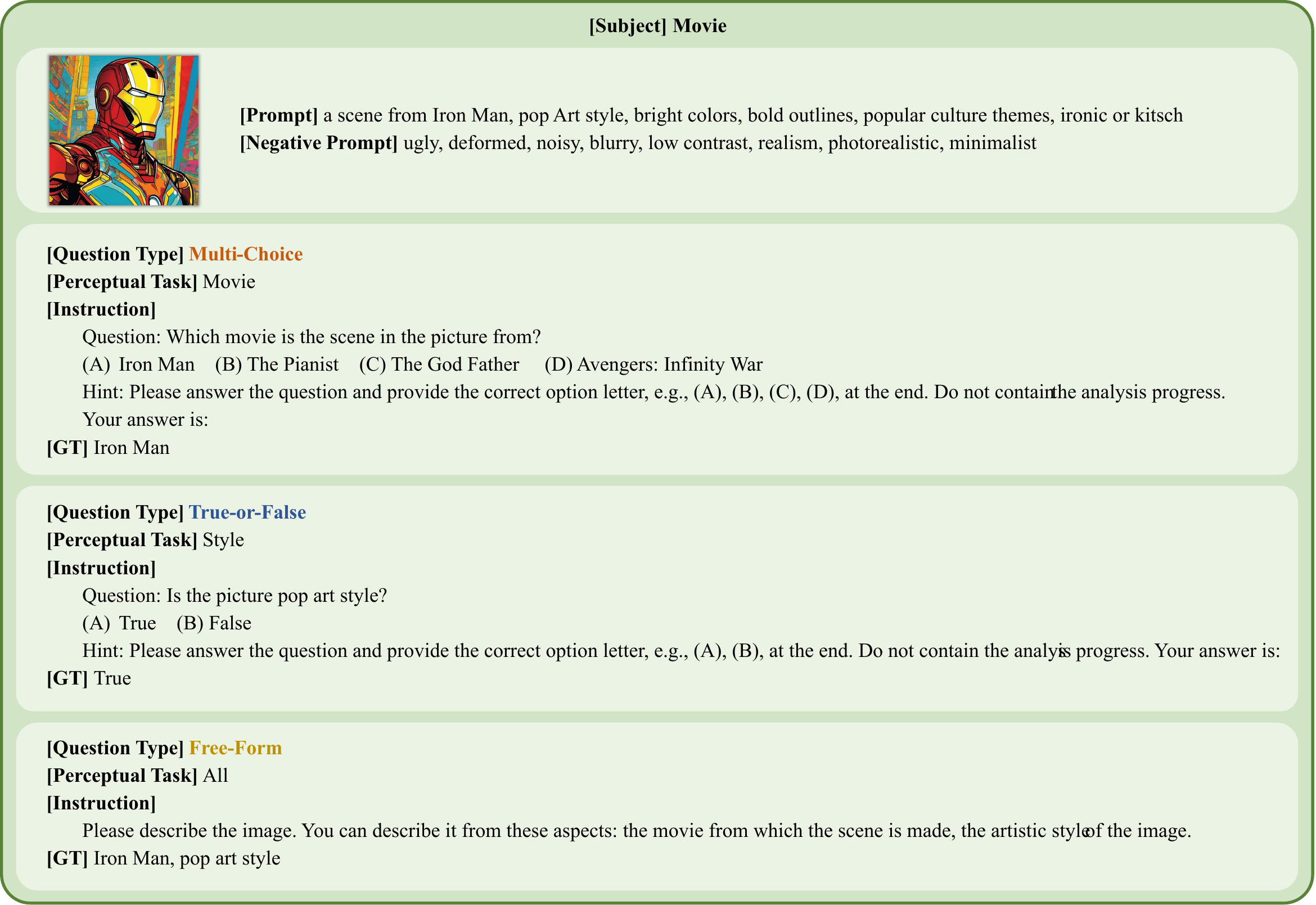}
    \caption{Movie}
    \label{fig:movie_example}
\end{figure}

\begin{figure}[h]
    \centering
    \includegraphics[width=1\linewidth]{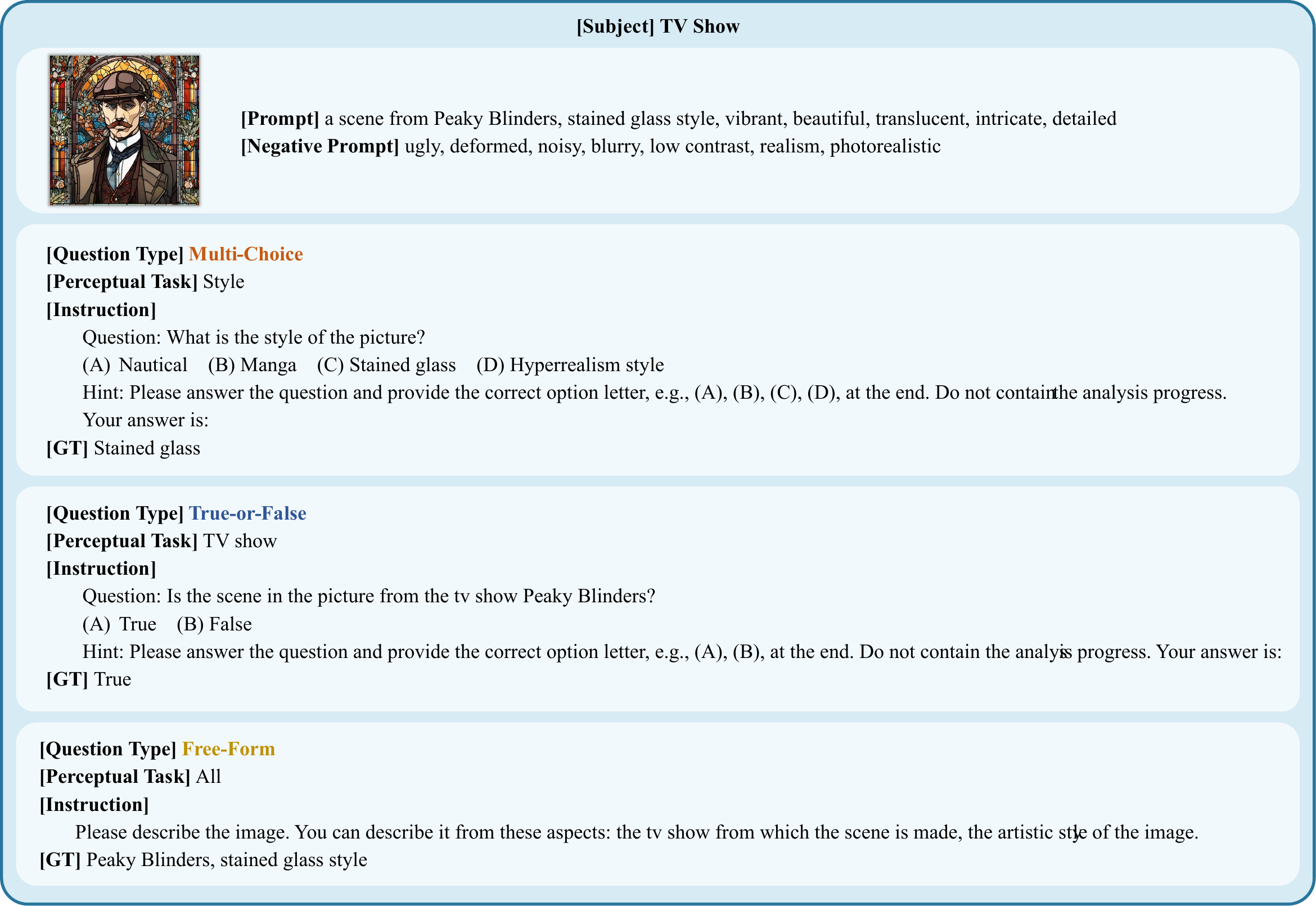}
    \caption{TV}
    \label{fig:tv_example}
\end{figure}

\begin{figure}[h]
    \centering
    \includegraphics[width=1\linewidth]{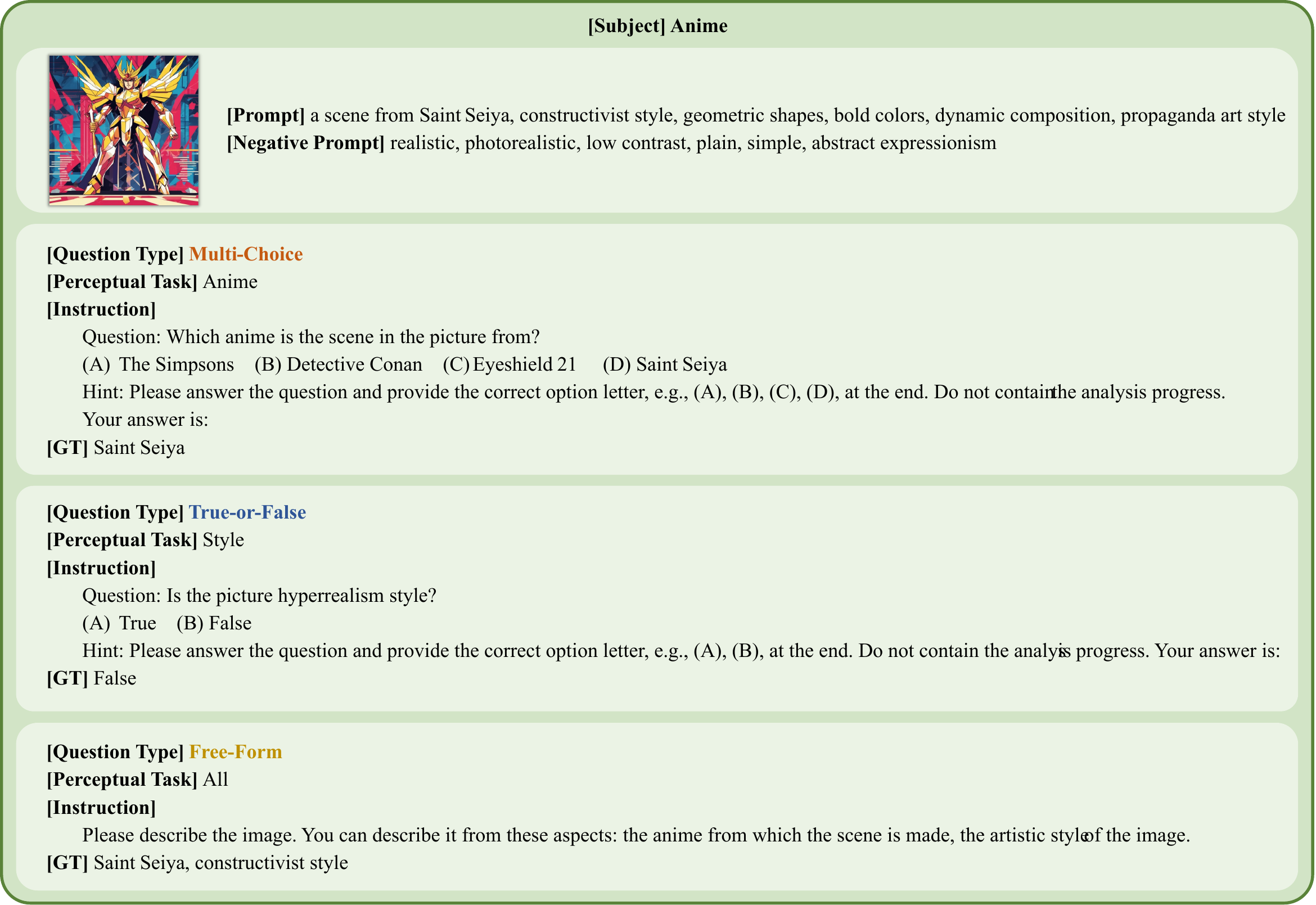}
    \caption{Anime}
    \label{fig:anime_example}
\end{figure}

\begin{figure}[h]
    \centering
    \includegraphics[width=1\linewidth]{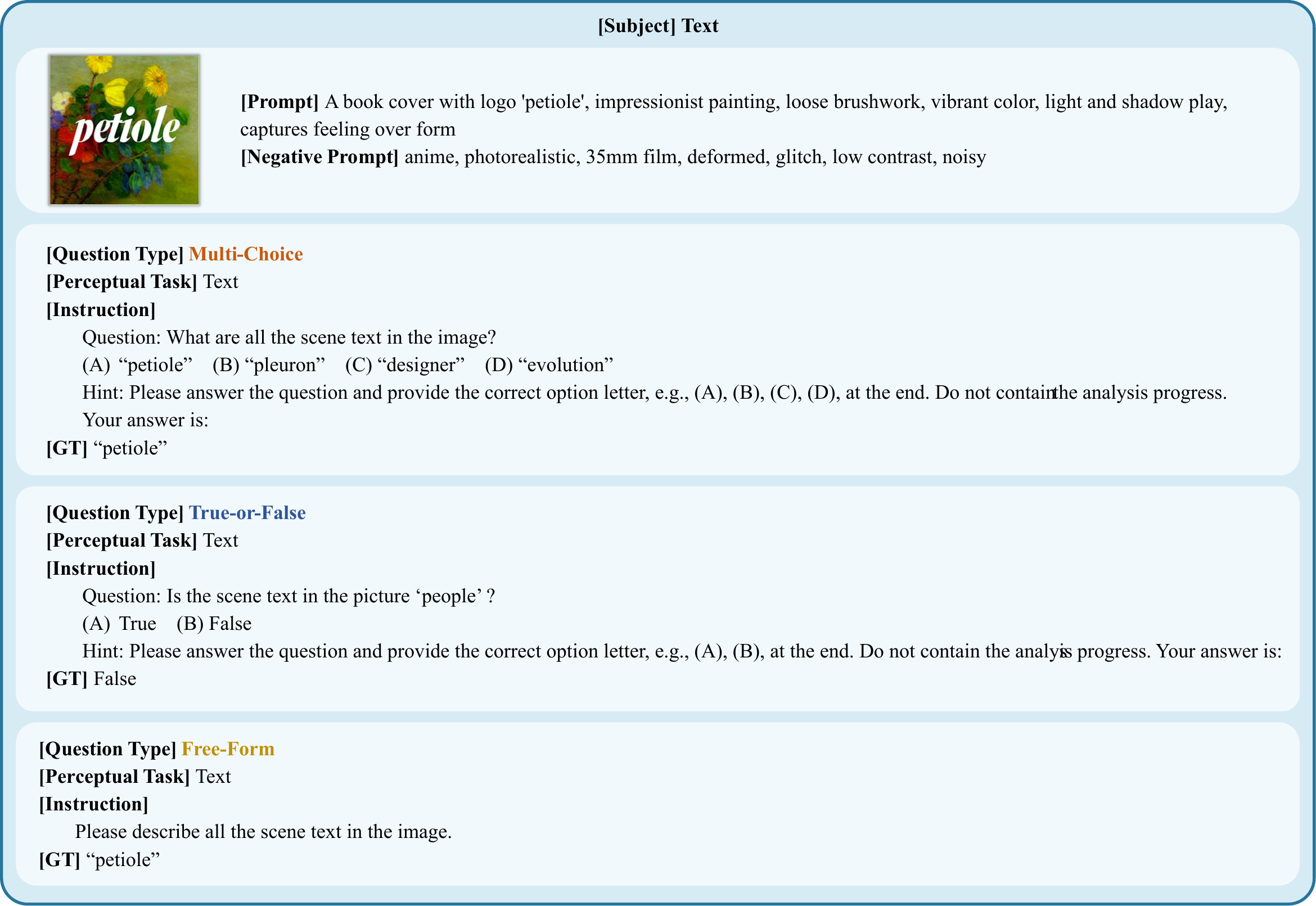}
    \caption{OCR}
    \label{fig:ocr_example}
\end{figure}

\clearpage
\section{JSON structure for evaluation}

Here, We present two example JSON structures generated by Dysca for evaluation in Listing \ref{lst:json-example}.

\lstdefinestyle{jsonStyle}{
  basicstyle=\ttfamily\small,   
  keywordstyle=\color{blue}, 
  commentstyle=\color{green},
  stringstyle=\color{red}, 
  breaklines=true,  
  postbreak=\mbox{$\hookrightarrow$},
  showstringspaces=false, 
  xleftmargin=15pt,  
  frame=single, 
}

\begin{figure}[h!]
  \centering
  \begin{lstlisting}[style=jsonStyle, caption={Example JSON structure for evaluation}, label={lst:json-example}]
{
    "id": 1,
    "images": [
      "images/1.png"
    ],
    "prompt": "face portrait of a 25 years old female, Indian race, calm and placid, on Mars, retro arcade style, 8-bit, pixelated, vibrant, classic video game, old school gaming, reminiscent of 80s and 90s arcade games",
    "negative_prompt": "modern, ultra-high resolution, photorealistic, 3D",
    "instruction": "Question: What is the background in the picture?\nChoices:\n(A) on the lawn\n(B) under the water background\n(C) sunset background\n(D) on Mars\nHint: Please answer the question and provide the correct option letter, e.g., (A), (B), (C), (D), at the end. Do not contain the analysis progress.\nYour answer is:",
    "question": "What is the background in the picture?",
    "answer": "on Mars",
    "question_type": "multi choice",
    "task": "recognition",
    "options": [
      "on the lawn",
      "under the water background",
      "sunset background",
      "on Mars"
    ],
    "question_majority": "background",
    "granularity": "coarse"
},
{
    "id": 2,
    "images": [
      "images/2.png"
    ],
    "prompt": "1 white color cat, on Mars, thick layered papercut art of, deep 3D, volumetric, dimensional, depth, thick paper, high stack, heavy texture, tangible layers",
    "negative_prompt": "2D, flat, thin paper, low stack, smooth texture, painting, drawing, photo, deformed",
    "instruction": "Question: Is there a cat in the picture?\nChoices:\n(A) False\n(B) True\nHint: Please answer the question and provide the correct option letter, e.g., (A), (B), (C), (D), at the end. Do not contain the analysis progress.\nYour answer is:",
    "question": "Is there a cat in the picture?",
    "answer": "True",
    "question_type": "true or false",
    "task": "recognition",
    "options": [
      "False",
      "True"
    ],
    "question_majority": "animal",
    "granularity": "fine"
}
  \end{lstlisting}
\end{figure}

\section{Data Sheet}

We follow the documentation frameworks provided by Gebru et al. \cite{gebru2021datasheets}.

\subsection{Motivation}

\textbf{For what purpose was the dataset created?} Was there a specific task in mind? Was there a specific gap that needed to be filled? Please provide a description.

\begin{itemize}
    \item The proposed dataset is used for evaluating current LVLMs perception ability. We use the synthesis images to prevent the potential data leakage problem in current benchmarks. The dataset test LVLMs in 20 subtasks under 4 scenarios and 3 question type, revealing the existing drawbacks of current LVLMs.
\end{itemize}

\textbf{Who created the dataset (e.g., which team, research group) and on behalf of which entity (e.g., company, institution, organization)?}

\begin{itemize}
    \item The dataset is created by the AI Safety and Trustworthiness Group on behalf of Key Laboratory of AI Safety of CAS, Institute of Computing Technology.
\end{itemize}

\textbf{Who funded the creation of the dataset?} If there is an associated grant, please provide the name of the grant or and the grant name and number.

\begin{itemize}
    \item This work is partially supported by Strategic Priority Research Program of the Chinese Academy of Sciences (No. XDB0680202),
Beijing Nova Program (20230484368), Suzhou Frontier
Technology Research Project (No. SYG202325),
and Youth Innovation Promotion Association CAS.
\end{itemize}

\subsection{Composition}

\textbf{What do the instances that comprise the dataset
    represent (e.g., documents, photos, people, countries)?} Are there multiple types of instances (e.g., movies, users, and ratings; people and interactions between them; nodes and edges)? Please provide a description.

  \begin{itemize}
    \item We show the instances list in Tab. \ref{tab:category_info}. The detailed word we collect for metadata $M$ are shown at \url{https://github.com/Robin-WZQ/Dysca}.
\end{itemize}

\textbf{How many instances are there in total (of each type, if appropriate)?}

\begin{itemize}
    \item There are a total of 20 subtasks in our Dysca. For details of each subtasks please see refer Fig. \ref{fig:subtasks}.
\end{itemize}

\textbf{Does the dataset contain all possible instances or is it a sample (not necessarily random) of instances from a larger set?}  If the dataset is a sample, then what is the larger set? Is the sample representative of the larger set (e.g., geographic coverage)? If so, please describe how this representativeness was validated/verified. If it is not representative of the larger set,  please describe why not (e.g., to cover a more diverse range of instances, because instances were withheld or unavailable).

\begin{itemize}
    \item No. The images in Dysca are completely generated from scratch.
\end{itemize}

\textbf{What data does each instance consist of?} ``Raw'' data
  (e.g., unprocessed text or images) or features? In either case, please provide a description.

\begin{itemize}
    \item Each instance consists of the prompt, the image generated by stable diffusion, the question and corresponding answer.
\end{itemize}

\textbf{Is there a label or target associated with each
    instance?} If so, please provide a description.

\begin{itemize}
    \item Yes, Dysca provides the ground truth for each instance.
\end{itemize}

\textbf{Is any information missing from individual instances?}
  If so, please provide a description, explaining why this information
  is missing (e.g., because it was unavailable). This does not include
  intentionally removed information, but might include, e.g., redacted
  text.

\begin{itemize}
    \item No.
\end{itemize}

\textbf{Are relationships between individual instances made
    explicit (e.g., users' movie ratings, social network links)?} If so, please describe how these relationships are made explicit.

\begin{itemize}
    \item There are no relationships between individual instances.
\end{itemize}

\textbf{Are there recommended data splits (e.g., training,
    development/validation, testing)?} If so, please provide a
  description of these splits, explaining the rationale behind them.

\begin{itemize}
    \item Following our motivation, the entire proposed dataset is used for testing purposes.
\end{itemize}

\textbf{Are there any errors, sources of noise, or redundancies in the dataset?} If so, please provide a description.

\begin{itemize}
    \item Errors in image generation resulting from stable diffusion are unavoidable. However, we have performed dataset cleaning to minimize these errors. Furthermore, the stability experiment in Appendix B demonstrates that these errors do not affect the overall evaluation results of the dataset.
\end{itemize}

\textbf{Is the dataset self-contained, or does it link to or
    otherwise rely on external resources (e.g., websites, tweets, other datasets)?} If it links to or relies on external resources, a) are there guarantees that they will exist, and remain constant, over time; b) are there official archival versions of the complete dataset (i.e., including the external resources as they existed at
    the time the dataset was created); c) are there any restrictions (e.g., licenses, fees) associated with any of the external resources that might apply to a dataset consumer? Please provide descriptions of all external resources and any restrictions associated with them, as well as links or other access points, as appropriate.

\begin{itemize}
    \item The proposed Dysca dose not rely on any external resources. 
\end{itemize}

\textbf{Does the dataset contain data that might be considered confidential (e.g., data that is protected by legal privilege or by doctor-patient confidentiality, data that includes the content of individuals' non-public communications)?} If so, please provide a description.

\begin{itemize}
    \item No.
\end{itemize}

\textbf{Does the dataset contain data that, if viewed directly,
    might be offensive, insulting, threatening, or might otherwise
    cause anxiety?} If so, please describe why.

\begin{itemize}
    \item No. To ensure that the generated images do not contain offensive, insulting, threatening, or anxiety-inducing content, we manually filter out words from the metadata $M$ that could potentially trigger the diffusion model to generate such images. Safety checker also used to further avoid unsafe image generation. 
\end{itemize}

\textbf{Does the dataset relate to people? } If not, you may skip the remaining questions in this section.

\begin{itemize}
    \item Yes.
\end{itemize}

\textbf{Does the dataset identify any subpopulations (e.g., by
    age, gender)?} If so, please describe how these subpopulations are identified and provide a description of their respective distributions within the dataset.

\begin{itemize}
    \item Yes. There are the age, gender and race recognition subtasks in Dysca. Each of them are divided to several subpopulations and the selection of these subpopulations is based on the ability of stable diffusion to generate the representative subpopulations.
\end{itemize}

\textbf{Is it possible to identify individuals (i.e., one or
    more natural persons), either directly or indirectly (i.e., in combination with other data) from the dataset?} If so, please describe how.

\begin{itemize}
    \item Yes. There is the celebrity recognition task in our dataset, where 50 well-know celebrity are chosen. We choose the celebrity who can be generated well by stable diffusion XL.
\end{itemize}

\textbf{Does the dataset contain data that might be considered
    sensitive in any way (e.g., data that reveals race or ethnic origins, sexual orientations, religious beliefs, political opinions or union memberships, or locations; financial or health data; biometric or genetic data; forms of government identification, such as social security numbers; criminal history)?} If so, please provide a description.

\begin{itemize}
    \item No, our benchmark does not contain any sensitive data.
\end{itemize}

\subsection{Collection Process}

\textbf{How was the data associated with each instance acquired?} Was the data directly observable (e.g., raw text, movie ratings), reported by subjects (e.g., survey responses), or indirectly inferred/derived from other data (e.g., part-of-speech tags, model based guesses for age or language)? If data was reported by subjects or indirectly inferred/derived from other data, was the data validated/verified? If so, please describe how.

\begin{itemize}
    \item We display the detailed explanation in Tab. \ref{tab:category_info}.
\end{itemize}

\textbf{What mechanisms or procedures were used to collect the
data (e.g., hardware apparatus or sensor, manual human curation, software program, software API)?} How were these mechanisms or procedures validated?

\begin{itemize}
    \item We collect the data by manual human curation.
\end{itemize}

\textbf{If the dataset is a sample from a larger set, what was the
 sampling strategy (e.g., deterministic, probabilistic with
 specific sampling probabilities)?}

\begin{itemize}
    \item No.
\end{itemize}

\textbf{Who was involved in the data collection process (e.g., students, crowdworkers, contractors) and how were they compensated (e.g., how much were crowdworkers paid)?}

\begin{itemize}
    \item We collect the metadata of Tab. \ref{tab:category_info} by authors. The images are generated by stable diffusion and labels of each image are also automatically generated.
\end{itemize}

\textbf{Over what timeframe was the data collected? Does this timeframe match the creation timeframe of the data associated with the instances (e.g., recent crawl of old news articles)?} If not, please describe the timeframe in which the data associated with the instances was created.

\begin{itemize}
    \item Our dataset was conducted in April of 2024, but the results do not depend on the date of data collection.
\end{itemize}

\textbf{Were any ethical review processes conducted (e.g., by an
 institutional review board)?} If so, please provide a description
 of these review processes, including the outcomes, as well as
 a link or other access point to any supporting documentation.

\begin{itemize}
    \item No.
\end{itemize}

\textbf{Did you collect the data from the individuals in question directly, or obtain it via third parties or other sources (e.g., websites)?}

\begin{itemize}
    \item No.
\end{itemize}

\textbf{Were the individuals in question notified about the data
 collection?} If so, please describe (or show with screenshots
 or other information) how notice was provided, and provide a
 link or other access point to, or otherwise reproduce, the exact
 language of the notification itself.

\begin{itemize}
    \item N/A. Our Dysca does not involve the collection from the individuals.
\end{itemize}

\textbf{Did the individuals in question consent to the collection
 and use of their data?} If so, please describe (or show with
 screenshots or other information) how consent was requested
 and provided, and provide a link or other access point to, or otherwise reproduce, the exact language to which the individuals consented.

 \begin{itemize}
    \item N/A. Our Dysca does not involve the collection from the individuals.
\end{itemize}

\textbf{If consent was obtained, were the consenting individuals
 provided with a mechanism to revoke their consent in the
 future or for certain uses?} If so, please provide a description,
 as well as a link or other access point to the mechanism (if
 appropriate).

\begin{itemize}
    \item N/A. Our Dysca does not involve the collection from the individuals.
\end{itemize}

\textbf{Has an analysis of the potential impact of the dataset and
 its use on data subjects (e.g., a data protection impact
 analysis) been conducted?} If so, please provide a description
 of this analysis, including the outcomes, as well as a link or
 other access point to any supporting documentation.

\begin{itemize}
    \item No.
\end{itemize}

\subsection{Preprocessing/cleaning/labeling}

\textbf{Was any preprocessing/cleaning/labeling of the data done
    (e.g., discretization or bucketing, tokenization, part-of-speech
    tagging, SIFT feature extraction, removal of instances, processing
    of missing values)?} If so, please provide a description. If not,
  you may skip the remaining questions in this section.

\begin{itemize}
    \item Yes. We leverage the off-the-shelf models, i.e., PP-OCRv3 \cite{li2022ppocrv3} and CLIP-L-14 \cite{radford2021learning_clip}, to clean the data. PP-OCRv3 \cite{li2022ppocrv3} is leveraged as the filter to exclude the failure image that TextDiffusion2 \cite{chen2023textdiffuser} generates the wrong text on the image. For the other images, we use CLIP-L-14 \cite{radford2021learning_clip} to filter out the images with low text-image consistency. 
\end{itemize}

\textbf{Was the ``raw'' data saved in addition to the preprocessed/cleaned/labeled data (e.g., to support unanticipated future uses)?} If so, please provide a link or other access point to the ``raw'' data.

\begin{itemize}
    \item Yes. We have saved all the data. However, most of these images are filtered and considered to be useless.
\end{itemize}

\textbf{Is the software that was used to preprocess/clean/label the data available?} If so, please provide a link or other access point.

\begin{itemize}
    \item Yes. CLIP-L-14 can be downloaded at \url{https://huggingface.co/docs/transformers/v4.41.3/en/model_doc/clip#transformers.CLIPModel}. 
    
    PP-OCRv3 can be downloaded at \url{https://github.com/PaddlePaddle/PaddleOCR/blob/main/README_en.md}
\end{itemize}

\subsection{Uses}

\textbf{Has the dataset been used for any tasks already?} If so, please provide a description.

\begin{itemize}
    \item No. The proposed dataset is the novel one which is used for evaluation current LVLMs perception ability.
\end{itemize}

\textbf{Is there a repository that links to any or all papers or systems that use the dataset?} If so, please provide a link or other access point.

\begin{itemize}
    \item Yes. We plan to create a section on the project homepage to keep track of
 LVLMs papers for researchers to analyze and compare.
\end{itemize}

\textbf{What (other) tasks could the dataset be used for?}

\begin{itemize}
    \item In this work, we do not explore the possibility of utilizing our benchmark for model training / fine-tuning. Our primary goal in this paper is to provide a large-scale evaluation benchmark that addresses the issue of data leakage in current multimodal evaluation benchmarks and offers evaluation results across multiple subtasks, scenarios, question types and styles. Nevertheless, considering that Dysca has the capability to synthesize high-resolution and unlimited amounts of annotated multimodal data, we believe that Dysca also holds potential as a training data synthesis tool for LVLMs.
\end{itemize}

\textbf{Is there anything about the composition of the dataset or the way it was collected and preprocessed/cleaned/labeled that might impact future uses?} For example, is there anything that a dataset consumer might need to know to avoid uses that could result in unfair treatment of individuals or groups (e.g., stereotyping, quality of service issues) or other risks or harms (e.g., legal risks, financial harms)? If so, please provide a description. Is there anything a dataset consumer could do to mitigate these risks or harms?

\begin{itemize}
    \item Yes.
\end{itemize}

\textbf{Are there tasks for which the dataset should not be used?} If so, please provide a description.

\begin{itemize}
    \item The proposed dataset should not be used to generate offensive data. 
\end{itemize}

\subsection{Distribution}

\textbf{Will the dataset be distributed to third parties outside of the entity (e.g., company, institution, organization) on behalf of which the dataset was created?} If so, please provide a description.

\begin{itemize}
    \item Yes.
\end{itemize}

\textbf{How will the dataset will be distributed (e.g., tarball on website, API, GitHub)?} Does the dataset have a digital object identifier (DOI)?

\begin{itemize}
    \item We will open-source our dataset on our GitHub project homepage. At
 the moment, we do not have a DOI number.
\end{itemize}

\textbf{When will the dataset be distributed?}

\begin{itemize}
    \item The dataset can be downloaded right now.
\end{itemize}

\textbf{Will the dataset be distributed under a copyright or other intellectual property (IP) license, and/or under applicable terms of use (ToU)?} If so, please describe this license and/or ToU, and provide a link or other access point to, or otherwise reproduce, any relevant licensing terms or ToU, as well as any fees associated with these restrictions.

\begin{itemize}
    \item The licence of Dysca is \href{https://huggingface.co/stabilityai/stable-diffusion-xl-base-1.0/blob/main/LICENSE.md}{"CreativeML Open RAIL++-M"}, which follows the licence set by the Stable Diffusion XL.
\end{itemize}

\textbf{Have any third parties imposed IP-based or other restrictions on the data associated with the instances?} If so, please describe these restrictions, and provide a link or other access point to, or otherwise reproduce, any relevant licensing terms, as well as any fees associated with these restrictions.

\begin{itemize}
    \item No.
\end{itemize}

\textbf{Do any export controls or other regulatory restrictions apply to the dataset or to individual instances?} If so, please describe these restrictions, and provide a link or other access point to, or otherwise reproduce, any supporting documentation.

\begin{itemize}
    \item Not yet.
\end{itemize}

\subsection{Maintenance}

\textbf{Who will be supporting/hosting/maintaining the dataset?}

\begin{itemize}
    \item Jie Zhang and Zhongqi Wang are hosting and maintaining the dataset.
\end{itemize}

\textbf{How can the owner/curator/manager of the dataset be contacted (e.g., email address)?}

\begin{itemize}
    \item  Email: zhangjie@ict.ac.cn
\end{itemize}

\textbf{Is there an erratum?} If so, please provide a link or other access point.

\begin{itemize}
    \item No.
\end{itemize}

\textbf{Will the dataset be updated (e.g., to correct labeling
    errors, add new instances, delete instances)?} If so, please describe how often, by whom, and how updates will be communicated to dataset consumers (e.g., mailing list, GitHub)?

\begin{itemize}
    \item There are no plans at the moment, but if there are updates, they will
 be announced, and the download source will be updated on the project
 homepage.
\end{itemize}

\textbf{If the dataset relates to people, are there applicable
    limits on the retention of the data associated with the instances (e.g., were the individuals in question told that their data would be retained for a fixed period of time and then deleted)?} If so, please describe these limits and explain how they will be enforced.

\begin{itemize}
    \item No.
\end{itemize}

\textbf{Will older versions of the dataset continue to be           
    supported/hosted/maintained?} If so, please describe how. If not, please describe how its obsolescence will be communicated to dataset consumers.

\begin{itemize}
    \item Yes. If there are any updates, the previous version of the dataset will also be shared on website for download.
\end{itemize}

\textbf{If others want to extend/augment/build on/contribute to the dataset, is there a mechanism for them to do so?} If so, please provide a description. Will these contributions be
  validated/verified? If so, please describe how. If not, why not? Is there a process for communicating/distributing these contributions to dataset consumers? If so, please provide a description.

\begin{itemize}
    \item Yes. We welcome and encourage researchers to extend/augment/build
 on/contribute to our dataset for non-profit purposes without the need
 for prior notification.
\end{itemize}

\end{document}